\documentclass[letterpaper]{article} % DO NOT CHANGE THIS
\usepackage[submission]{aaai23}  % DO NOT CHANGE THIS
\usepackage{times}  % DO NOT CHANGE THIS
\usepackage{helvet}  % DO NOT CHANGE THIS
\usepackage{courier}  % DO NOT CHANGE THIS
\usepackage[hyphens]{url}  % DO NOT CHANGE THIS
\usepackage{graphicx} % DO NOT CHANGE THIS
\usepackage{natbib}  % DO NOT CHANGE THIS AND DO NOT ADD ANY OPTIONS TO IT
\usepackage{caption} % DO NOT CHANGE THIS AND DO NOT ADD ANY OPTIONS TO IT
\usepackage{algorithm}
\usepackage{algorithmic}

\usepackage{newfloat}
\usepackage{listings}
\DeclareCaptionStyle{ruled}{labelfont=normalfont,labelsep=colon,strut=off} % DO NOT CHANGE THIS
\floatstyle{ruled}
\newfloat{listing}{tb}{lst}{}
\floatname{listing}{Listing}

\usepackage{amsmath}
\usepackage{amssymb}

\newtheorem{corollary}{Corollary}

\newcommand{\argmax}{arg\,max}

\newcommand{\eg}{\textit{e.g. }}
\newcommand{\ie}{\textit{i.e. }}

\title{Certified Robustness in Federated Learning}
\author{
    Written by AAAI Press Staff\textsuperscript{\rm 1}\thanks{With help from the AAAI Publications Committee.}\\
    AAAI Style Contributions by Pater Patel Schneider,
    Sunil Issar,\\
    J. Scott Penberthy,
    George Ferguson,
    Hans Guesgen,
    Francisco Cruz\equalcontrib,
    Marc Pujol-Gonzalez\equalcontrib
}
\affiliations{
    \textsuperscript{\rm 1}Association for the Advancement of Artificial Intelligence\\
    1900 Embarcadero Road, Suite 101\\
    Palo Alto, California 94303-3310 USA\\
    publications23@aaai.org
}

\usepackage{bibentry}
\begin{document}

\maketitle

\begin{abstract}
Federated learning has recently gained significant attention and popularity due to its effectiveness in training machine learning models on distributed data privately.
However, as in the single-node supervised learning setup, models trained in federated learning suffer from vulnerability to imperceptible input transformations known as adversarial attacks, questioning their deployment in security-related applications.
In this work, we study the interplay between federated training, personalization, and certified robustness.
In particular, we deploy randomized smoothing, a widely-used and scalable certification method, to certify deep networks trained on a federated setup against input perturbations and transformations.
We find that the simple federated averaging technique is effective in building not only more accurate, but also more certifiably-robust models, compared to training solely on local data.
We further analyze personalization, a popular technique in federated training that increases the model's bias towards local data, on robustness.
We show several advantages of personalization over both~(that is, only training on local data and federated training) in building more robust models with faster training. 
Finally, we explore the robustness of mixtures of global and local~(\ie personalized) models, and find that the robustness of local models degrades as they diverge from the global model.
\end{abstract}

\section{Introduction}
Machine learning models in general, and deep neural networks (DNNs) in particular, have demonstrated remarkable performance in several domains, including computer vision~\cite{krizhevsky2012imagenet} and natural language processing~\cite{hinton2012speech}, among others~\cite{lecun2015deep}. 
Despite this success, DNNs are brittle against small carefully-crafted perturbations in their input space~\cite{szegedy2014intriguing, goodfellow2015explaining}. 
That is, a DNN $f_\theta : \mathbb R^d\rightarrow \mathcal P(\mathcal Y)$ can produce different predictions for the inputs $x \in \mathbb R^d$ and $x+\delta$, although the adversarial perturbation $\delta$ is small, and thus $x$ and $x+\delta$ are indistinguishable to the human eye. 
Moreover, DNNs are also susceptible to input transformations that preserve an image's semantic information, such as rotation, translation, and scaling~\cite{engstrom2018rotation}. 
This observation raises security concerns about deploying such models in security-critical applications, \ie self-driving cars.
Consequently, this phenomenon sparked substantial research efforts towards building machine learning models that are not only accurate, but also robust.
That is, building models that both correctly classify an input $x$, and maintain their prediction as long $x$ preserves its semantic content. 
Further interest arose in theoretically characterizing the output of models whose input was subjected to perturbations.
Whenever a theoretical guarantee exists about a model's robustness for classifying inputs, $f_\theta$ is said to be \emph{certifiably} robust~\cite{wong2018provable}.
Among several methods that built such certifiably robust models, Randomized Smoothing (RS)~\cite{cohen2019certified} is arguably one of the most effective, scalable, and popular approaches to certify DNNs. 
In a nutshell, RS predicts the most probable class when the classifier's input is subjected to additive Gaussian noise. 
While RS directly operated on pixel-intensity perturbations, a later work extended this technique to certify against input deformations, such as rotation and translation~\cite{deformrs}. 

The problem of training large-scale machine learning models in the real world may need to account for data privacy constraints that arise in distributed setups.
Federated learning (FL)~\cite{fedavg, FEDLEARN, konevcny2016federated} is a viable solution to this problem, which is now commonly used in popular services~\cite{kairouz2021advances}.
Although the certified robustness of models in traditional learning settings has been widely studied, few works consider the interaction between certified robustness and the federated setup.
Specifically, federated learning distributes the training process across a (possibly large) set of clients (each with their own local data) and privately builds a global model without leaking nor sharing local raw data.
In practice, federated training is usually performed through a series of communication rounds between the clients and an orchestrating server. 
In each round, a subset of clients trains a local model for multiple steps and sends the model to the server; the server then aggregates the received models and communicates the result back to selected clients~\cite{kairouz2021advances}.
In the federated learning setup, \citet{zizzo2020fat} showed that (1)~adversarial attacks can reduce the accuracy of state-of-the-art federated learning models to 0\%, and (2)~adversarial robustness is higher in the centralized setting than in the federated setting, illustrating the difficulty of enhancing adversarial robustness.
We further investigate these observations, and identify two key and challenging settings for analyzing certified robustness in federated learning: 
\textbf{(i)} each client has insufficient data to fully train a robust model, and
\textbf{(ii)} each client may be interested in building a model that is robust against different transformations.
These settings raise questions about whether and when adversarial robustness benefits from federated training. 

In this work, we set to study the certified robustness of models trained in a federated fashion. 
We deploy Randomized Smoothing to measure certified robustness against both pixel-intensity perturbations and input deformations. 
We first analyze when federated training benefits model robustness. 
We show that in the few local-data regimes---where clients have insufficient data---federated averaging  improves not only model performance but also its robustness. 
Furthermore, we explore how certified robustness gains from personalization~\cite{smith2017federated, mansour2020three, arivazhagan2019federated, explicit_personalization}, \ie fine-tuning the global model on local data. 
We show that personalization can provide significant robustness improvements even when federated averaging operates on models trained without augmentations aimed at enhancing robustness. 
At last, we analyze a version of the recently-proposed mixture between global and personalized models from federated learning~\cite{implicit_personalization} from a certified robustness lens.
In summary, our contributions are three-fold:
\begin{enumerate}
    \item We provide a large-scale study of certified robustness in the federated learning setup. 
    We do so by characterizing the setting, in which federated learning is more beneficial for model robustness than training solely on local data.
    
    \item We present a comprehensive analysis of the personalization effect on robustness. 
    We show how personalization, combined with robust training, compares favorably to robust training on local data, even when individual clients do not employ robust training.
    
    \item  We experimentally analyze the certified robustness benefits of using a mixture of global and local models. 
    We find that as the local model gets closer to the global one via local training steps, certified robustness against both pixel perturbations and input deformations improves.
    
\end{enumerate}

%%%%%%%%%%%%%%%%%%%%%%%%%%%%%%%%%%%%%%%%%%%%%%%%%%%%%%%%%%%%%%%%%%%%%%%%%%%

\section{Related Work}

\paragraph{Certified Adversarial Robustness.}
The seminal work of \citet{szegedy2014intriguing} demonstrated how DNNs are vulnerable to small perturbations in their input, now called adversarial attacks.
Subsequent works noted how the brittleness against these attacks was widespread and easily exploitable~\cite{goodfellow2015explaining}.
This observation led to further works on defense mechanisms that improved adversarial robustness~\cite{kannan2018adversarial, madry2018towards}, and on attacks that aimed at breaking these mechanisms~\cite{carlini2017towards, moosavi2016deepfool}.
The development of defenses and attacks of increasing sophistication, akin to an arms race, highlighted the difficulty of improving robustness.
In particular, the work of \citet{athalye2018obfuscated} identified how various defenses provided a false sense of security, and attributed such misperception to the difficulty of assessing adversarial robustness~\cite{carlini2019evaluating}.
This difficulty sparkled curiosity on defenses whose robustness could be formally proven~\cite{wong2018provable}; that is, defense mechanisms that feature formal statements about the nonexistence of successful adversarial attacks.
To this aim, \citet{cohen2019certified} proposed Randomized Smoothing~(RS), an approach for certification that proved to successfully scale with models and datasets.
Notably, RS has been successfully combined with adversarial training~\cite{salman2019provably}, regularization~\cite{Zhai2020MACER:}, and parameter optimization~\cite{alfarra-data, eiras-ancer} for improved robustness.
The original RS formulation certified models against pixel-level additive perturbations, and was studied in the federated learning setup by \citet{chen2021certifiably}.
Recently, DeformRS~\cite{deformrs} reformulated RS to consider parameterized deformations, granting robustness certificates against more general and semantic perturbations, such as rotation and translation.
In this work, we leverage RS and DeformRS to study the certified robustness of models learnt in federated learning.
% against semantic perturbations.

\paragraph{Federated Learning.} 
Federated Learning (FL) is a data-decentralized machine learning setting in which a central server coordinates a large number of clients (\eg mobile devices or whole organizations) to collaboratively train a model~\cite{fedavg, FEDLEARN,konevcny2016federated, wang2021field}.
FL has proven to be a valuable setting to study, finding interest both from research and practical applications~\cite{kairouz2021advances}. 
In particular, FL has been successfully deployed by Google~\cite{hard2018federated}, \citet{apple19wwdc}, \citet{nvidia2020}, and many others, even in safety- and security-critical settings, which motivates our work.
Given how the inherent multiple-devices  nature of FL introduces heterogeneity in the data, interest spurred towards techniques that considered model personalization~\cite{FL-MAML-Jakub,implicit_personalization, hanzely2020lower, li2021ditto} to improve performance on each device's data~\cite{mansour2020three, explicit_personalization, jiang2019improving, paulik2021federated}. 
% Moreover, \citet{yu2020salvaging} showed how naïve federated training may
% % , in the end, 
% be detrimental for some 
% % of the 
% participants.

\paragraph{Robustness in FL.}
Concerns about the adversarial vulnerability of machine learning models have also been studied in the FL setup~\cite{bhagoji2019analyzing}.
While most prior works focused on training-time adversarial attacks such as Byzantine attacks~\cite{he2020byzantine}, and model~\cite{bhagoji2019analyzing} or data poisoning~\cite{steinhardt2017certified, sun2021data}, recent works studied test-time robustness.
In particular, the works of \citet{zizzo2020fat} and \citet{shah2021adversarial} studied the incorporation of adversarial training~\cite{madry2018towards}, arguably the most effective empirical defense against adversarial attacks, into the FL setup.
Closest to our study, is the recent work of \citet{chen2021certifiably} that explored certification via RS in FL.
In contrast to \citet{chen2021certifiably}, instead of pixel-level perturbations, we study realistic (\eg rotation and translation) perturbations.
Furthermore, we provide (1)~a characterization of the setting in which federated averaging---a standard technique in FL---enhances robustness, and (2)~analyze how personalization affects model robustness.

%%%%%%%%%%%%%%%%%%%%%%%%%%%%%%%%%%%%%%%%%%%%%%%%%%%%%%%%%%%%%%%%%%%%%%%%%%%

\section{Methodology}

% \egornote{Maybe I missed it, but I think it is needed to be said directly that we consider only the image classification problem.}

% Our methodology builds on the framework of Randomized Smoothing (RS)~\cite{cohen2019certified} and DeformRS~\cite{deformrs}~(the extension of RS to general input deformations). 
We start with a brief overview on Randomized Smoothing.
% of both. 
% We refer interested readers to the original works for more details.

\subsection{Background on Randomized Smoothing}
\paragraph{Notation.} 
We consider the standard classification problem where a classifier $f_\theta : \mathbb R^d \rightarrow \mathcal P(\mathcal Y)$, \eg a DNN with parameters $\theta$, maps the input $x\in \mathbb R^{d}$ to the probability simplex $\mathcal P(\mathcal Y)$ over $k$ classes, where $\mathcal Y = \{1, 2, \dots, k\}$. 
That is, $\sum_{i=1}^k f_\theta^i(x) = 1$ with $f_\theta^i(x) \geq 0$ being the $i^{\text{th}}$ element of $f_\theta(x)$, representing the probability $x$ belongs to the $i^\text{th}$ class. 
% During inference (test time) and for an input-label pair $(x, y)$ where $y\in\mathcal Y$, the model predicts the label with the highest probability through
% \[
% \hat y = \text{arg}\max_i f^i(x).
% \]
% The model is correctly classifying $x$ if $\hat y = y$. Throughout this chapter, we will assume that $f$ is a trained classifier and will refer to it as $f$ for short.

\paragraph{Randomized Smoothing.}
Randomized smoothing is a method for constructing a ``smooth" classifier from an arbitrary base classifier $f_\theta$. 
In a nutshell, when queried at input $x$, the smooth classifier returns $f_\theta$'s average prediction when its input is subjected to isotropic additive Gaussian noise:
\begin{equation}\label{eq:soft-smoothing}
    g (x) = \mathbb E_{\epsilon \sim \mathcal N(0, \sigma^2 I)}\left [f_\theta(x+\epsilon)\right].
\end{equation}

Let $g$ predict label $c_A$ for input $x$ with some confidence, \ie $\mathbb{E}_\epsilon[f^{c_A}_\theta(x+\epsilon)] = p_A \ge p_B = \max_{c \neq c_A} \mathbb{E}_\epsilon[f^c_\theta(x+\epsilon)]$, then, as proved in~\cite{Zhai2020MACER:}, $g_\theta$'s prediction is certifiably robust at $x$ with certification radius:
\begin{equation}
\begin{aligned}
    \label{eq:certification_radius}
        R =  \frac{\sigma}{2} \left(\Phi^{-1}(p_A) - \Phi^{-1}(p_B)\right).
\end{aligned}
\end{equation}
That is, $\text{arg}\max_i g^i(x+\delta) = \text{arg}\max_i g^i(x),\:\forall \|\delta\|_2 \leq R$, where $\Phi^{-1}$ is the inverse CDF of the standard Gaussian distribution. 

\subsection{Background on DeformRS}
While the original RS framework certified against additive pixel perturbations in images, DeformRS~\cite{deformrs} extended RS to certify against image \textit{deformations}. 
DeformRS achieved this objective by proposing a \textit{parametric-domain smooth classifier}.
In particular, given an image $x$ with pixel coordinates $p$, a parametric deformation function $\nu_\phi$ with parameters $\phi$ (\eg if $\nu$ is a rotation, then $\phi$ is the angle of rotation) and an interpolation function $I_T$, DeformRS defined the parametric smooth classifier as:
\begin{equation}\label{eq:parametric-smooth-classifier}
g_\phi(x, p) = \mathbb E_{\epsilon\sim \mathcal D}\left[ f_\theta(I_T(x, p+\nu_{\phi+\epsilon})) \right].   
\end{equation}
In a nutshell, $g$ outputs the average prediction of $f_\theta$ over transformed versions of the input $x$. 
Note that, in contrast to traditional RS, this formulation perturbs the pixels' \textit{location}, rather than their intensities. % here no longer perturbs pixel intensities, but rather pixel locations. 
DeformRS showed that
% , analogous to the smoothed classifiers in RS,
parametric-domain smooth classifiers are certifiably robust against perturbations to the parameters of the deformation function via the following corollary.

\begin{corollary}\label{cor:parametric-certification} (restated from \cite{deformrs}) Suppose that $g$ predicts class $c_A$ for an input $x$, \ie $c_A = \argmax_i g^i_\phi(x, p)$ with
\[
p_A = g_\phi^{c_A}(x, p) \,\, \emph{and} \,\, p_B = \max_{c\neq c_A}g_\phi^c(x, p)
\]
Then $\argmax_i g^i_{\phi+\delta}(x, p) = c_A$ for all parametric perturbations $\delta$ satisfying:
\begin{align*}
\begin{split}
    &\|\delta\|_1 \leq \sigma \left(p_A - p_B\right) \qquad \qquad \qquad \quad \,\,\,\, \text{for } \mathcal D = \mathcal{U}[-\sigma, \sigma], \\
    &\|\delta\|_2 \leq \frac{\sigma}{2}\left(\Phi^{-1}(p_A) - \Phi^{-1}(p_B)\right) \qquad \text{for } \mathcal D = \mathcal N(0, \sigma^2I).
\end{split}
\end{align*}
\end{corollary}
In short, Corollary~\ref{cor:parametric-certification} states that, as long as the parameters of the deformation function (\eg rotation angle) are perturbed by a quantity upper bounded by the certified radius, $g$'s prediction will remain constant. 
This theoretical result then allowed for certifying parametric image deformations.
% certification of classifiers against various image deformations in~\cite{deformrs, perez20223deformrs}. 

Experimentally, we consider three common image corruptions: (1)~additive pixel-intensity perturbations, (2)~pixel-location translations, and (3)~pixel-location rotations. 
Perturbing pixel intensities corresponds to RS' original formulation from Eq.~\eqref{eq:soft-smoothing}, in which the input $x$ is additively perturbed by noise.
The other two corruptions require DeformRS' formulation from Eq.~\eqref{eq:parametric-smooth-classifier}.
For this formulation, let $\Omega^{\mathbb{Z}}$ be the grid of an $N\times M$-sized image, where $p_{n,m} = (n,m)\in \Omega^{\mathbb{Z}}$ represents a pixel location.
If this grid is additively perturbed by a vector field $\nu_\phi(p_{n,m}) = (u_{n,m}, v_{n,m}) \in \mathbb R^2$, modeling translations and rotations corresponds to defining $\nu_\phi(p_{n,m})$ as follows:
% ~\cite{deformrs}:
% \egornote{What is meant by the "field" here?}
% \juannote{@Egor is the field terminology clearer now?}
% \egornote{Ideally I would put a small example picture (like on from DeformRS paper, as it clarifies instantly), but maybe making a reference is enough for someone who wants more details or needs additional explanation.}

\paragraph{Translation.} 
Image translation requires two parameters $\phi = \{t_u,\, t_v\}$, \ie the translation magnitude along each axis, namely $\nu_\phi(p_{n,m}) = (t_u, t_v) ~~\forall p$ $~\forall n,m$ as per Eq.~\eqref{eq:parametric-smooth-classifier}. 

\paragraph{Rotation.} 
2D rotation is parameterized by the rotation angle, thus $\phi = \{\beta\}$, where $u_{n,m} = n (\cos(\beta)-1) - m \sin(\beta)$ and $\nu_{n,m} = n \sin(\beta) + m (\cos(\beta)-1)$.

% \egornote{Do we need to specify the Threat Model model here, or its fully clear from the description of the defences?}

% \juannote{I think the formulation should be clearer now. Can someone make a pass and fix as desired?}

% \paragraph{Affine.} Our formulation for the certification of the parametric family of deformations is general and covers all affine vector fields as special cases. In particular, affine vector fields are parameterized by 6 parameters, namely $\phi = \{a,b,c,d,e,f\}$, where $u_{n,m} = an + bm + e$ and $v_{n,m} = cn + dm + f$. Note that this class naturally covers composite deformations, such as scaling and translation jointly.

\begin{figure*}
    \centering
    \includegraphics[trim={0 10.4cm 0 0},clip , width=0.8\textwidth]{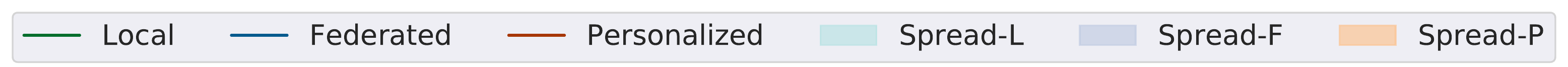}
    \includegraphics[width=0.3\textwidth]{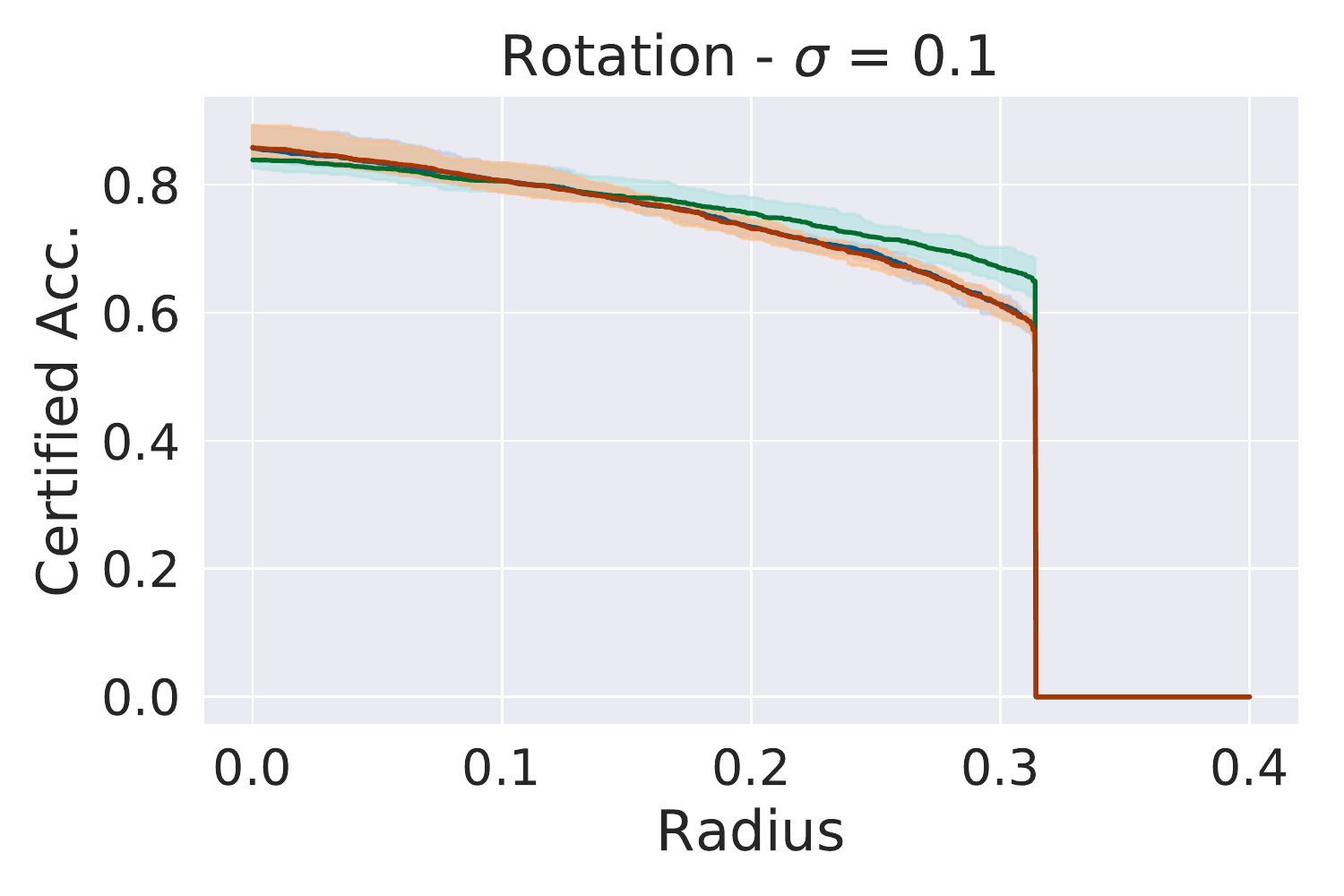}
    \includegraphics[width=0.3\textwidth]{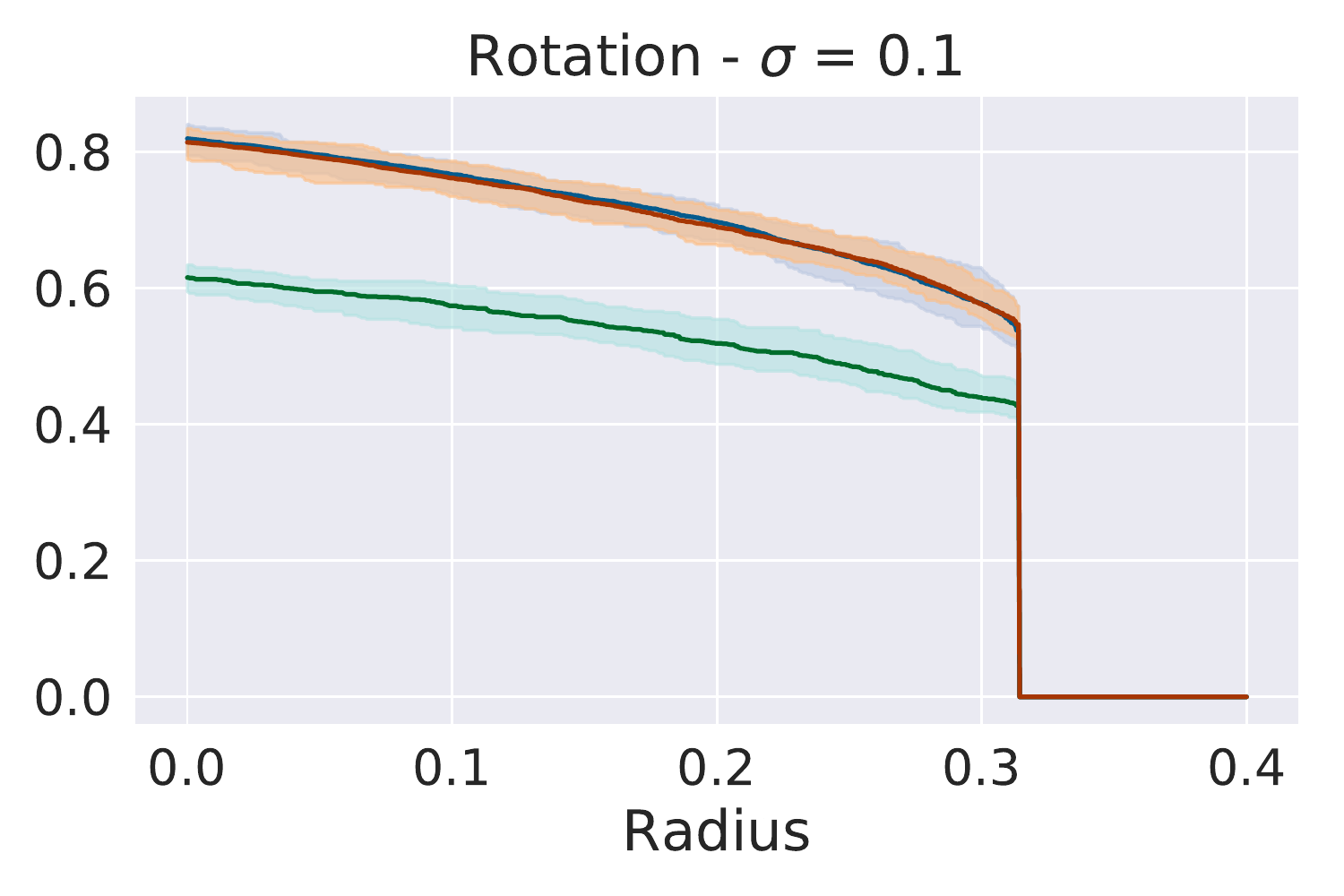}
    \includegraphics[width=0.3\textwidth]{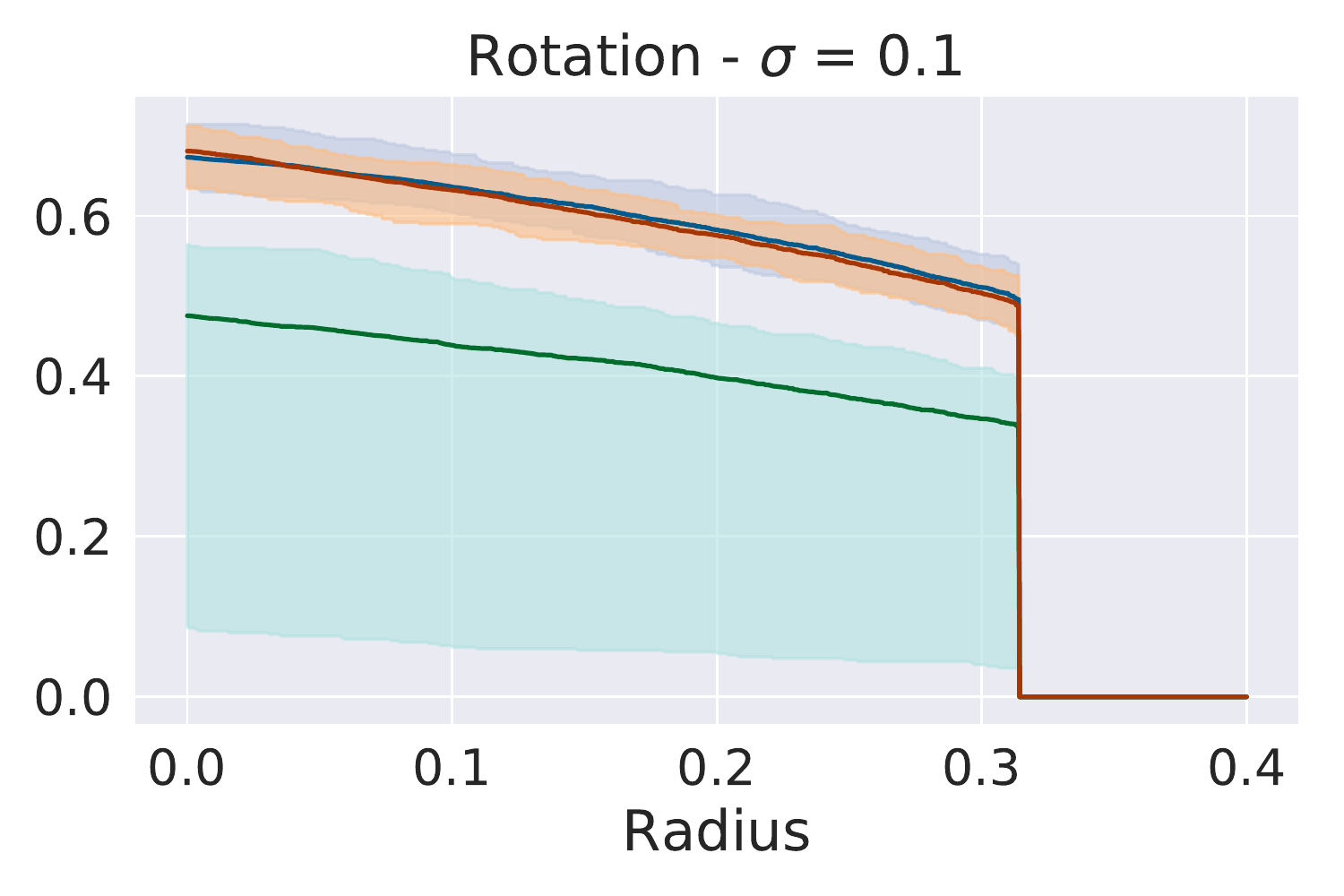} \\
    \includegraphics[width=0.3\textwidth]{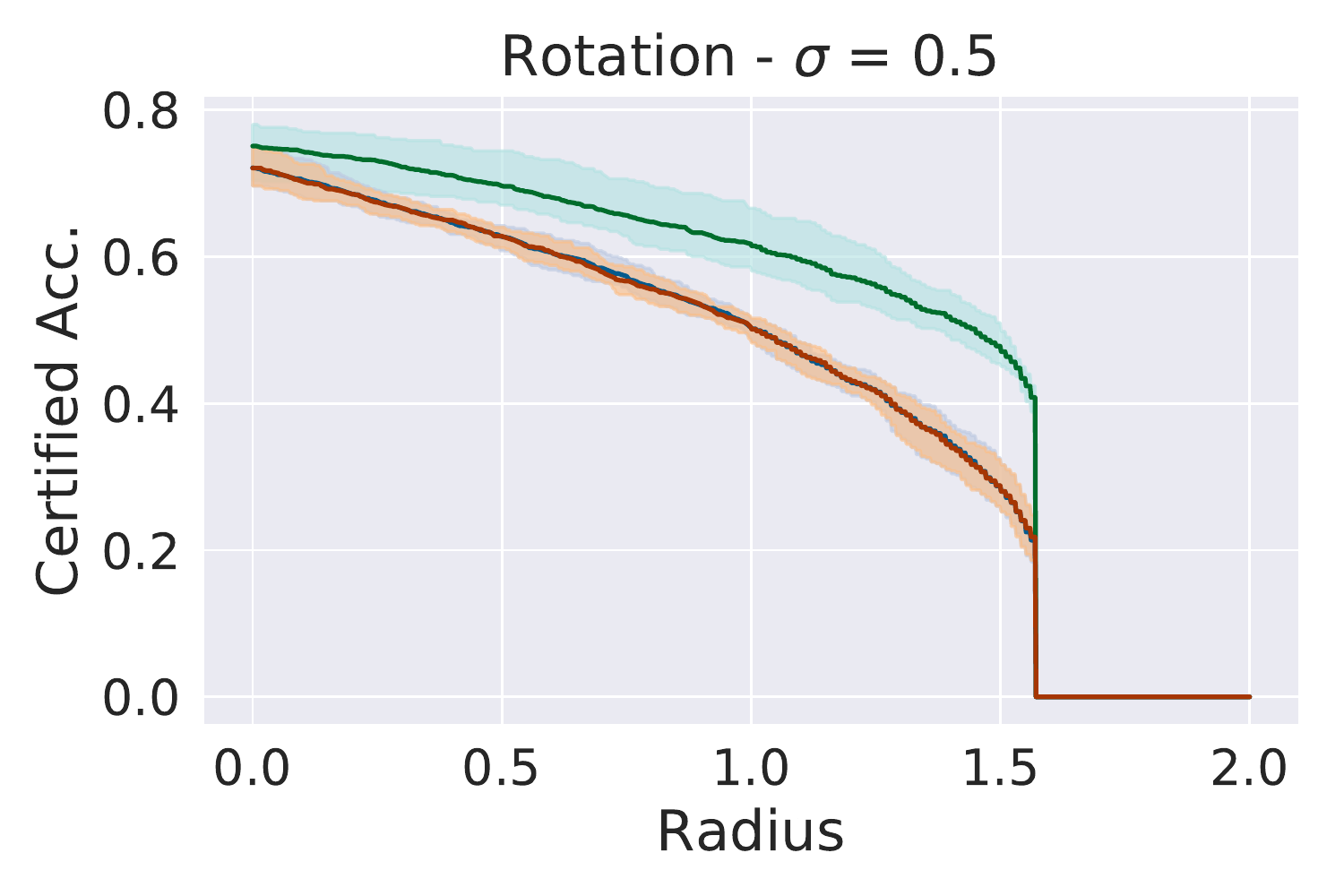}
    \includegraphics[width=0.3\textwidth]{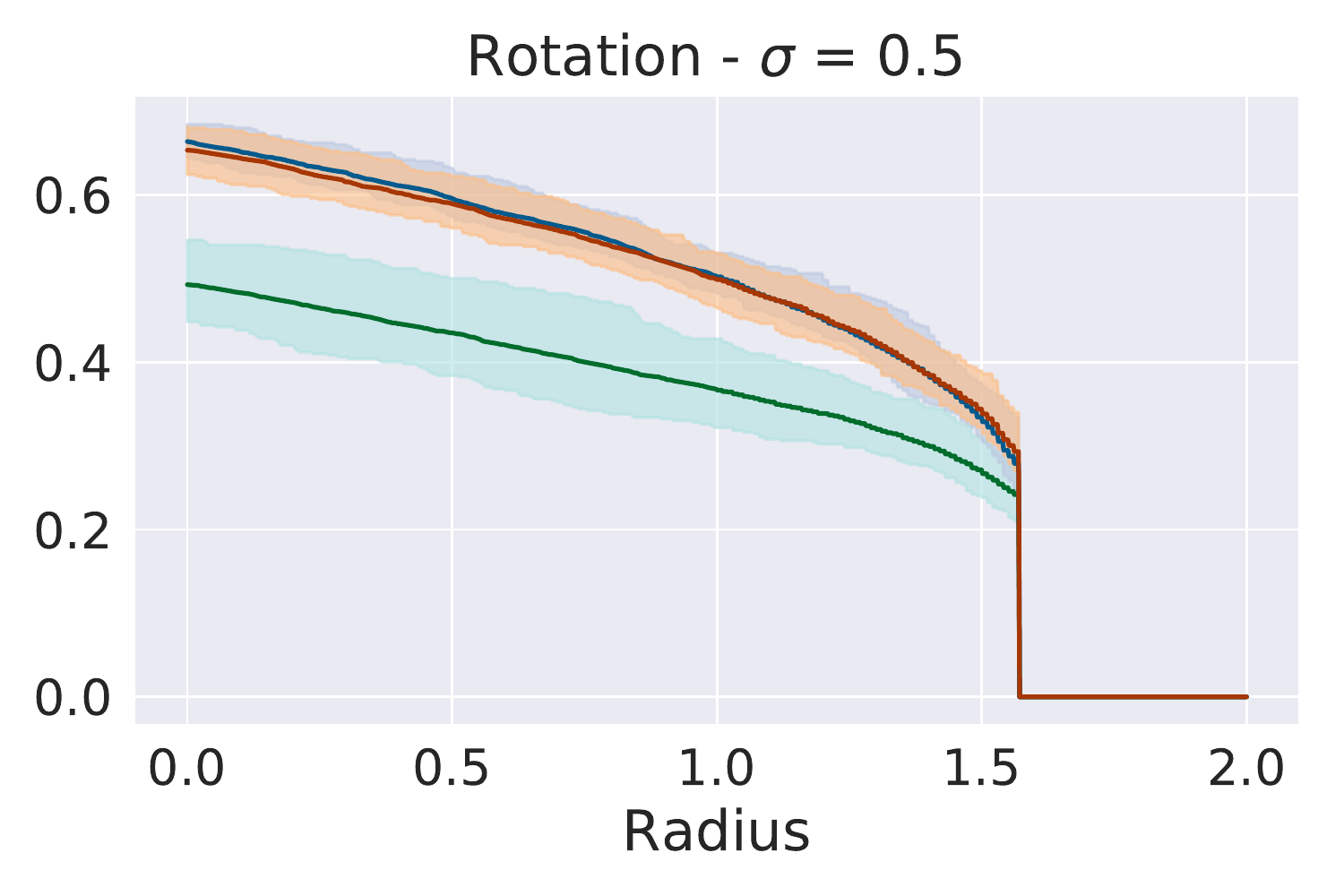}
    \includegraphics[width=0.3\textwidth]{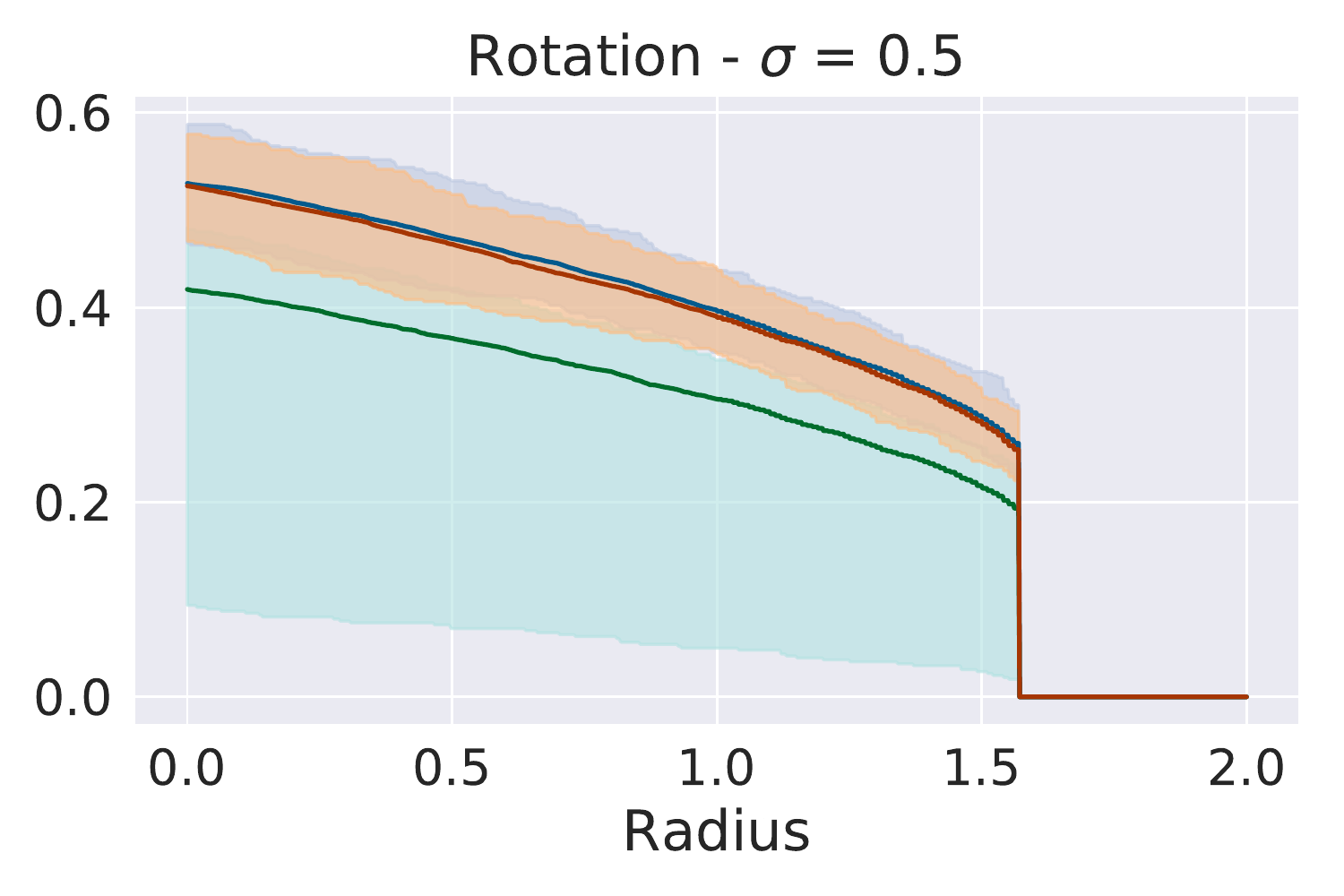}
    \vspace{-0.4cm}
    \caption{\textbf{Certified Accuracy against rotation.} 
    We plot the certified accuracy curves for rotation deformation with varying $\sigma \in\{0.1, 0.5\}$ in the top and bottom rows respectively. 
    We vary the number of clients on which we divide the dataset. 
    Left: 4 clients, middle: 10 clients, right: 20 clients. 
    As the number of clients increases, performance of local training worsens.}\vspace{-0.5cm}
    \label{fig:rotation-ablating-num-clients}
\end{figure*}
\subsection{Specialization to Federated Training}
Earlier works proposed various training frameworks to enhance the robustness of smooth classifiers. 
The schemes studied by these frameworks ranged from simple data augmentation~\cite{cohen2019certified} to more sophisticated techniques, including adversarial training~\cite{salman2019provably} and regularization~\cite{Zhai2020MACER:}. 
In this work, we robustify the classifier via simple and nondemanding data augmentation.
We select this approach since our main aim is to analyze the effect of federated training on certified robustness in isolation of additional and more sophisticated training schemes. 
In addition, this setup allows compatibility with other FL-specific constraints and technologies, such as differential privacy and secure aggregation, which can be essential for real-world deployment~\cite{kairouz2021advances}.

We analyze certified robustness against three different input transformations. 
In particular, we consider pixel perturbations~\cite{cohen2019certified}, and image rotation and translation~\cite{deformrs}. 
Note that the first setting perturbs pixel \textit{intensities}, whereas the other two perturb pixel \textit{locations}. 
We measure the certified radius for each sample according to Eq.~\eqref{eq:certification_radius} for input perturbations, and leverage Corollary~\ref{cor:parametric-certification} for measuring certification radii for rotation and translation. 
Following~\citet{deformrs}, we smooth with a uniform distribution for rotation, and Gaussian smoothing for pixel perturbations and translation.

% \egornote{I do not recommend mentioning "homogeneous and non-overlapping manner" so bluntly here because it kinda contradicts the realistic FL setting.}

As in the Federated Learning setup, we employ the common federated averaging~\cite{Povey2014,fedavg} technique. 
In a nutshell, we consider the case when a dataset is distributed across multiple clients in a non-overlapping manner. 
We assume that all clients share the architecture $f_\theta$, to allow for weight averaging across clients.
Furthermore, we study three different training schemes, where each client is minimizing the following regularized risk~\cite{implicit_personalization}:
\begin{equation}\label{eq:implicit-personalization}
    \min_{\theta_1, \dots, \theta_n} \frac{1}{n}\sum_{i=1}^n \lambda \mathcal L_i(\theta_i) + (1-\lambda)\left\| \theta_i - \bar \theta \right\|_2^2, 
\end{equation}
where $\mathcal L_i$ is a training loss function (\eg cross entropy) evaluated on local data of client $i$, $\theta_i$ represents the $i^\text{th}$ client's model parameters, $\lambda\in[0, 1]$ is a personalization parameter, and $\bar \theta = \frac{1}{n}\sum_{i=1}^n\theta_i$ is the average model. 

\paragraph{Local Training.} 
Each client trains solely on their own local data, without communication with other clients. 
At test time, each client employs their own local model. 
This setting is equivalent to minimizing the risk in Eq.~\eqref{eq:implicit-personalization} with $\lambda=1$.

\paragraph{Federated Training.} 
All clients are collaborating in building one global model. 
This is achieved by iteratively performing (1) client-side model updates with optimization on local data (\ie solving Eq.~\eqref{eq:implicit-personalization} with $\lambda=1$), and (2) communication of client models to the server, which averages all models and broadcasts the result back to the clients (\ie solving Eq.~\eqref{eq:implicit-personalization} with $\lambda=0$). 
% Each client then updates the received model with few local steps for the next communication round. 
This process is repeated until convergence or for a fixed number of communication rounds. 
At test time, all clients employ the global model sent from the server at the last communication round.

\paragraph{Personalized Training.} 
Since \citet{yu2020salvaging} exposed how the basic FL formulation with $\lambda = 1$ and $\theta_1 = \dots = \theta_n$ degrades the performance of some clients, we explore a personalization approach.
Namely, upon federated training convergence, each client fine-tunes the global model communicated in the last round with a few optimization steps on their local data (\ie solving Eq.~\eqref{eq:implicit-personalization} with $\lambda=1$). 
Despite its simplicity, this approach was shown to perform well in various tasks~\cite{yu2020salvaging}.

\begin{figure*}
    \centering
    \includegraphics[trim={0 7.4cm 0 0},clip , width=0.6\textwidth]{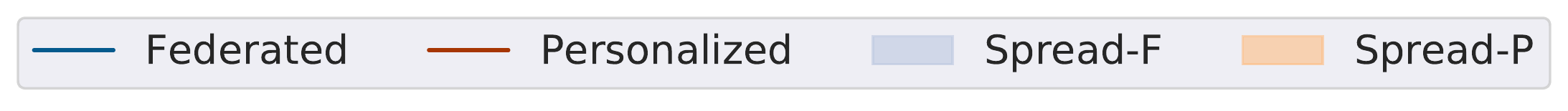} \\
    \includegraphics[width=0.3\textwidth]{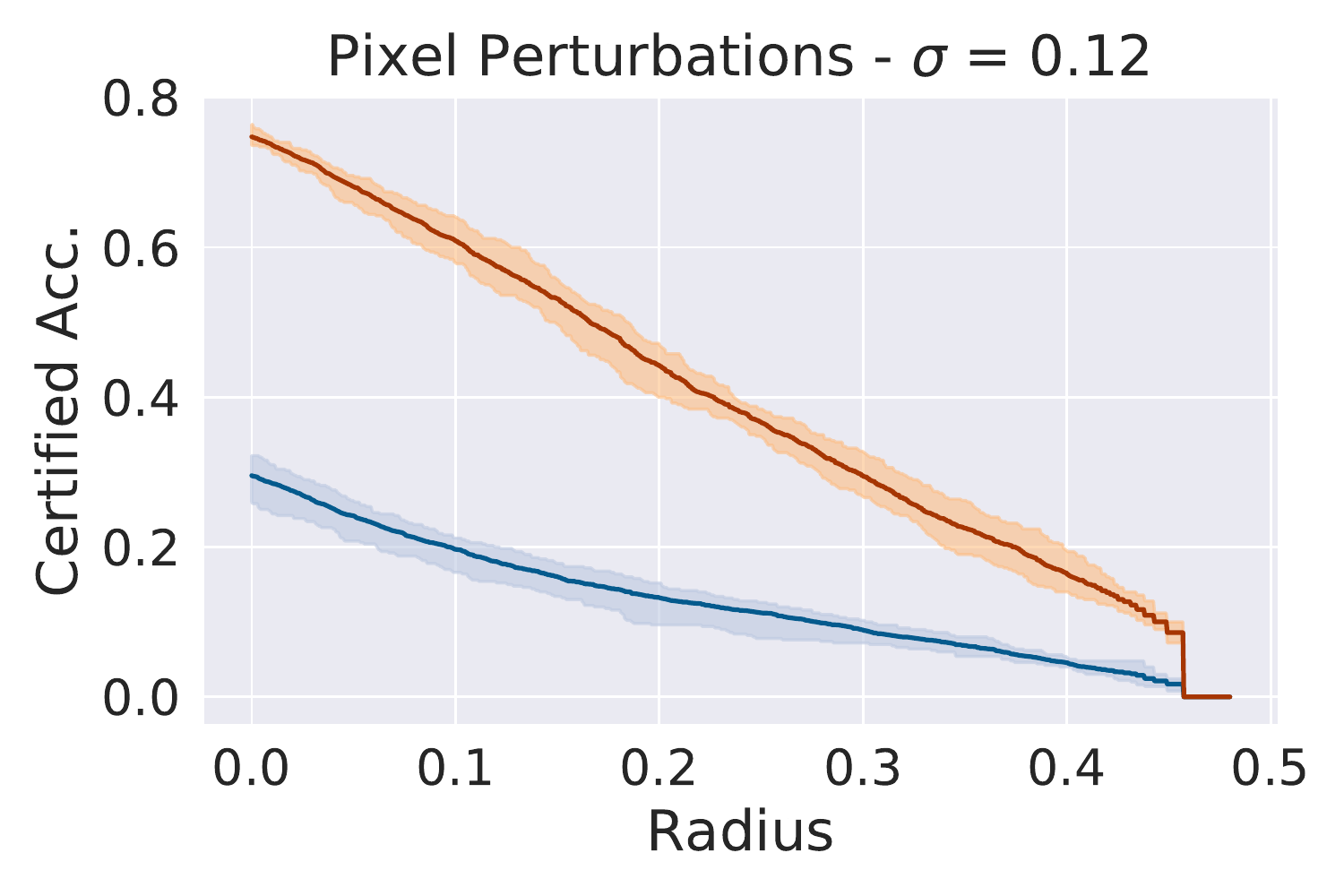}
    \includegraphics[width=0.3\textwidth]{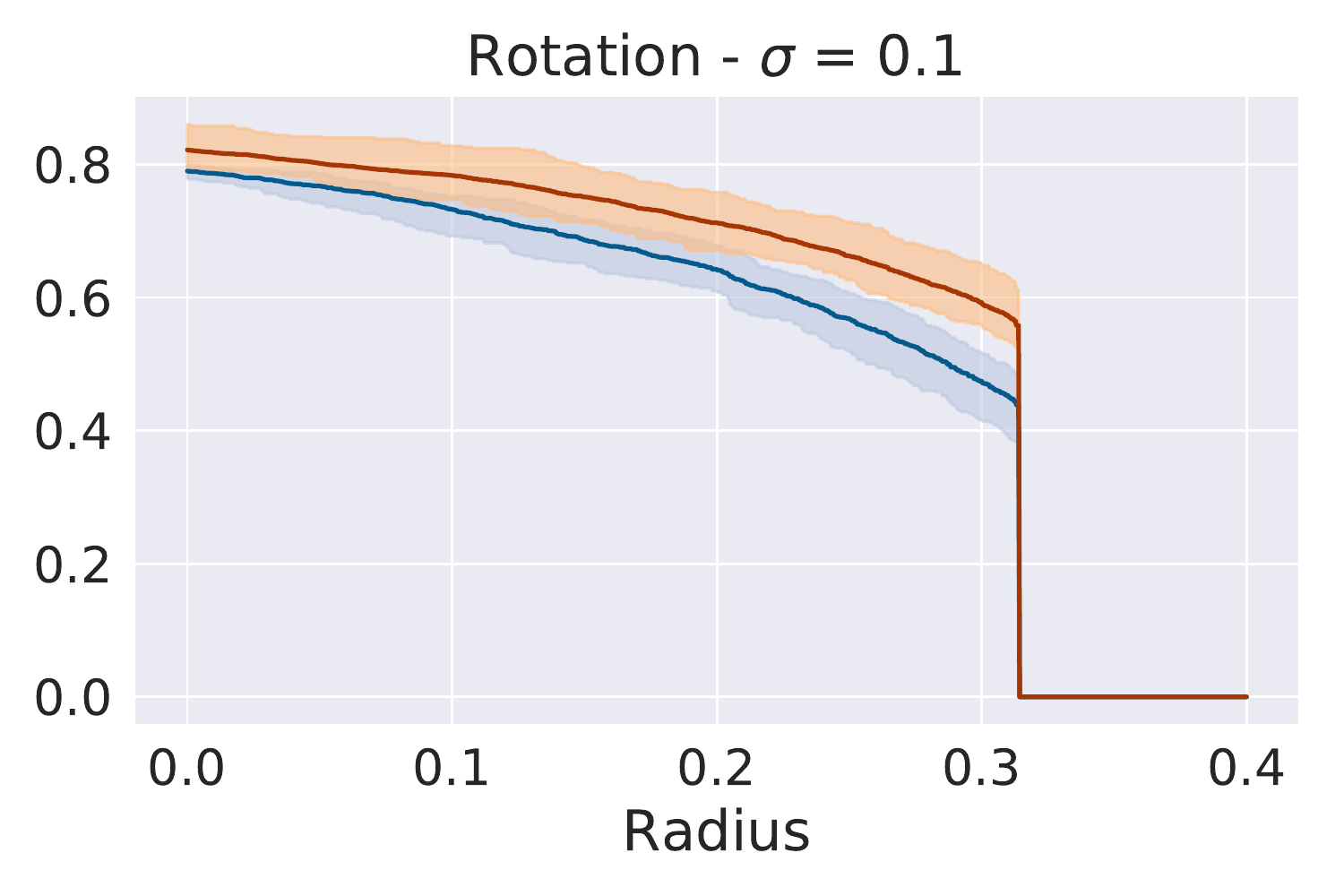}
    \includegraphics[width=0.3\textwidth]{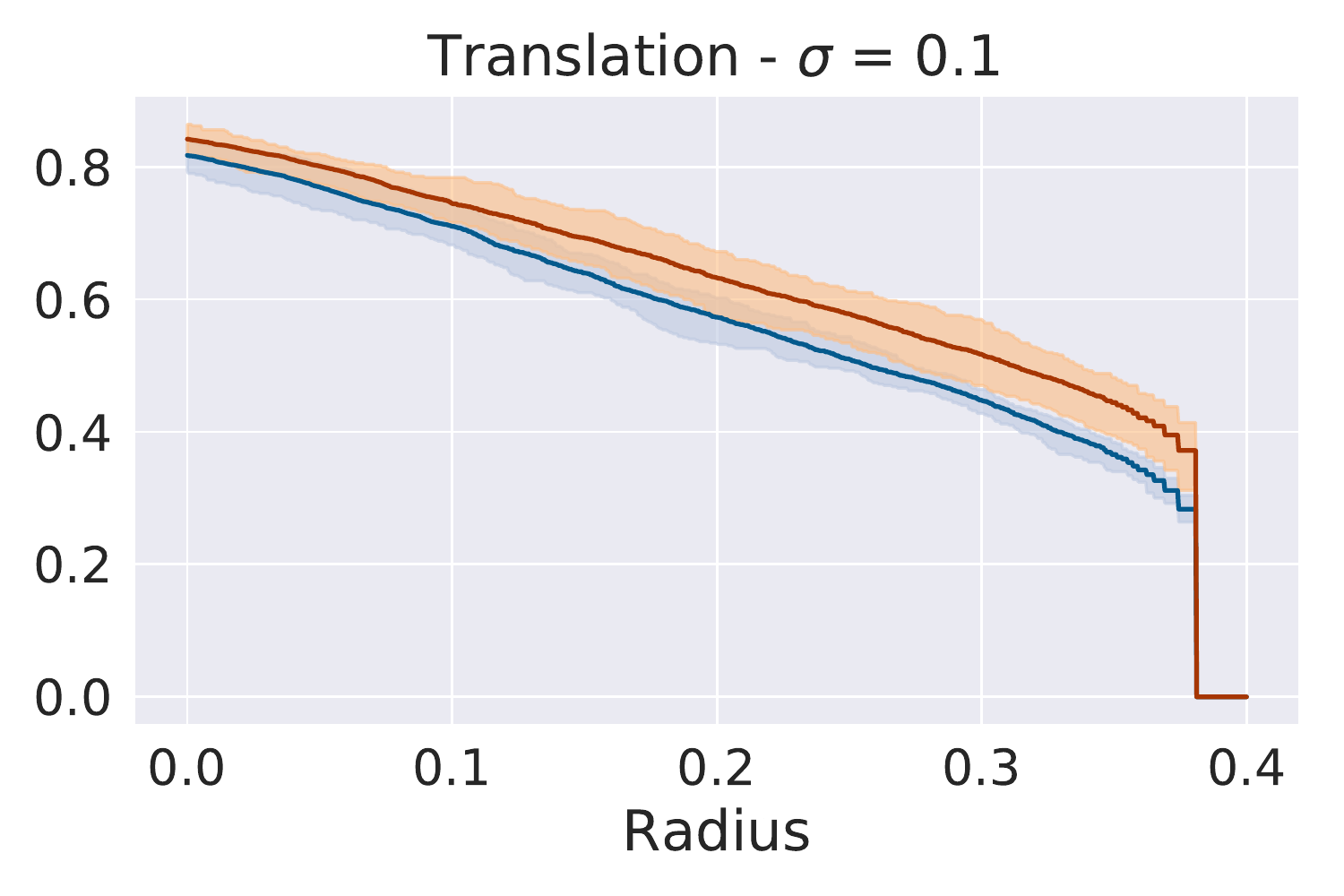} \\
    \includegraphics[width=0.3\textwidth]{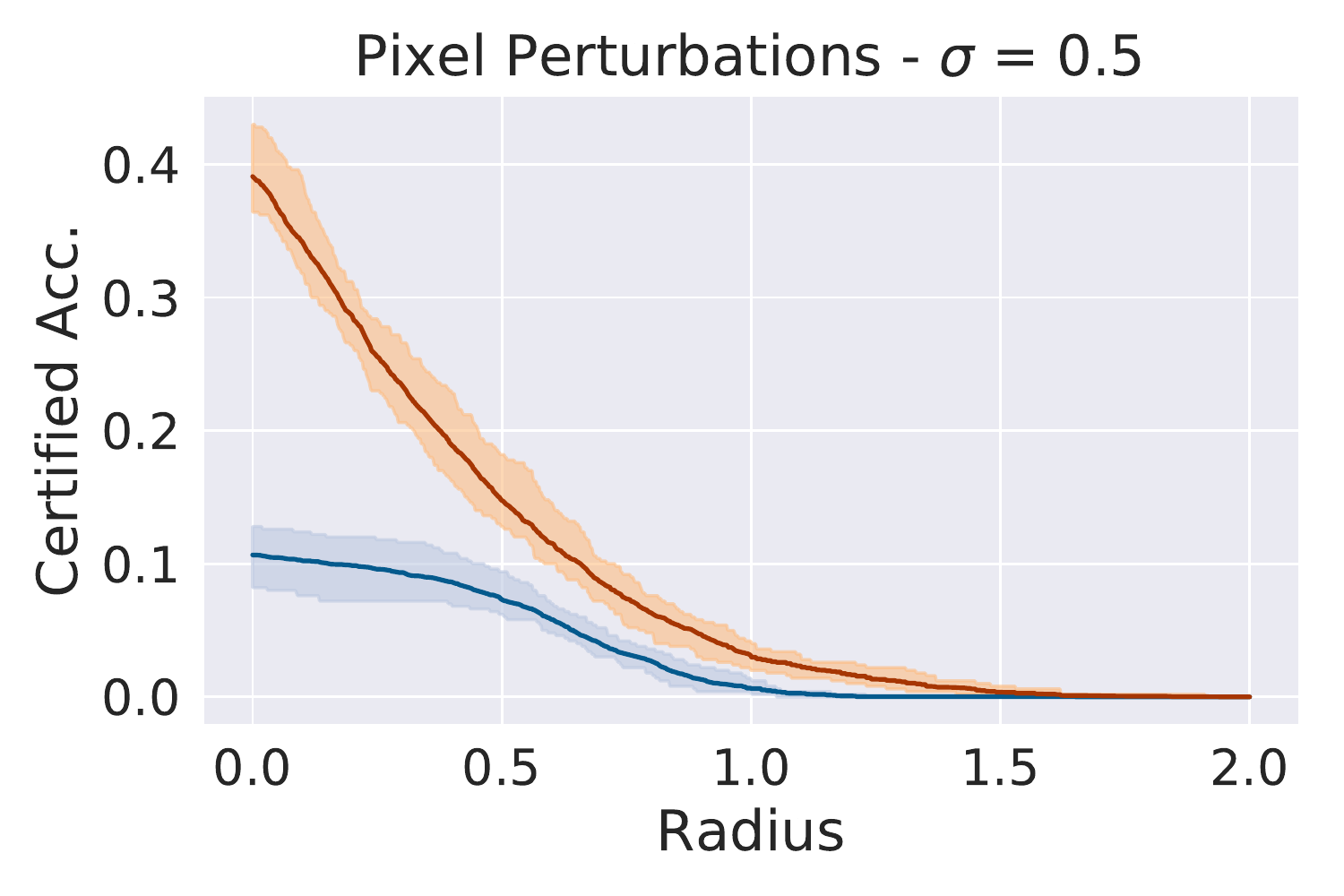}
    \includegraphics[width=0.3\textwidth]{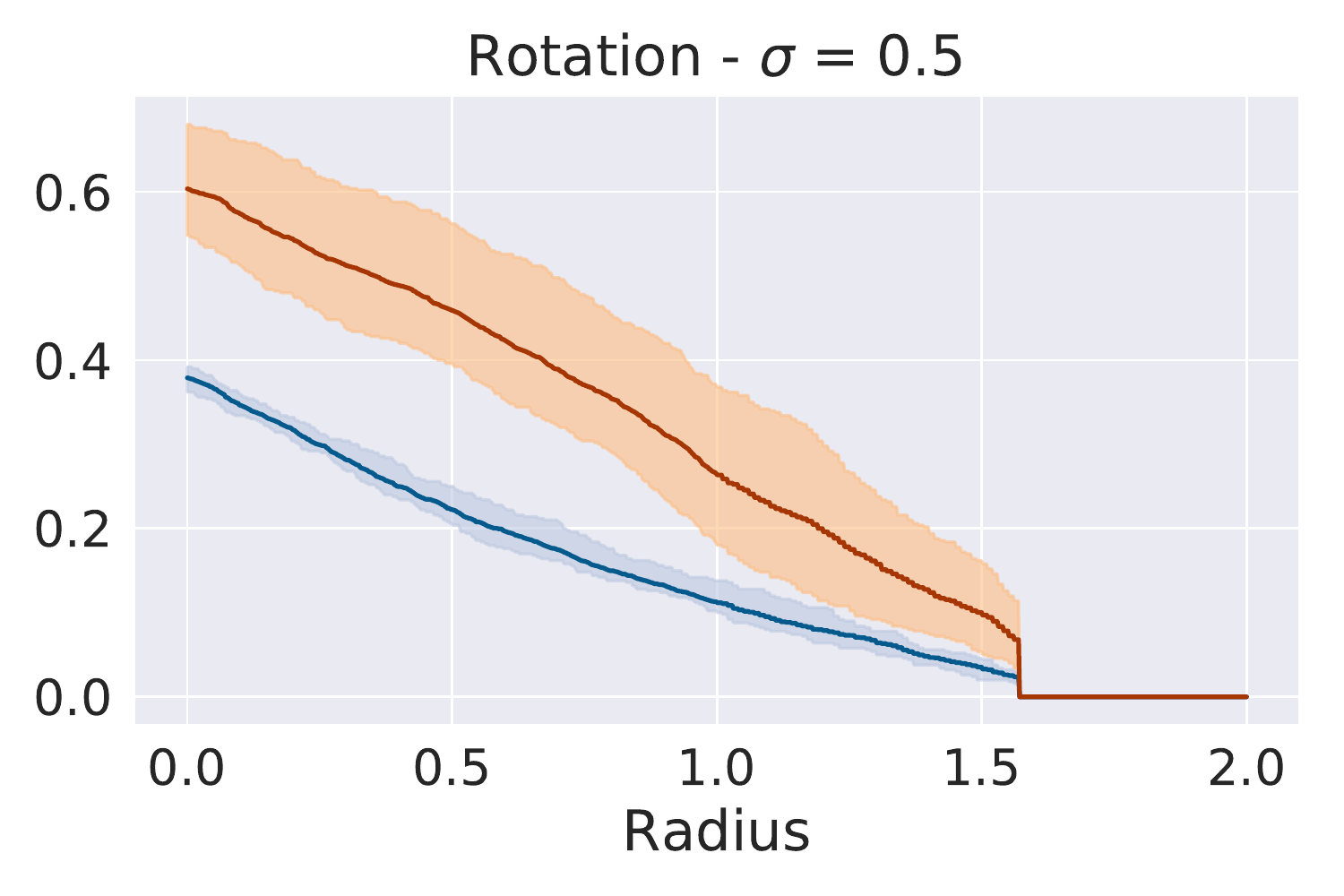}
    \includegraphics[width=0.3\textwidth]{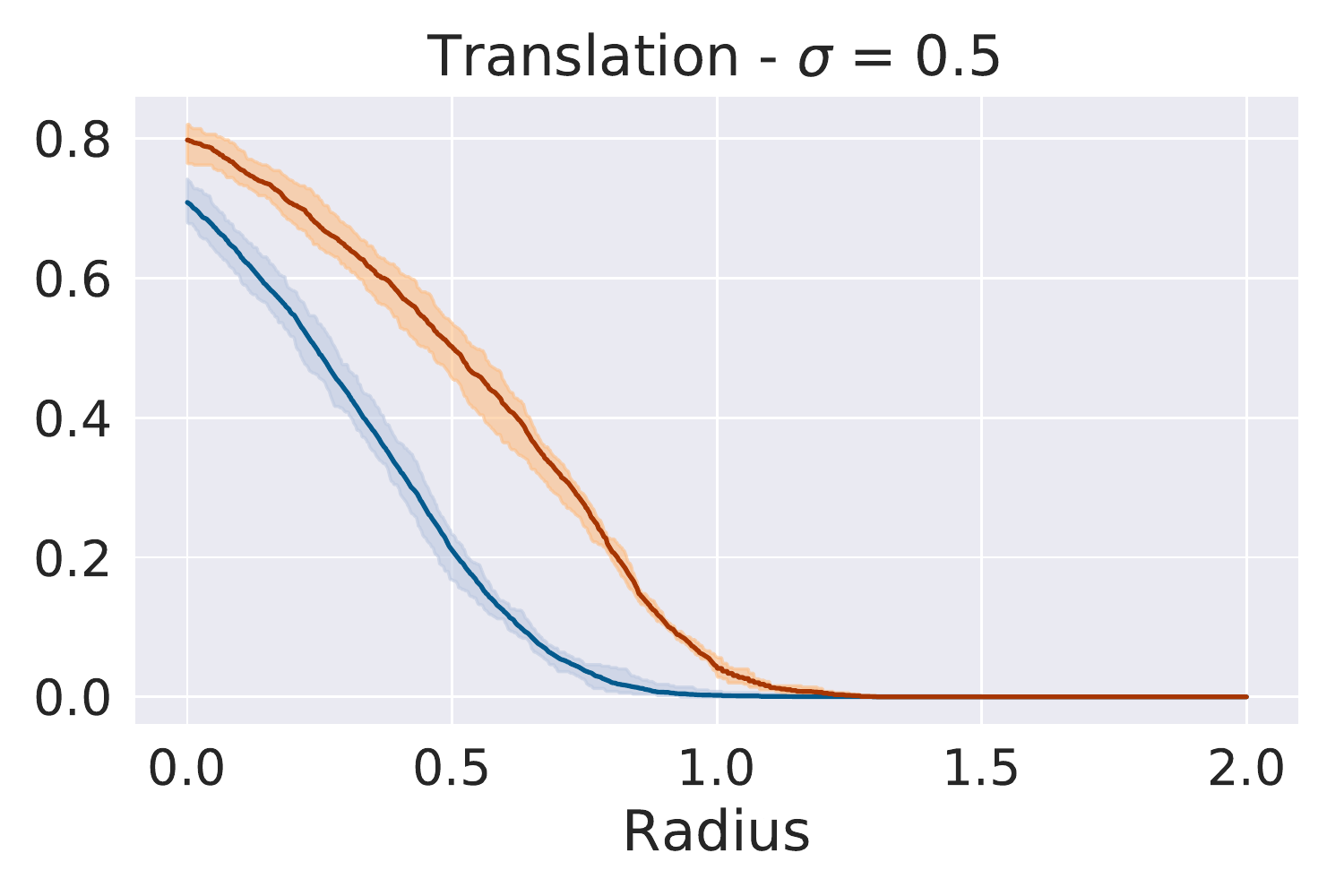}\vspace{-0.4cm}   
    \caption{\textbf{Personalization effect on certified robustness.} 
    Certified accuracy curves against pixel perturbations (left), rotation (middle), and translation (right). We use $\sigma \in \{0.12, 0.5\}$ for pixel perturbations and $\sigma \in\{0.1, 0.5\}$ for rotation and translation.}
    \vspace{-0.3cm}
    \label{fig:nominal-pretraining}
\end{figure*}

\begin{figure*}[t]
    \centering
    \includegraphics[width=0.235\textwidth]{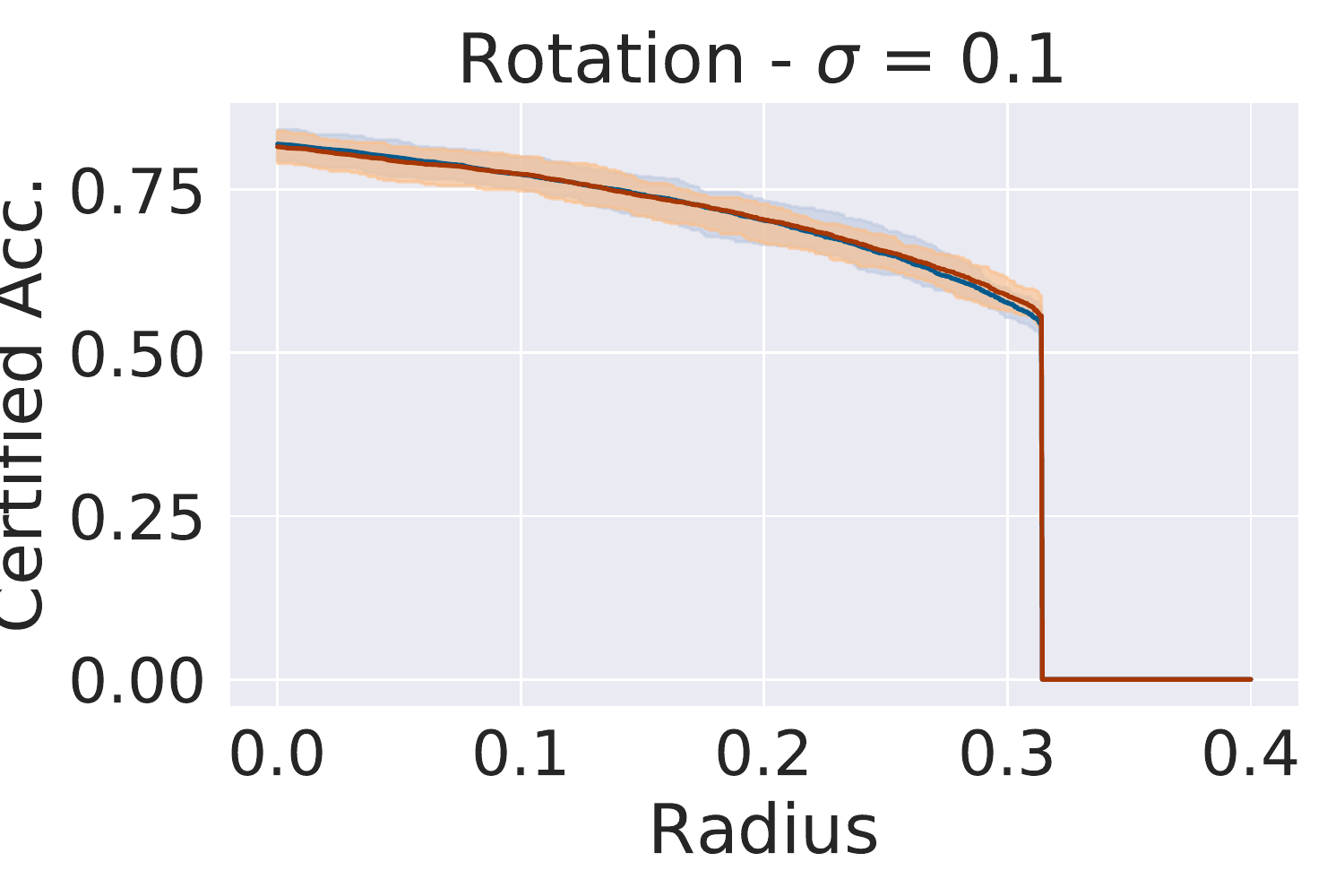}
    \includegraphics[width=0.235\textwidth]{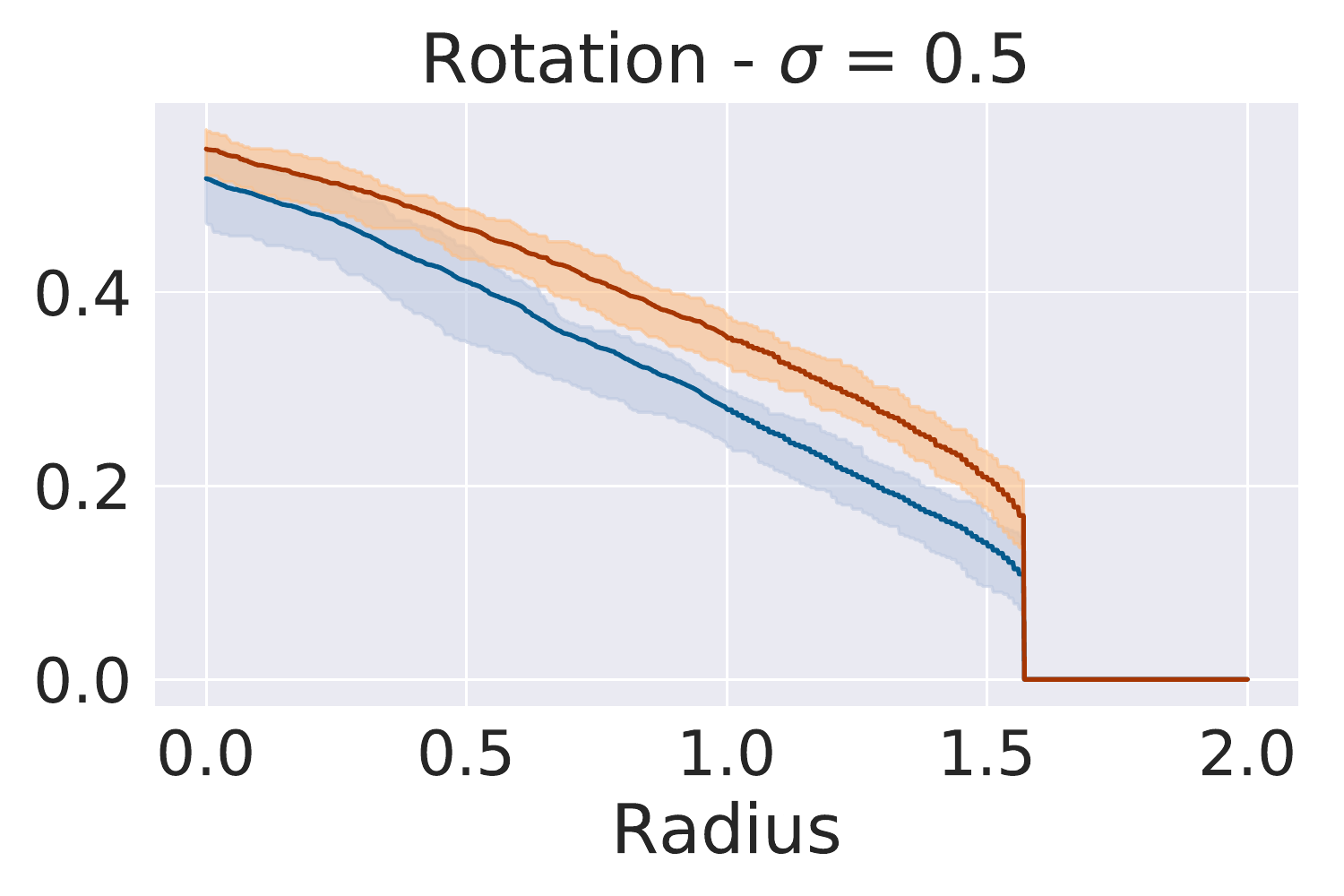}
    \includegraphics[width=0.235\textwidth]{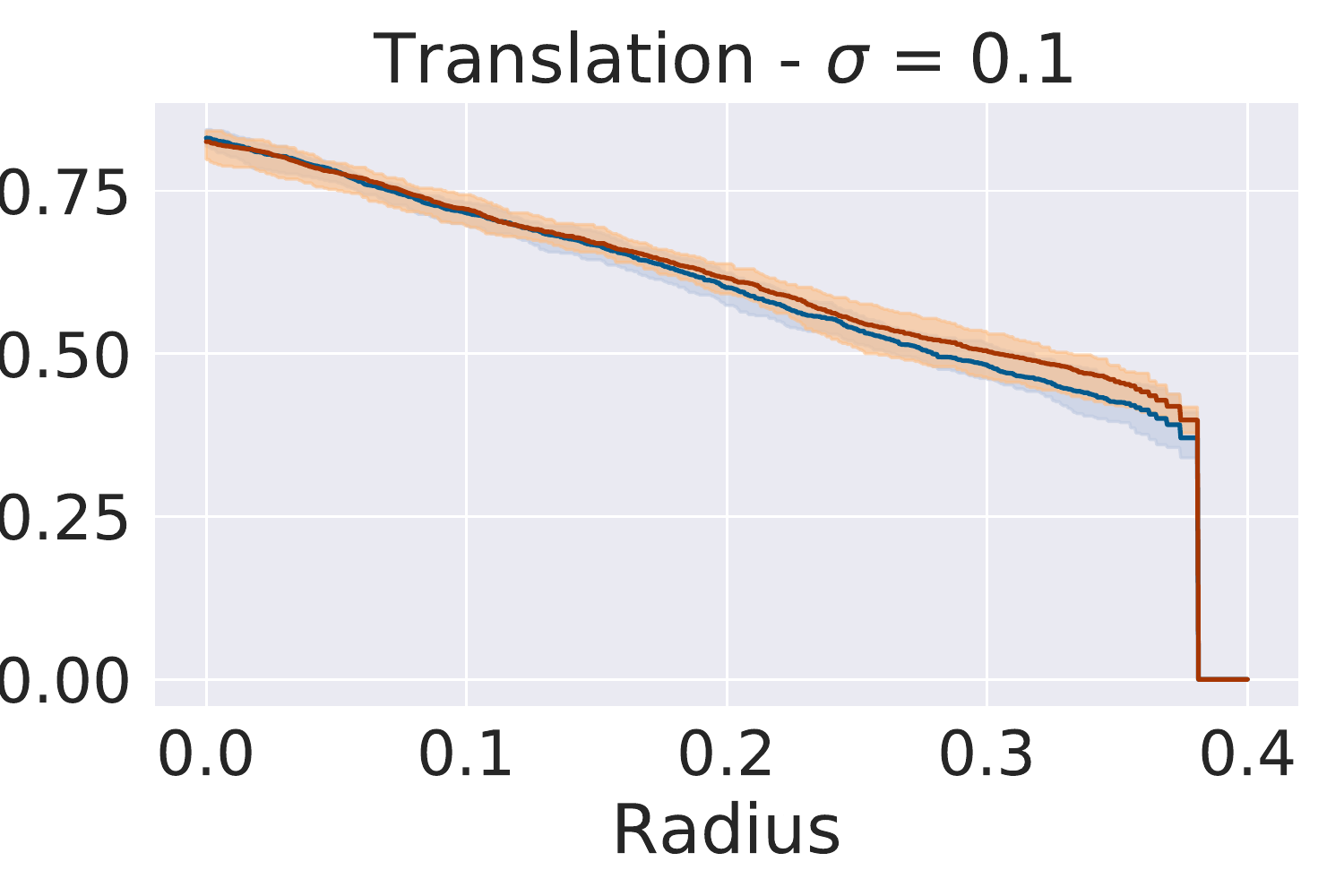} 
    \includegraphics[width=0.235\textwidth]{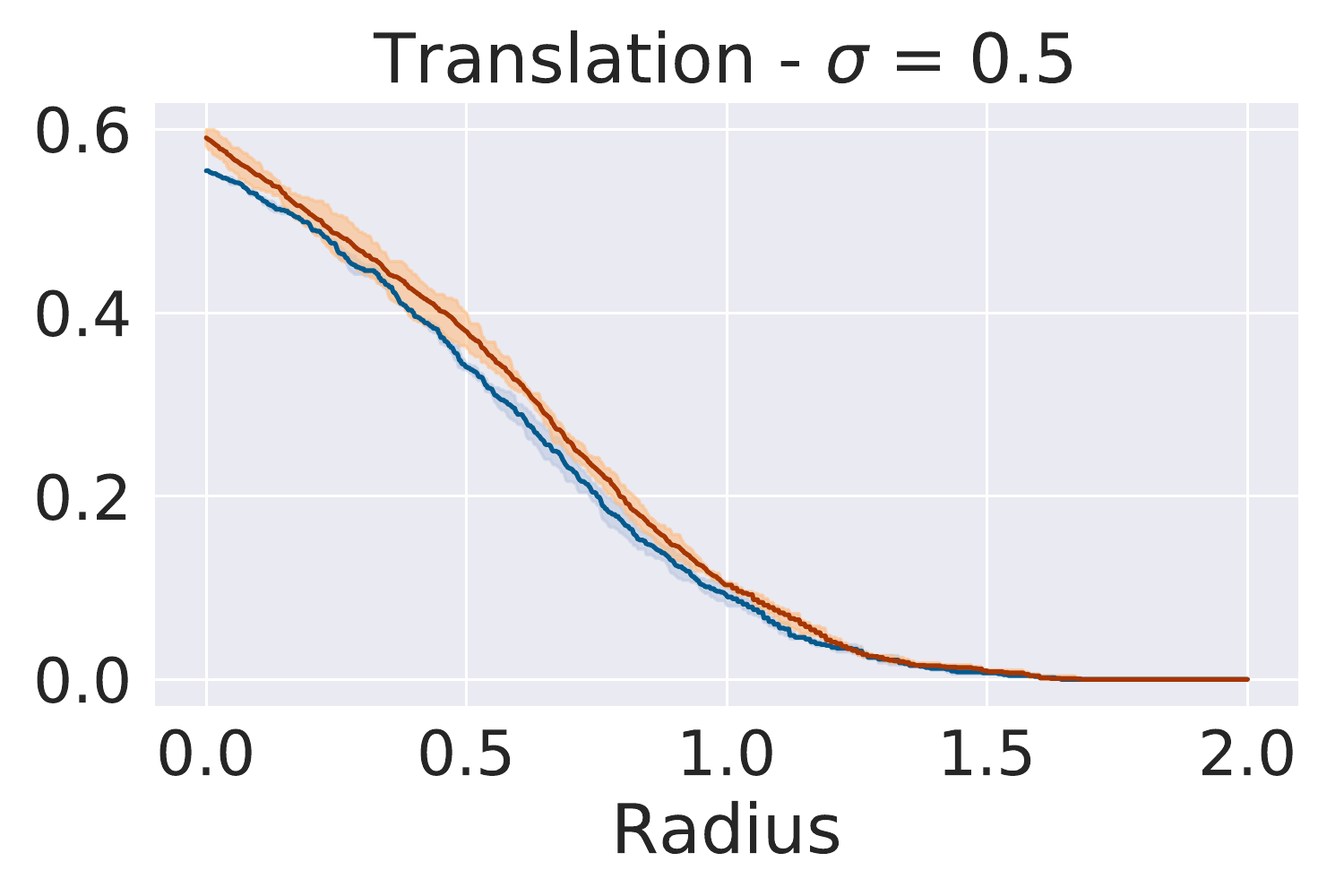} 
    \vspace{-0.4cm}   
    \caption{\textbf{Certified Robustness against Rotation starting from Affine. } % Effect of Personalization} 
    Left: personalization with rotation augmentation.
    % with $\sigma\in\{0.1, 0.5\}$. 
    Right: translation augmentation.
    % with $\sigma\in\{0.1, 0.5\}$.
    The color convention follows that of Figure~\ref{fig:nominal-pretraining}.}
    \vspace{-0.5cm}
    \label{fig:affine-pretraining}
\end{figure*}
\paragraph{Mixture of models.} 
% We address personalization with a mixture of global and local models~\cite{implicit_personalization, kulkarni2020survey}.
We also consider a more sophisticated \textit{personalized} ERM setting via a mixture of global and local models~\cite{implicit_personalization}, seeking an explicit trade-off between the global model and the local models for $\lambda\in(0,1)$.
We choose this approach because the formulation in Eq.~\eqref{eq:implicit-personalization} provably benefits from local training (as it is done in local SGD and FedAvg) in terms of communication efficiency. 
In fact, \citet{hanzely2020lower} showed that methods with local steps are optimal in this case.
In contrast, for standard ERM, local methods are known to be worse than their non-local counterparts in the heterogeneous data regime~\cite{woodworth2020minibatch}.
% (see \cite{ProxSkip} for a recent breakthrough in this area).
% We choose this approach due to recent advances~\cite{hanzely2020lower} in broader theoretical understanding of its properties for communication, oracle calls and local gradients (given by development of lower bounds and optimal algorithms).

Our implementation for solving problem~\eqref{eq:implicit-personalization} for every client $i$ requires local gradient descent steps (with step size $\gamma$) and a mixing of current global and local models through:
\begin{equation}\label{eq:mixture_update}
    \theta_i \leftarrow 
    \theta_i - \gamma \left(
    \lambda \nabla \mathcal{L}_i(\theta_i) + 2(1-\lambda) (\theta_i - \bar{\theta})\right),
\end{equation}
alternating with communication to the server for computing the current average model $\bar \theta$. 
At test time, each client uses its corresponding personalized model $\theta_i$.
% \juannote{@Egor, could you please list the reference for arguing that our personalization formulation is theoretically sound? I can introduce a sentence or two here based on the paper.}
% \egornote{It was first introduced in this paper https://arxiv.org/abs/2002.05516 and a follow-up (with lower bounds and optimal methods for convex case) https://arxiv.org/abs/2010.02372}
% \juannote{@Egor, I have written this.
% Should we add/remove/modify anything?}
% https://arxiv.org/abs/2002.10619
% https://ieeexplore.ieee.org/abstract/document/9210355

While the effect of federated training and personalization in improving model performance is widely explored in the literature, its impact on the trained model's certified robustness remains unclear.
Thus, in our experiments, we empirically study the following questions: 
\textbf{(i)} Is local training sufficient to build robust models? 
\textbf{(ii)} Does federated training improve the certified robustness of the trained model? 
\textbf{(iii)} Is personalization beneficial for improving the certified robustness of local models?
\textbf{(iv)} How would the mixture of models when setting $\lambda \in (0, 1)$ interact with the resultant certified accuracy of each local model?
%%%%%%%%%%%%%%%%%%%%%%%%%%%%%%%%%%%%%%%%%%%%%%%%%%%%%%%%%%%%%%%%%%%%%%%%%%%
\section{Experiments}

\begin{figure*}[t]
    \centering
    \includegraphics[trim={0 10.4cm 0 0},clip , width=0.8\textwidth]{sections/images/common_legend.pdf}
    \includegraphics[width=0.3\textwidth]{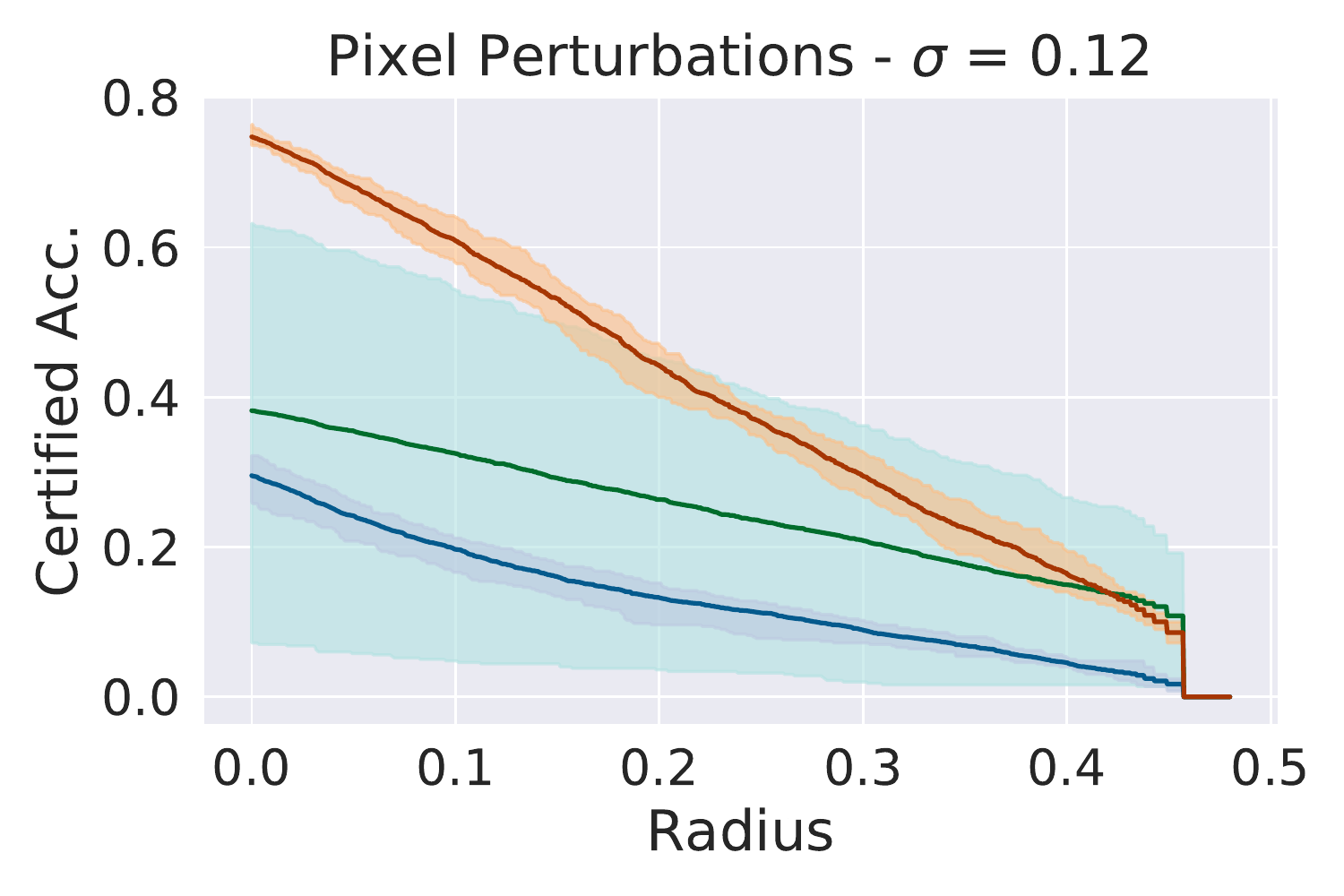}
    \includegraphics[width=0.3\textwidth]{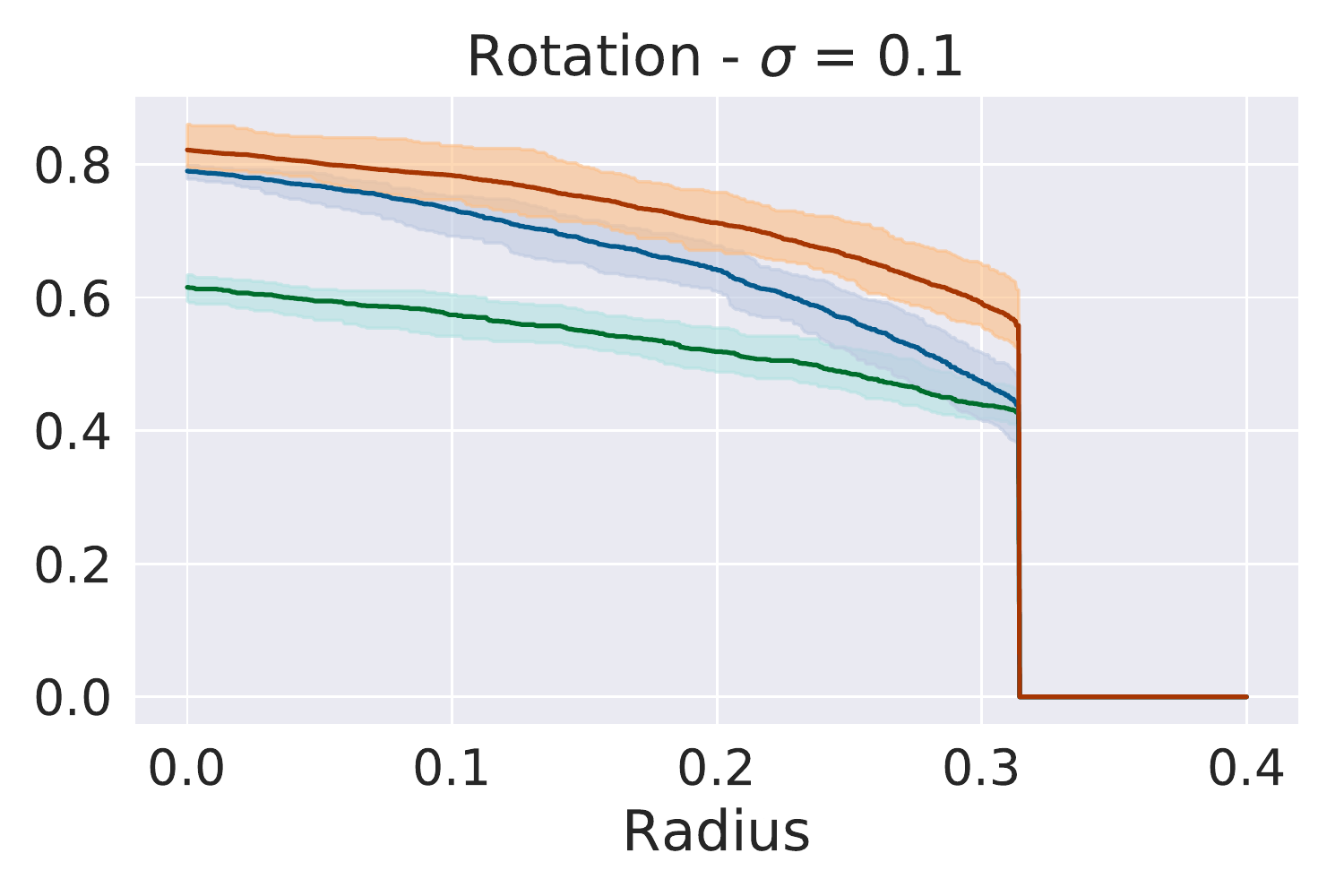}
    \includegraphics[width=0.3\textwidth]{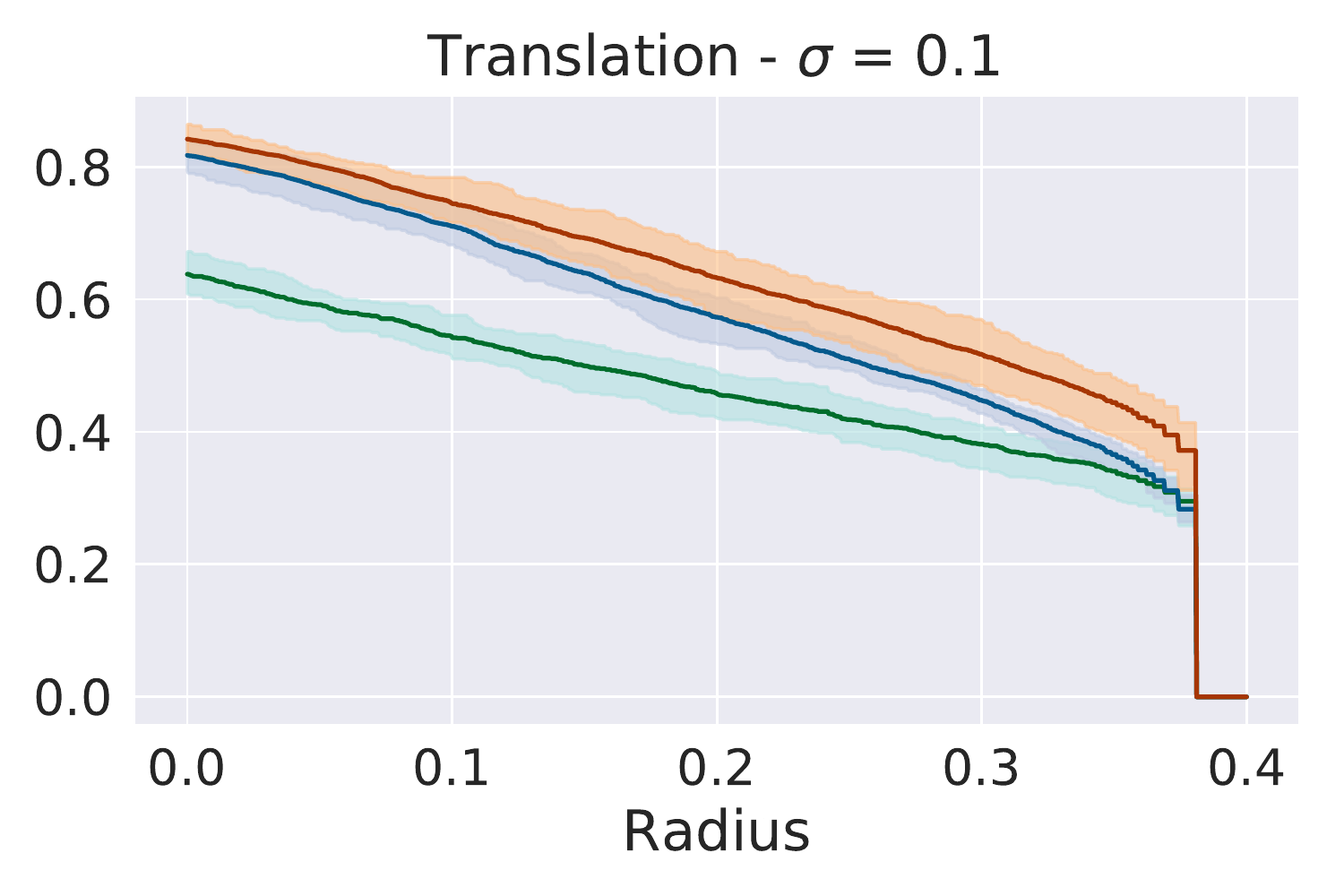} \\
    \includegraphics[width=0.3\textwidth]{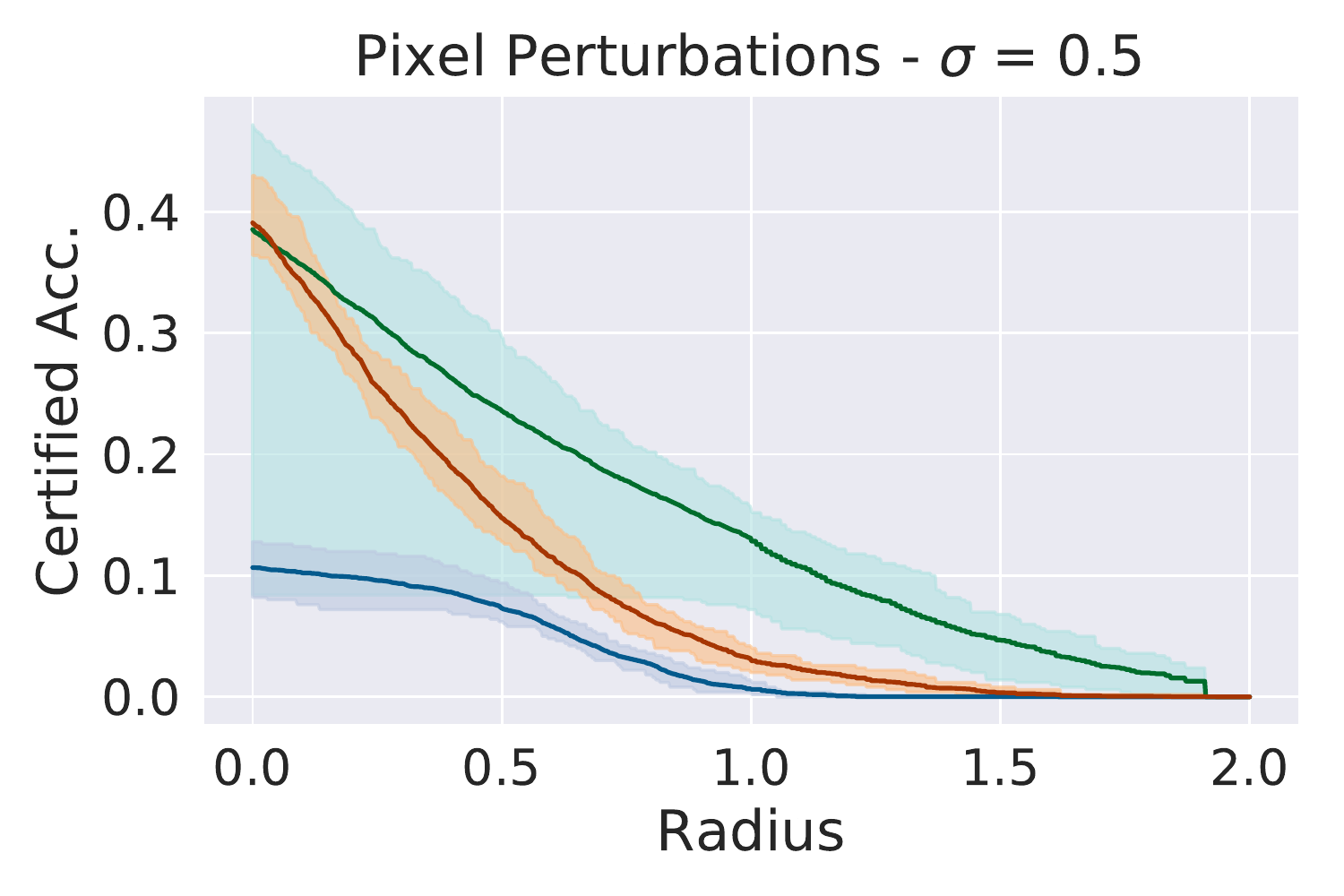}
    \includegraphics[width=0.3\textwidth]{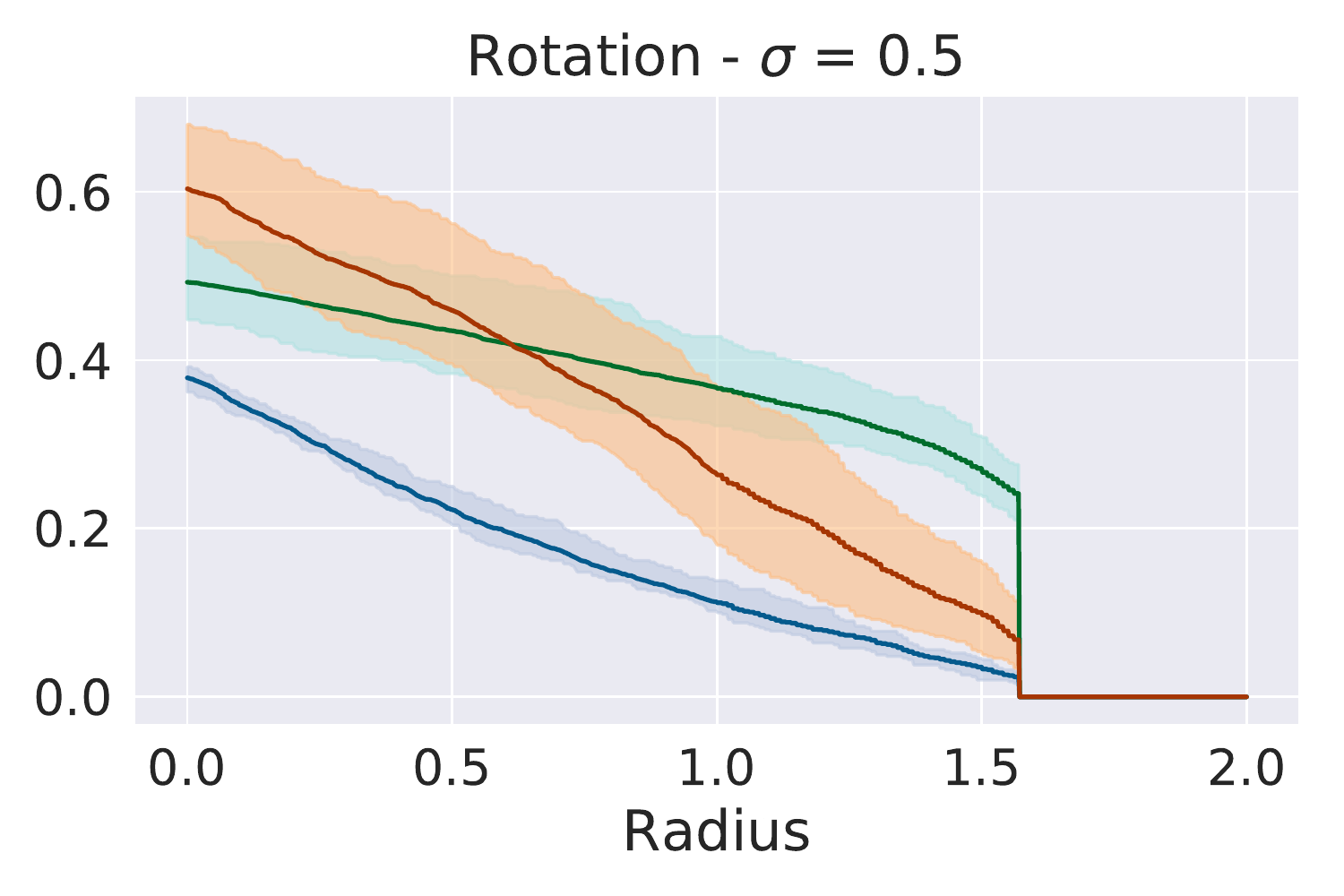}
    \includegraphics[width=0.3\textwidth]{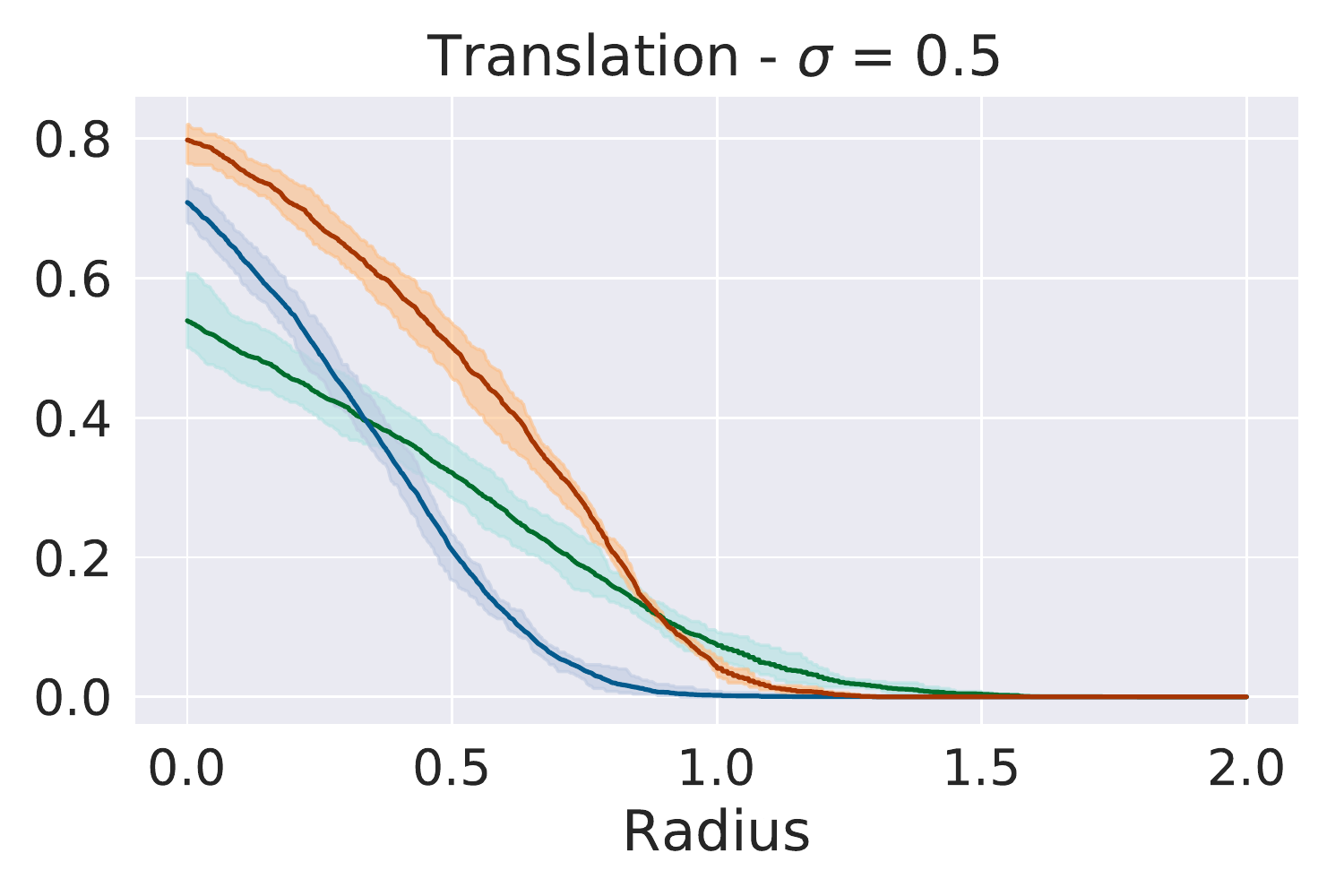}
     \includegraphics[width=0.3\textwidth]{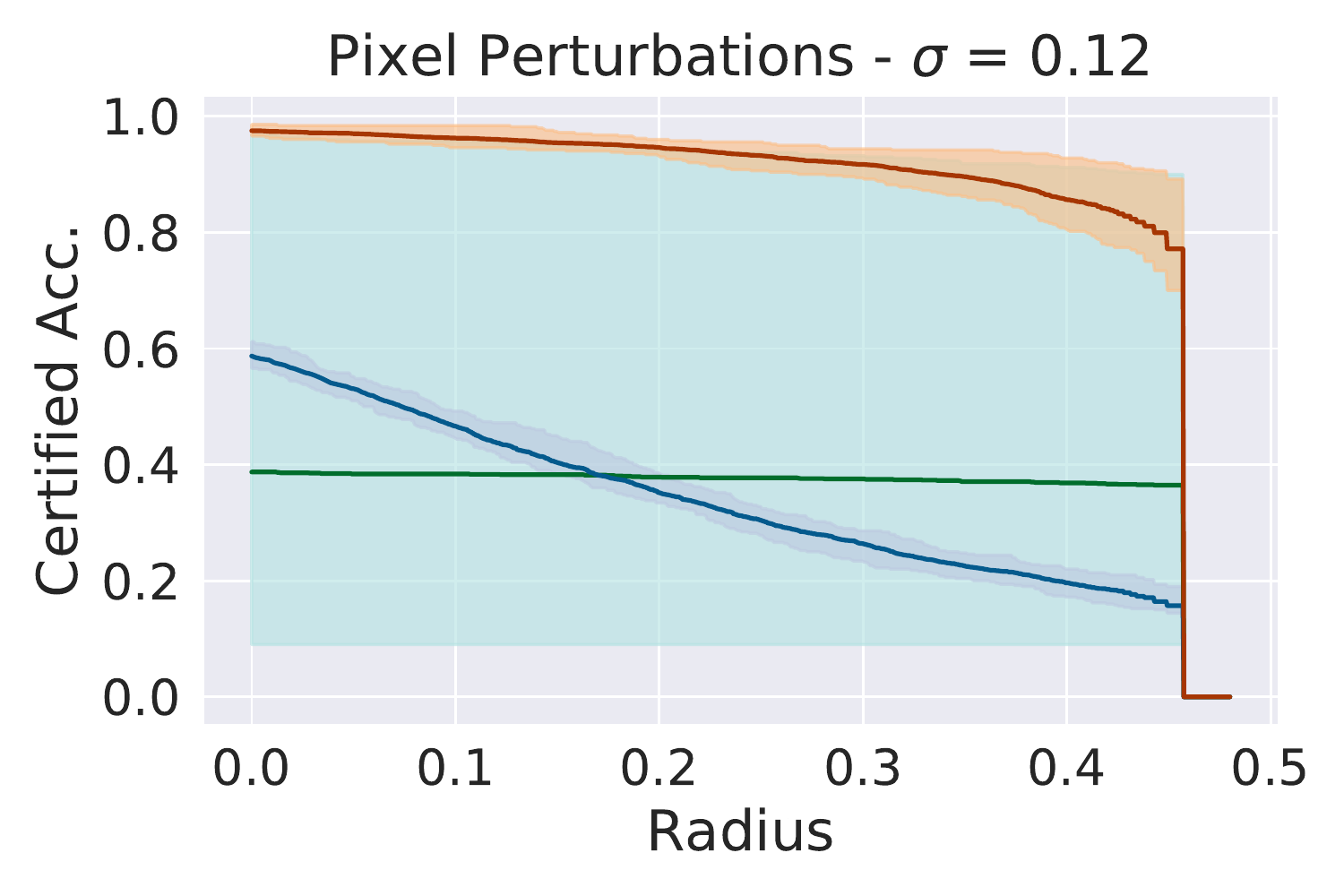}
    \includegraphics[width=0.3\textwidth]{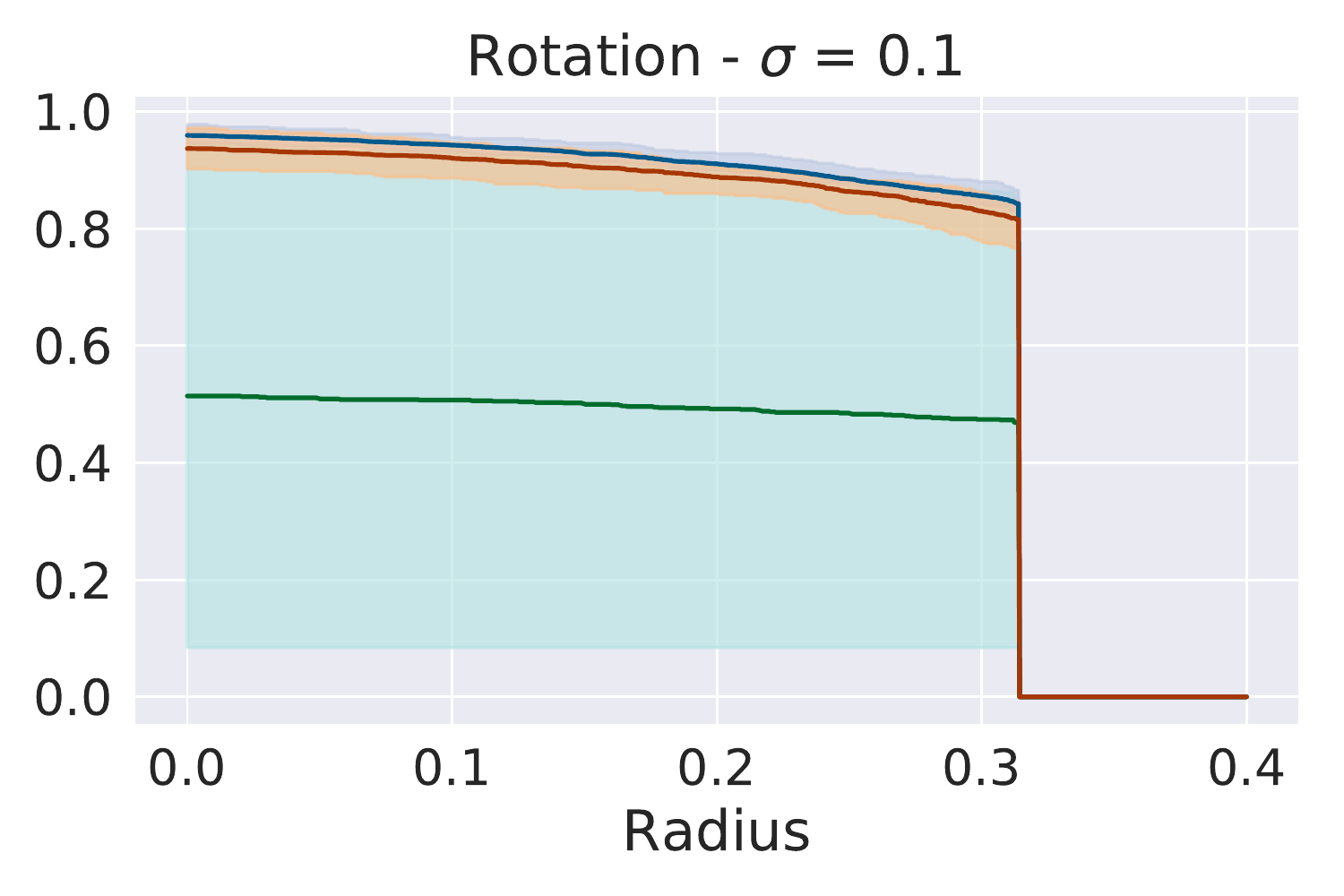}
    \includegraphics[width=0.3\textwidth]{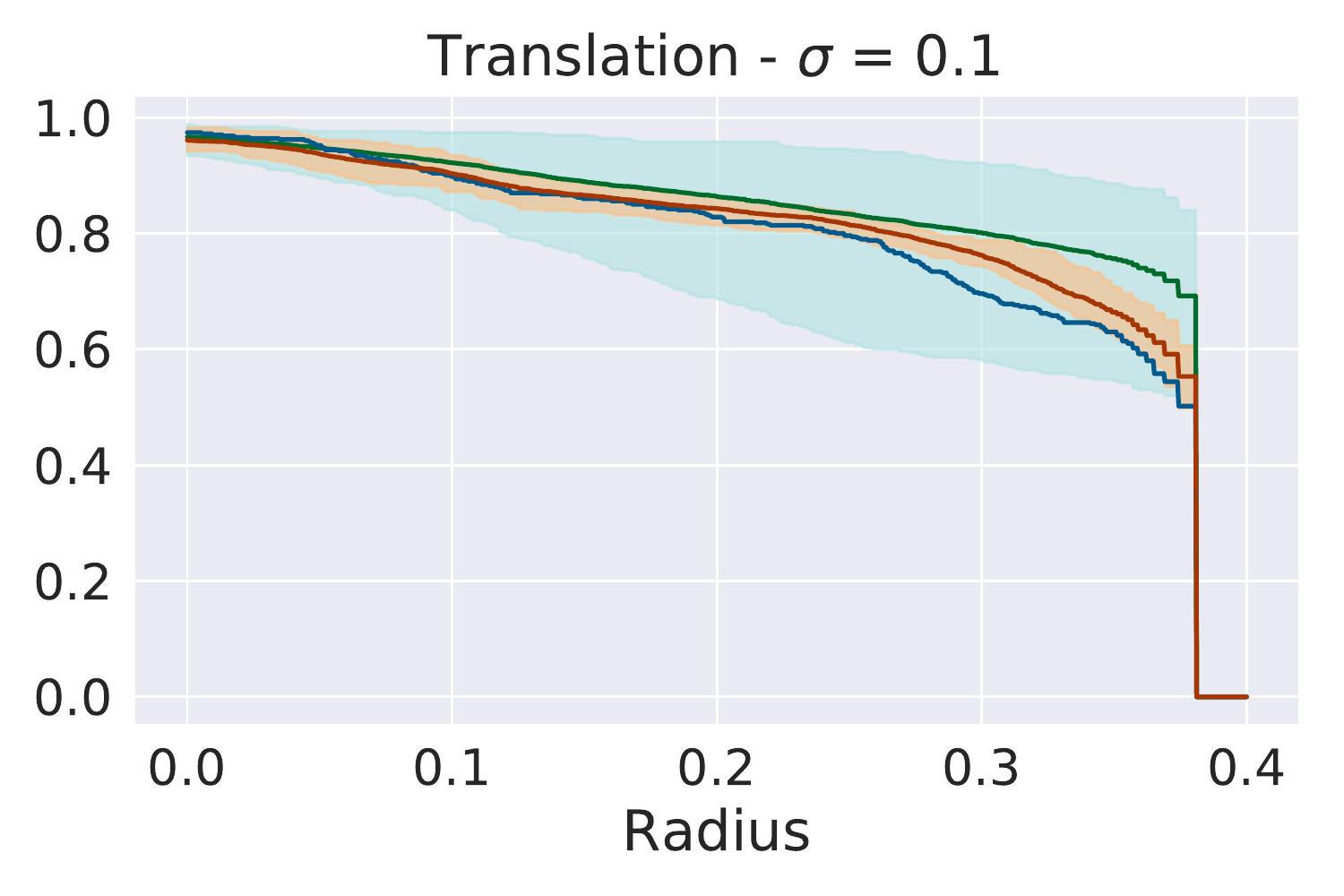}
    \vspace{-0.4cm}
    \caption{\textbf{Comparison between personalization and local training.} 
    We plot the certified accuracy curves against pixel perturbations (left), rotation (middle), and translation (right) with varying $\sigma \in\{0.1, 0.5\}$ for rotation and translation an $\sigma \in \{0.12, 0.5\}$ for pixel perturbations. Last row: Experiments on MNIST dataset.}
    \vspace{-0.5cm}
    \label{fig:nominal-pretraining-all}
\end{figure*}

% \egornote{For me the number of experimental details at the beginning seems overwhelming. That's why I recommend leaving some of them (standard for robustness and not essential for FL) to the Appendix.}
% \motinote{I think it is important to have detailed description since our experiments are the main contribution.}
\textbf{Training Setup.}
We fix ResNet18~\cite{resnet} as our architecture and train on the CIFAR10~\cite{cifars} and MNIST~\cite{lecun2010mnist} datasets. 
We split the training and test sets across clients in a homogeneous and non-overlapping fashion.
% , and leave experiments on non-homogeneous splits to the appendix. 
Note that this split forms a best-case condition for local training, allowing compatibility with federated training and its associated data-privacy constraints.
We conduct 90 epochs of training, divided in 45 communication rounds, \ie 2 epochs of local training per communication round. 
We set a batch size to 64, and use a starting learning rate of $0.1$, which is decreased by a factor of 10 every 30 epochs (15 communication rounds). 
Unless specified otherwise, given a transformation we are interested in robustifying against (\eg rotation), we train the model by augmenting the data with such transformation. 
This procedure is equivalent to estimating the output of the smooth classifier in Eq.~\eqref{eq:parametric-smooth-classifier} with one sample. 
We set the 
% standard deviation 
$\sigma\in\{0.1, 0.5\}$ for rotation and translation and $\sigma \in  \{0.12, 0,5\}$ for pixel perturbations, following~\citet{cohen2019certified, deformrs}.
For local models, we train each client on their local data only with the aforementioned setup without any communication.

\textbf{Certification Setup.} 
For certification, we use the public implementation from \citet{cohen2019certified} for pixel-intensity perturbations, and the one from \citet{deformrs} for rotation and translation.
We use $100$ Monte Carlo samples for computing the top prediction $c_A$, and $100$k samples for estimating a lower bound to the prediction probability $p_A$; we further use a failure probability of $\alpha = 0.001$, and bound the runner-up prediction via $p_B = 1-p_A$.
For experiments on deformations, we choose $I_T$ to be a bi-linear interpolation function, following~\citet{deformrs, perez20223deformrs}.
To evaluate each client's certification, we fix a random subset of $500$ local samples.
At last, we plot the average certified accuracy curves across clients, highlighting the minimum and maximum values as the limits of shaded regions. 
% We use ``Local" for performance of \textit{local} models, ``Federated" for the \textit{federated} model, and ``Personalized" for the \textit{personalized} models.
We report ``certified accuracy at radius $R$'', defined as the percentage of test samples that is both, classified correctly, and has a certified radius of at least $R$. 
We report the ablations for all clients in the appendix.
% Moreover, since image dimensions vary across datasets (square images of sizes 28, 32, 224 for MNIST, CIFAR10 and ImageNet, respectively), we normalize all image dimensions to $[-1, 1]\times[-1, 1]$.

\subsection{Is Federated Training Good for Robustness?}

\begin{figure*}
    \centering
    \includegraphics[trim={0 10.4cm 0 0},clip , width=0.9\textwidth]{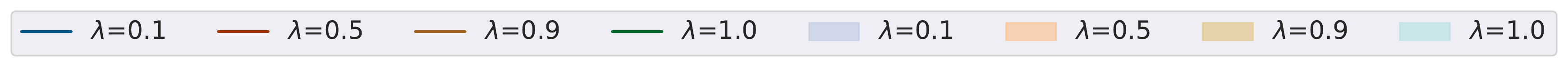}
    \includegraphics[width=0.3\textwidth]{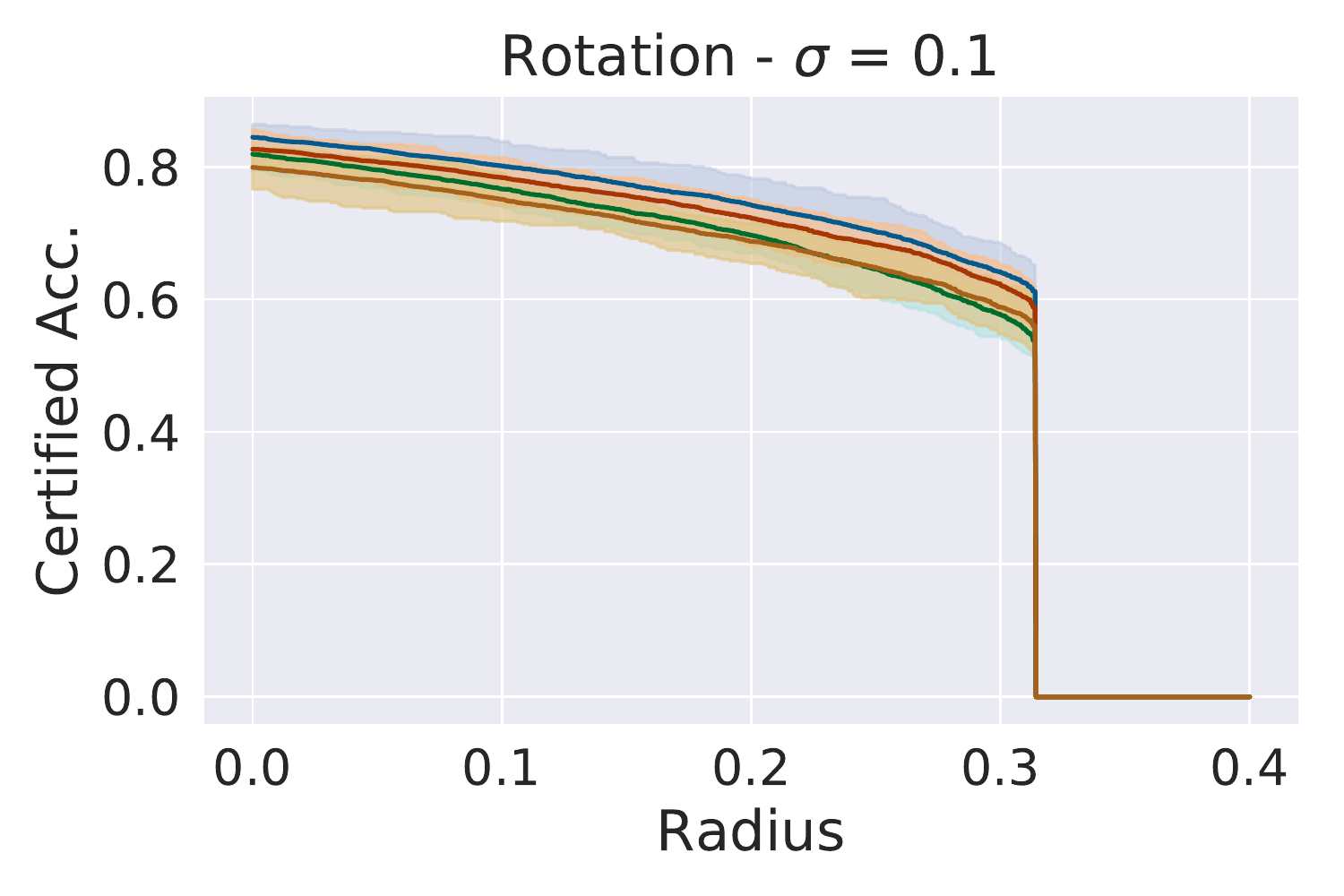}
    \includegraphics[width=0.3\textwidth]{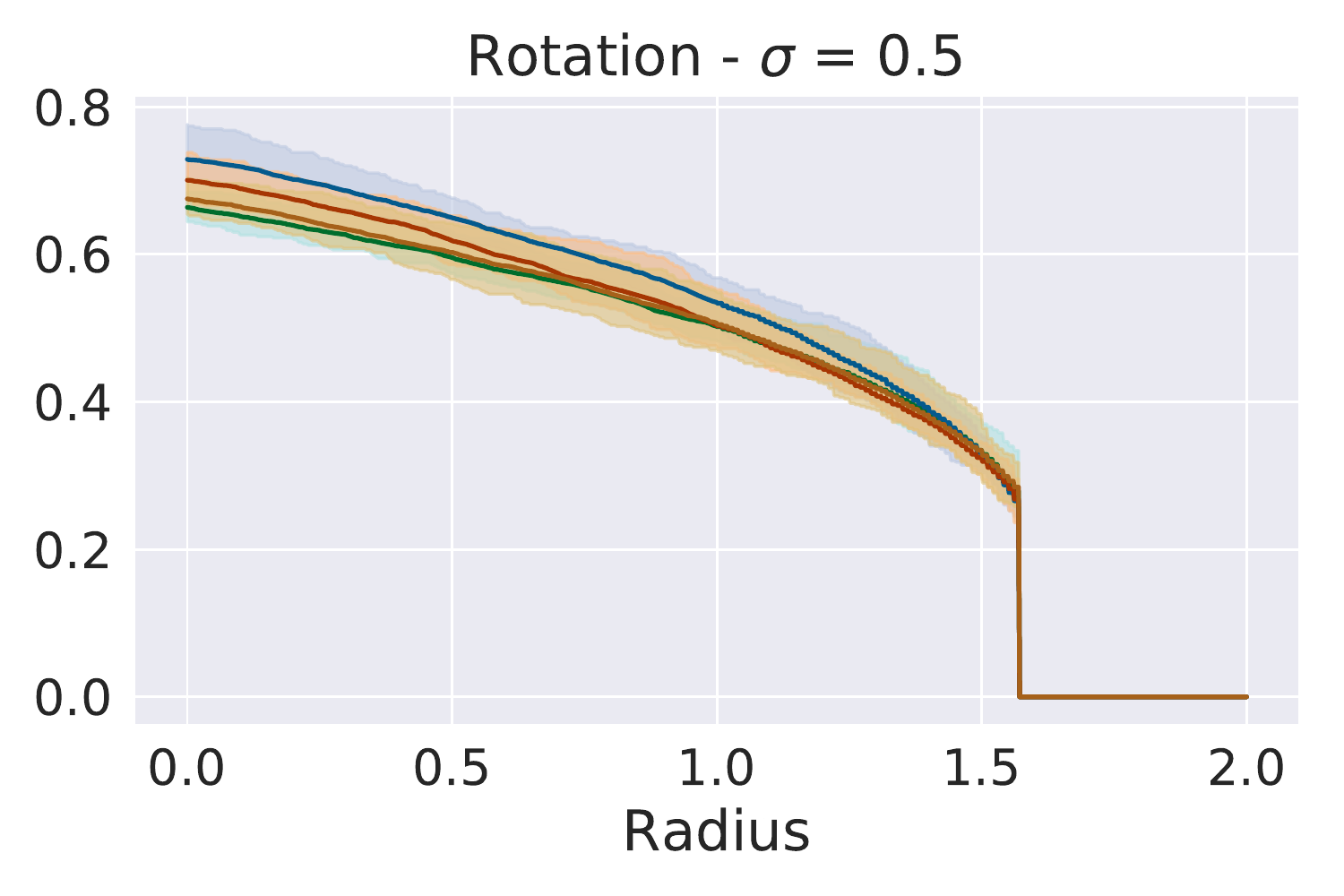}
    \includegraphics[width=0.3\textwidth]{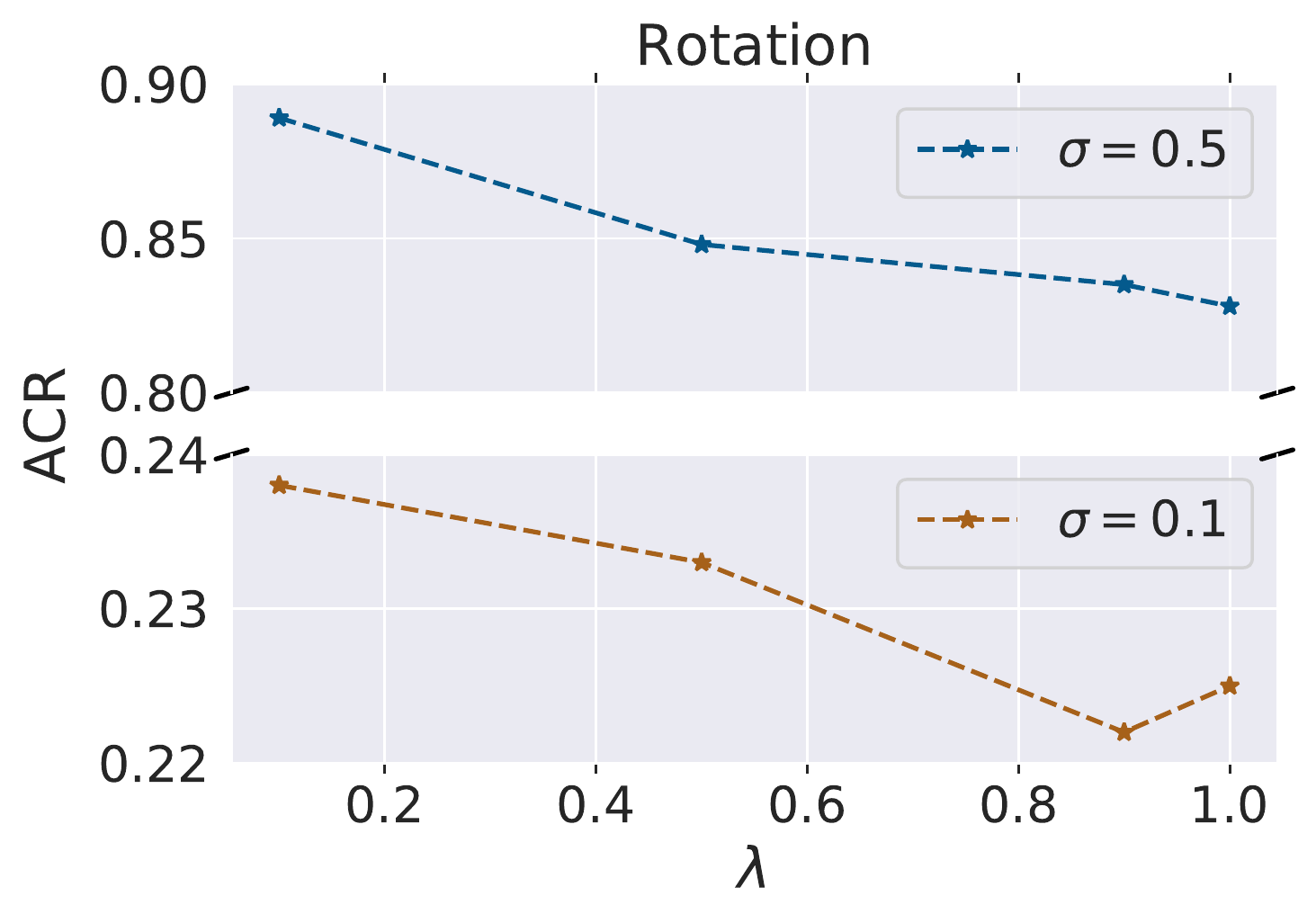} \\
    \includegraphics[width=0.3\textwidth]{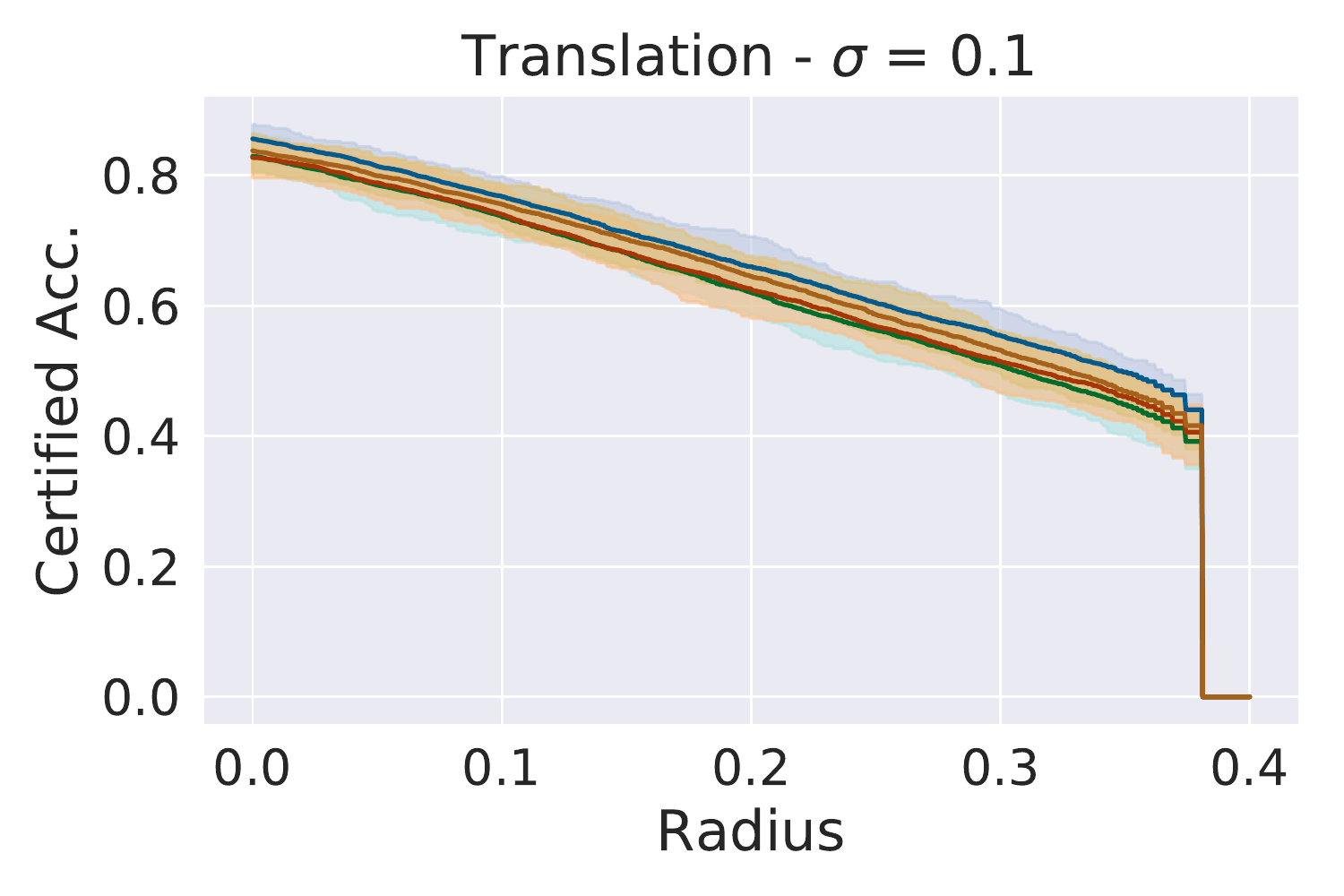}
    \includegraphics[width=0.3\textwidth]{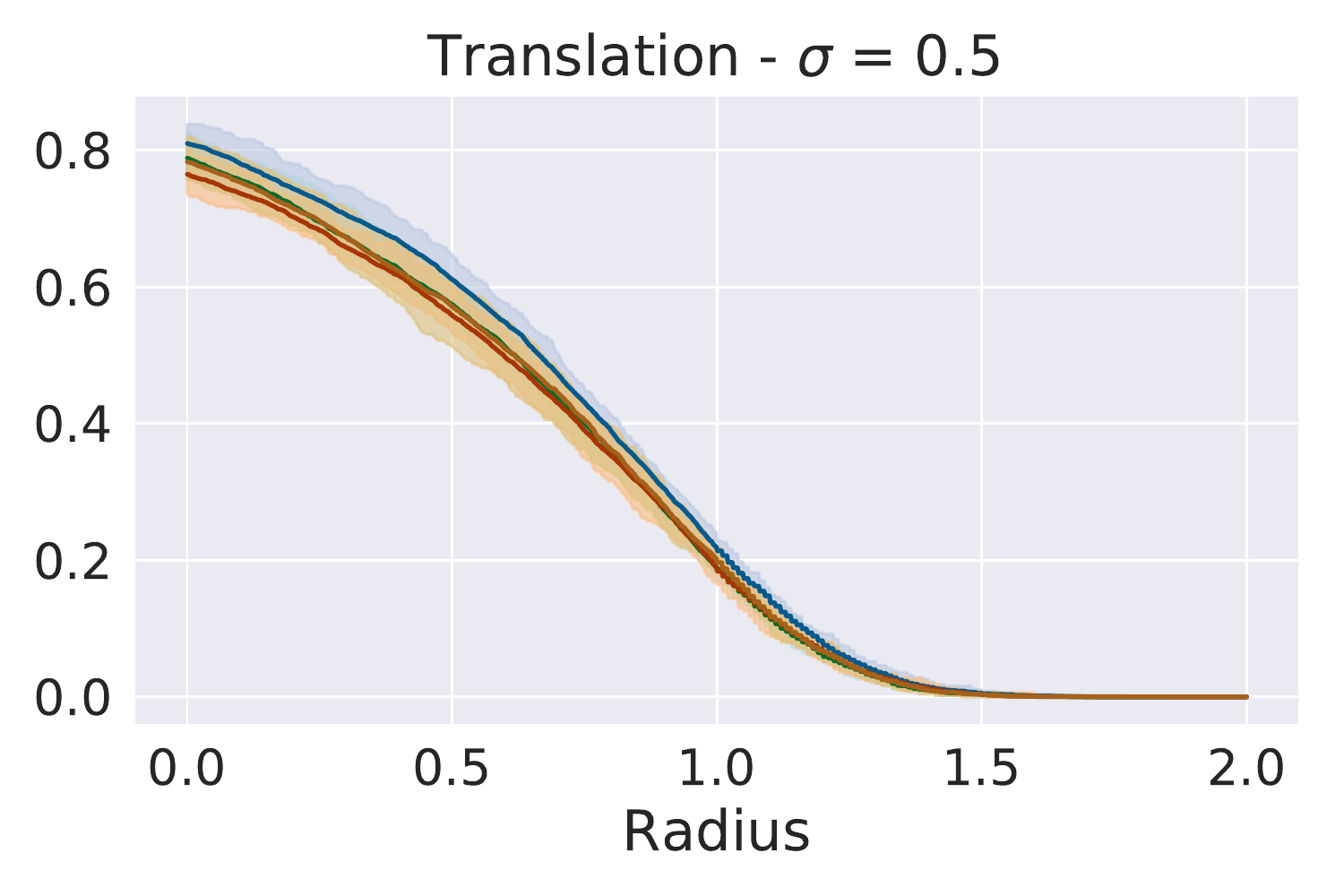}
    \includegraphics[width=0.3\textwidth]{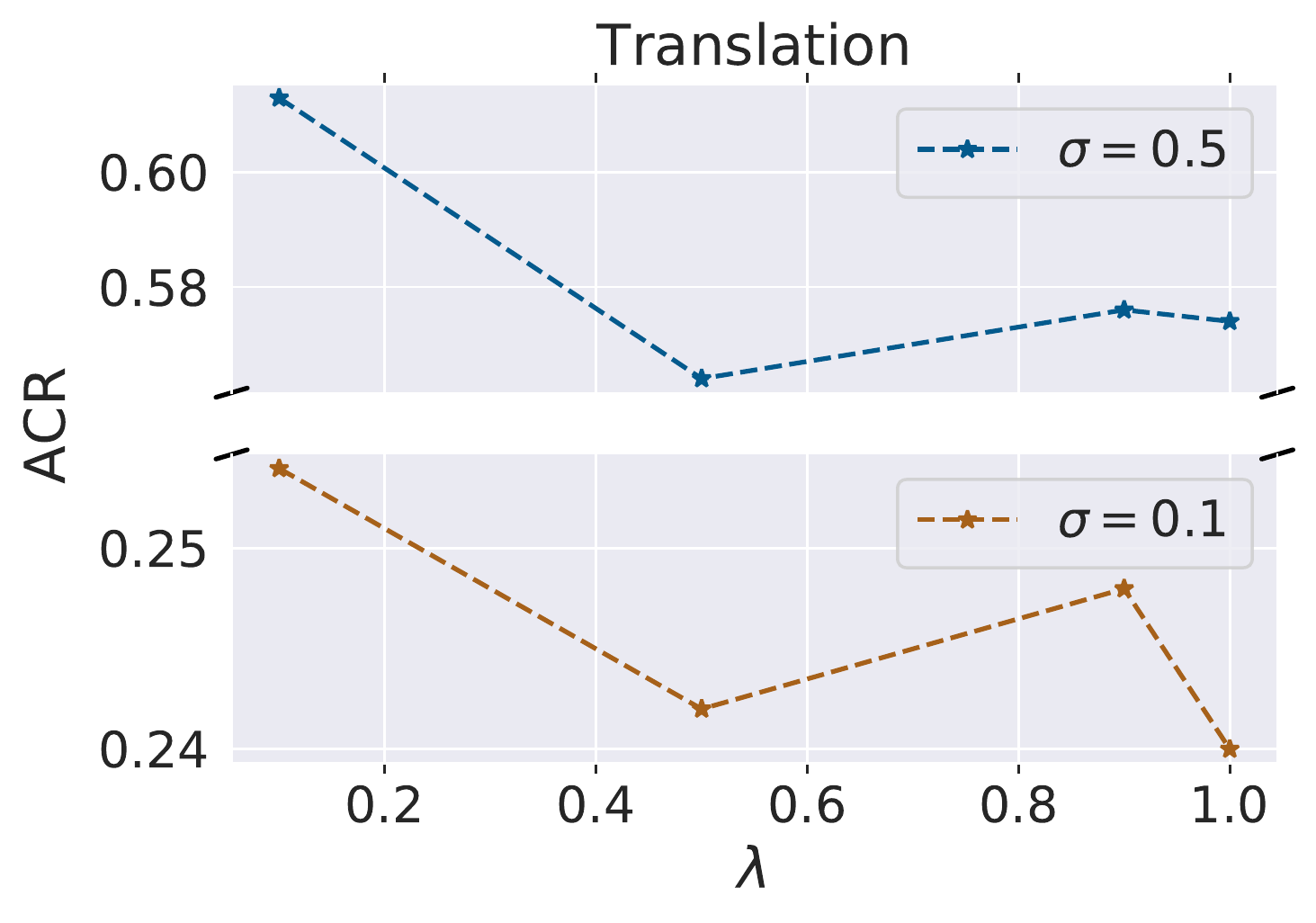}
\vspace{-0.4cm}
\caption{\textbf{Mixture of models.} 
    We explore the effect of ``intermediate'' levels of personalization by varying $\lambda \in \{0.1, 0.5, 0.9, 1.0\}$ from Eq.~\eqref{eq:implicit-personalization}.
    We experiment with rotation (top row) and translation (bottom row).
    The first two columns report certified accuracy curves, while the third one shows the associated the Average Certified Radius (ACR), \ie area under the curves, as a function of $\lambda$.
    The robustness of local models degrades, when they diverge from the global model.}
    \vspace{-0.5cm}
    \label{fig:models-mixture-translation-rotation}
\end{figure*}

We first investigate the benefits of federated training on certified robustness. 
To that regard, we use a rotation transformation and vary the number of clients on which the dataset is distributed.
We conduct experiments with 4, 10, and 20 clients, and report the results in Figure~\ref{fig:rotation-ablating-num-clients}. 
From these results, we draw the following observations. 
\textbf{(i)} When the number of clients is small (4 clients), \ie when each client has large data, local training can be sufficient for building a model that is accurate \textit{and} robust. 
\textbf{(ii)} As the number of clients increases, \ie a typical scenario in FL, and the amount of data available to each client is not large, federated averaging shines in providing a model with better performance and robustness. 
% That is, federated training can build models that are not only accurate, but also robust. 
In particular, we find federated training can bring robustness improvements of over 20\% in the few-local-data regime (10 or 20 clients).
This can be interpreted as follows: when local data can sufficiently approximate the real-data distribution, then there is no advantage to conducting federated training. 
On the other hand, when local data is insufficient for approximating the data-distribution, federated learning is useful.
\textbf{(iii)} The performance and robustness of personalized (fine-tuned) models are comparable to the global federated model. 
We attribute this observation to the federated model being trained with data augmentation, and thus converging to a reasonably-good local optimum. 
Hence, personalizing this model with few epochs of local training would not have very significant impact. 
We delve more into the benefits of personalization in the next section.
\textbf{(iv)} At last, we observe that, as the number of clients increases, the performance difference between the best and worst performing clients is significantly larger for local training. 
This is evidenced by the spread in Figure~\ref{fig:rotation-ablating-num-clients}, where the shaded area is notably larger for local training. 
% We provide more experiments elaborating this phenomenon in the appendix.

\subsection{Is Personalization Beneficial for Robustness?}

Next, we investigate the effect of personalization on certified robustness. 
In essence, the experiments in Figure~\ref{fig:rotation-ablating-num-clients} assumed all clients train robustly, aiming at collaboratively building a global robust model. 
However, in a more realistic federated setup, this is not necessarily the case. 
For instance, different clients could target robustness against different transformations or, in the extreme case, some clients could disregard robustness entirely and simply target accuracy.

To study this case, we use 10 clients and modify the averaging and personalization phases as follows. 
During the federated training phase, all models train without augmentations (\ie targeting accuracy and disregarding robustness). 
During the personalization phase, we augment data with the transformation for which the client is interested in being robust against. 
Experimentally, we personalize against pixel perturbations, rotations, and translations, and report results in Figure~\ref{fig:nominal-pretraining}. 
Our results show that personalization has a significant impact on improving model robustness. 
Moreover, this improvement is observed across the board, irrespective of the choice of the augmentation or 
% the smoothing parameter 
$\sigma$.

Next, we address the question: what if all other clients aim to build a model robust against  transformations that are different than the desired one?
We analyze the scenario when all clients train with an augmentation that differs from the one a certain client desires. 
Thus, we deploy augmentation with affine deformation~\cite{deformrs} during the federated training phase, and then personalize with either rotation or translation. 
We report the results in Figure~\ref{fig:affine-pretraining} showing that personalization enhances the robustness even when federated training is 
conducted with a different augmentation.
% (an affine deformation, in this case).

% I think it would be good to contrast this results to the first plot, when federated robust training is performed. Namely, how transferable is the resulted model in terms of robustness to other perturbation?

\subsection{Advantages of Personalization}\label{sec:advantages-of-personalization}

Further, we address the following question: when is it better to robustly train on local data than to do robust personalization (starting from a nominally trained federated model)?
We examine this question by comparing the performance of a nominally trained federated model that is robustly personalized (Figure~\ref{fig:nominal-pretraining}) against local robust training. 
We display this comparison in Figure~\ref{fig:nominal-pretraining-all}. 
We observe the following: \textbf{(i)} When the target local transformation is similar to the target federated one (small values of $\sigma$), personalized models significantly outperform local models, despite robust training only occurring in the personalization phase.
That is, with far less computation and much faster training\footnote{While our experiments used simple data augmentation, practitioners may use more effective (and computationally expensive~\cite{madry2018towards}) alternatives for robust training.}, federated training with personalization offers a better alternative to training only on local data. 
Furthermore, the performance gap increases with the number of clients---equivalent to assigning each client fewer local data. 
\textbf{(ii)} As the difference between the transformation in the federated and personalized settings increases, the performance of personalized models becomes comparable to that of their local counterparts. 
Note that this is an artifact of choosing a relatively small number of clients, on which the dataset is distributed. 
Indeed, in the appendix we report how a larger number of clients ($N=100$) displays significant robustness improvements of personalized models compared to local models.

% \juannote{I think the language in this section is too complex: nouns have names that are too long. I worked on it a bit, but it can certainly use more passes}

\subsection{FL with Mixture Model}\label{sec:mixture-model}

At last, we explore how using the mixture of models formulation in Eq.~\eqref{eq:implicit-personalization} affects robustness of the resulting local models. 
Note that through all the aforementioned experiments, we studied the case when $\lambda=1.0$ for local training, or $\lambda=0.0$ for federated training.
To this end, we compare the performance of models trained with various values $\lambda\in(0, 1)$,
% \in \{0.1, 0.5, 0.9, 1.0\}
leading to different personalization levels. 
That is, during local training, each client is leveraging the update step in \eqref{eq:mixture_update} before communicating the local model to the orchestrating server for an averaging step. 
% Note that setting $\lambda = 1.0$ corresponds to standard federated training for solving the classical ERM problem and one model-fits-all paradigm $\theta_1 = \dots = \theta_n$.

Figure~\ref{fig:models-mixture-translation-rotation} plots the results of this experiment. 
The first two columns display certified accuracy curves against rotation and translation for different $\lambda$ values. 
The last column summarizes the results in terms of Average Certified Radius (ACR), \ie the average certified radius of correctly-classified examples~\cite{Zhai2020MACER:}. 
From these results we observe that: \textbf{(i)} for every deformation (rotation, translation) and every $\sigma$ value, the mixture of models formulation outperforms the robustness of standard federated training. 
This phenomenon is consistent with almost \textit{any} value of the personalization parameter $\lambda$. 
In particular, we observe that setting $\lambda=0.1$, corresponding to the \textit{least} personalized local models, yields the best results. 
\textbf{(ii)} By contrasting these results to Figure~\ref{fig:nominal-pretraining-all}, we notice that, across clients, the spread of certified accuracy is smaller than both naïve personalization (via simple local fine-tuning) and local training.

%%%%%%%%%%%%%%%%%%%%%%%%%%%%%%%%%%%%%%%%%%%%%%%%%%%%%%%%%%%%%%%%%%%%%%%%%%%

\section{Conclusions}
We analyzed the certified robustness of deep networks trained in a federated learning setup. 
We found that federated averaging and personalization deliver  simple yet effective methods for providing models that are both more accurate and robust than locally-trained models. 
% Moreover, we studied the interplay between personalization and robustness when federated training is conducted in a non-robust fashion, 
% showing several advantages of personalization over local and global training.
% and found a non-trivial interaction between a model's level of personalization and its certified robustness.
At last, we examined the effect of deploying a mixture of local and global models on certified robustness.
% of the trained model.
% finding significant robustness enhancements in several setups.

% There exist several frameworks that characterize federated training and personalization. 
% A limitation of our work is the focus on the popular federated averaging technique. 
% A possible interesting avenue for future work is to benchmark different federated and personalization frameworks in terms of robustness. 
% % Moreover, new mixture models could be devised that explicitly target robustness enhancement are left for future works.
% Moreover, the development of new mixture models explicitly aimed at improving robustness and reliability is also left for future work.

% \section{Limitations and Future Work}
% PlaceHolder

\newpage
\bibliography{main}

% \appendix
\clearpage

\appendix

% \part{Supplementary material}
\onecolumn
\begin{center}
    \huge
    \textbf{Certified Robustness in Federated Learning \\ 
Supplementary Materials}
\end{center}

% \vspace{-50pt}

% \tableofcontents
\section{Additional Experiments}

\subsection{Is Federated Averaging Beneficial for Robustness}
\begin{figure}[b]
    \centering
    \includegraphics[width=0.245\textwidth]{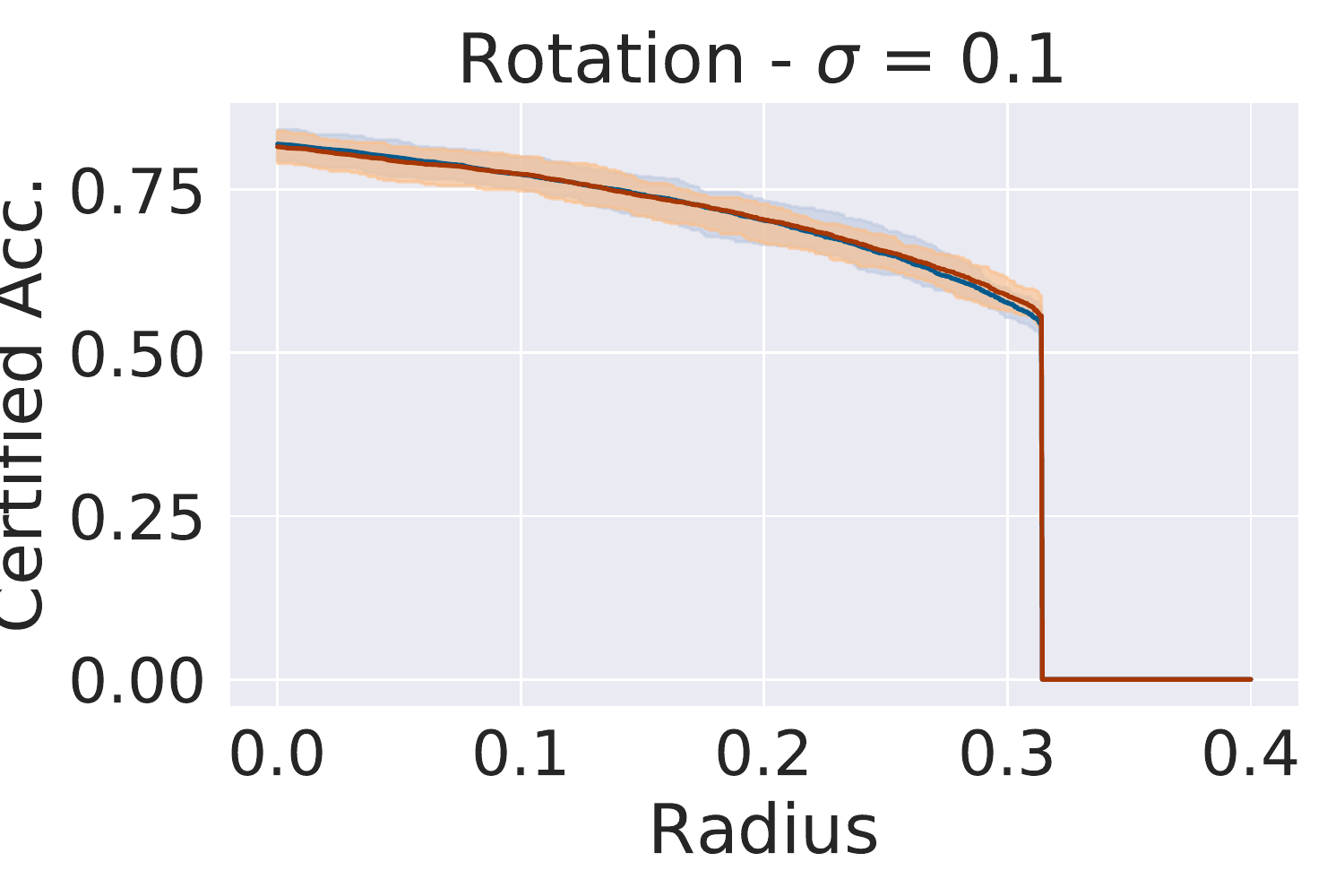}
    \includegraphics[width=0.245\textwidth]{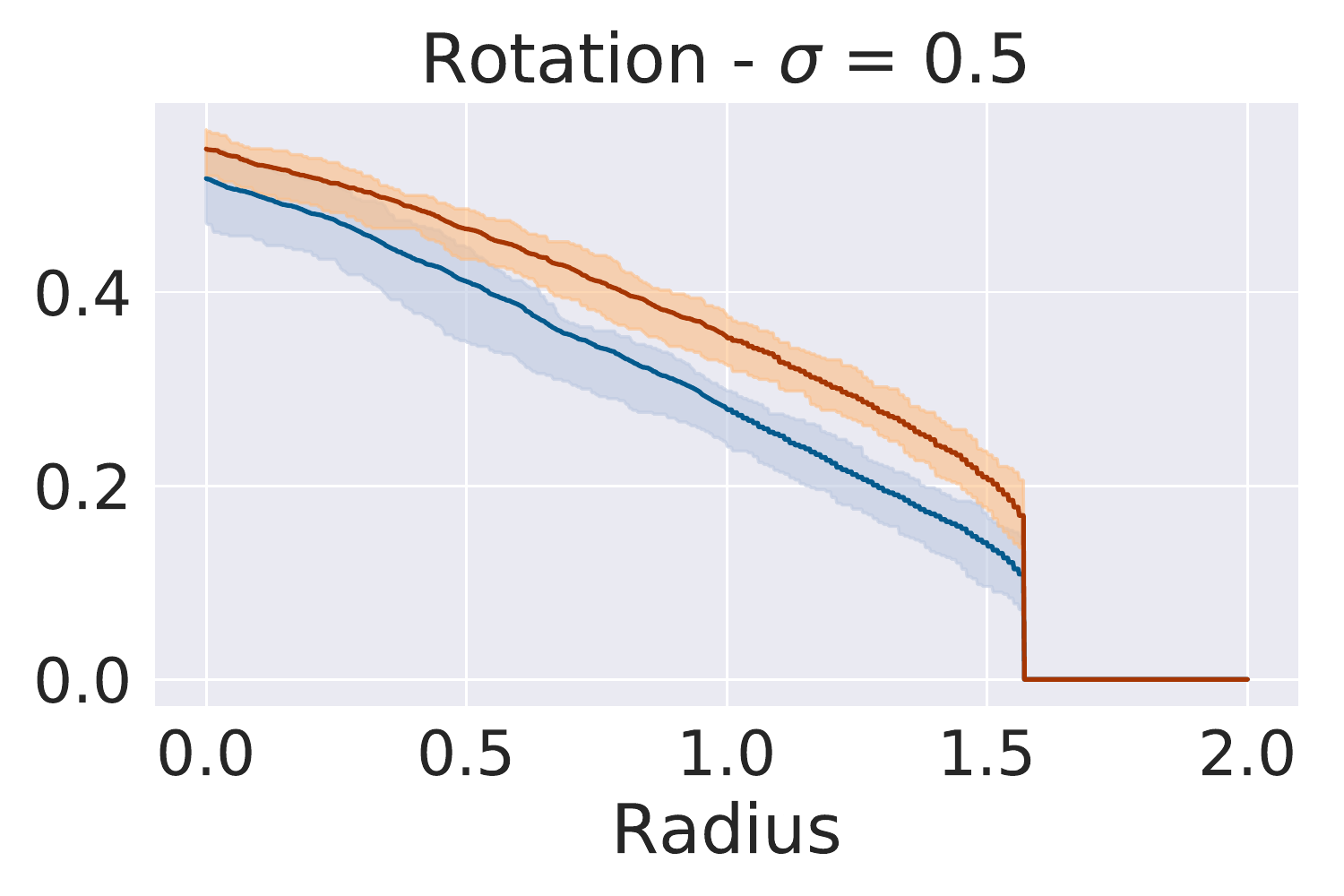}
    \includegraphics[width=0.245\textwidth]{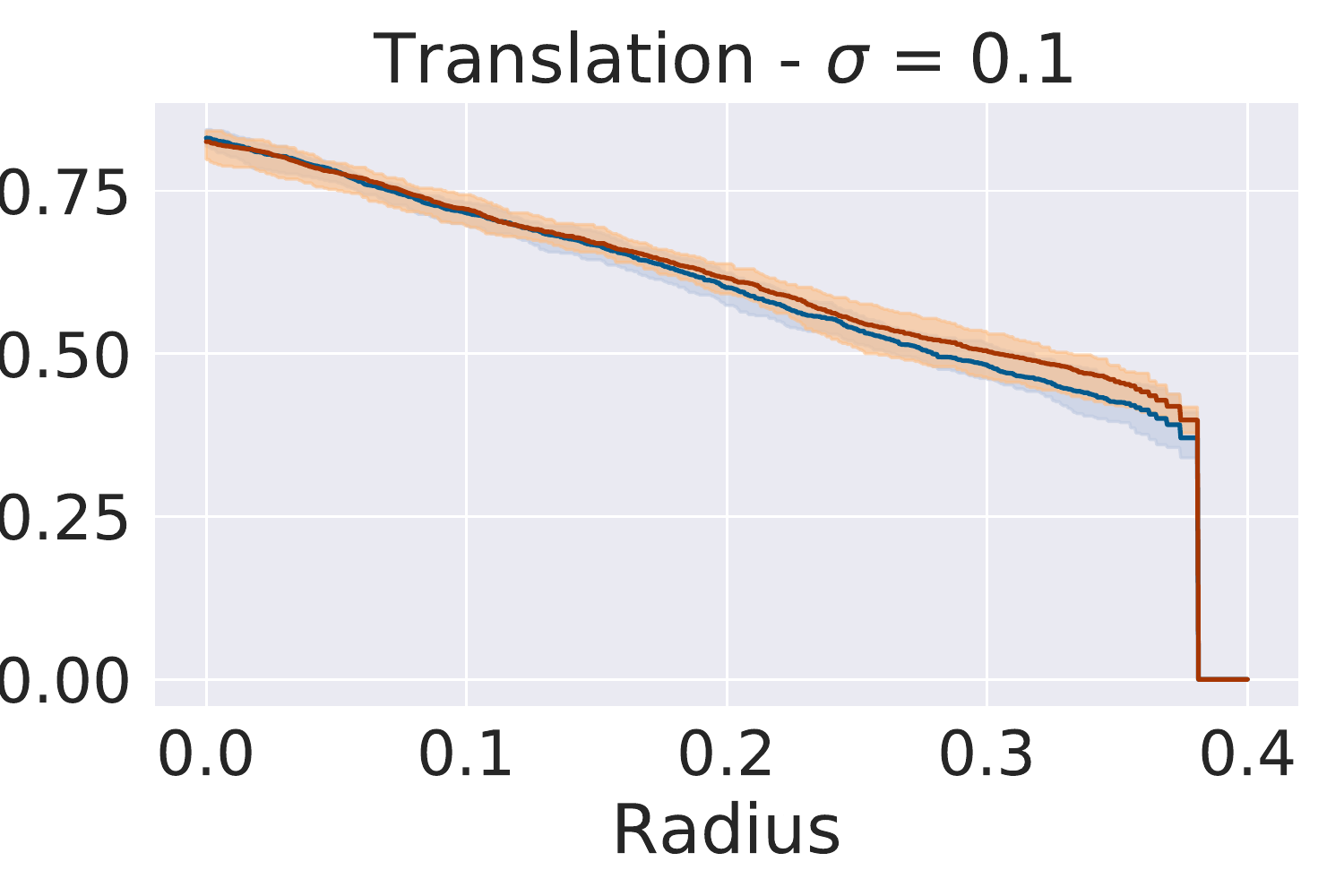} 
    \includegraphics[width=0.245\textwidth]{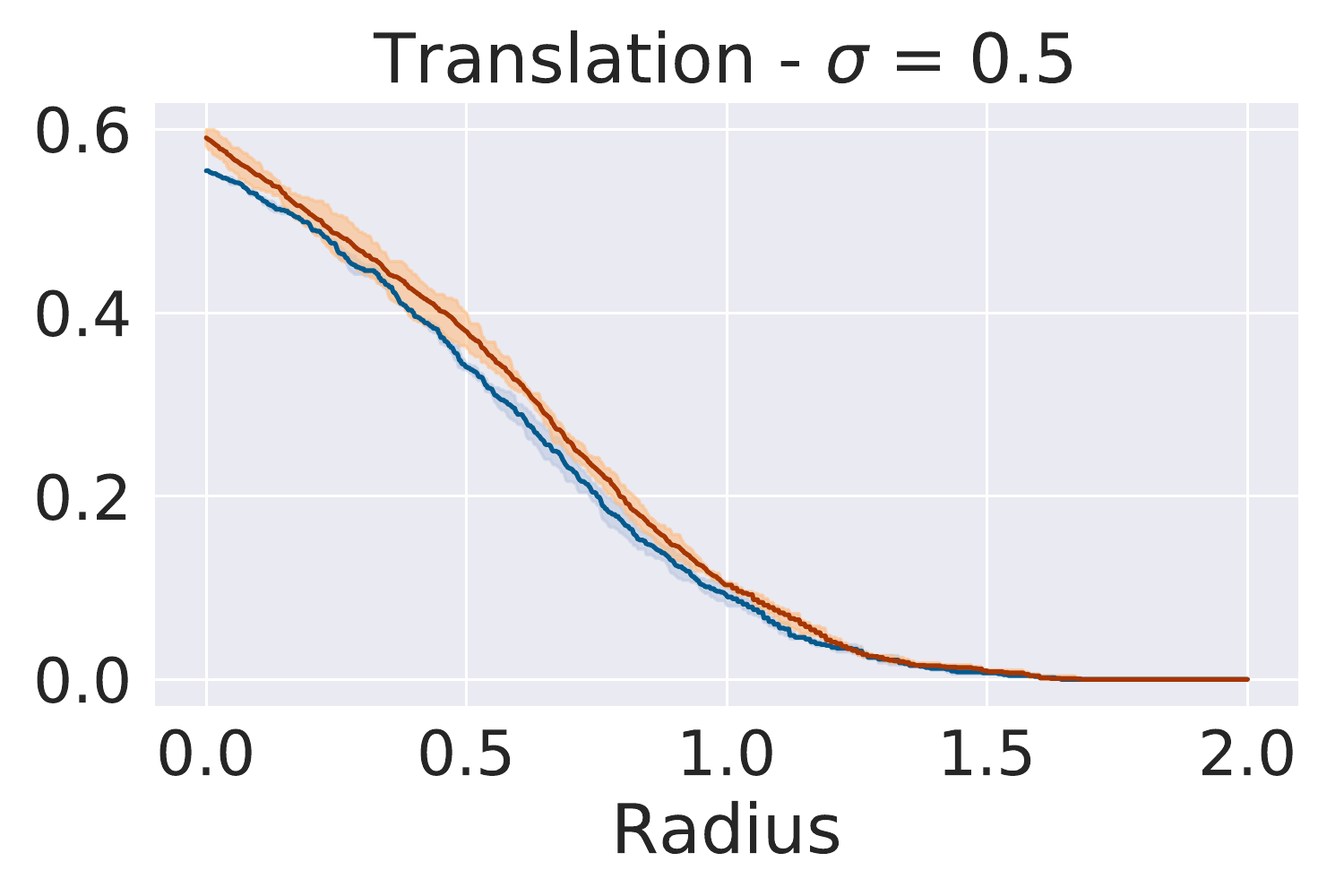}    
    \caption{\textbf{Certified Robustness against Rotation starting from Affine. Effect of Personalization} 
    Left: personalization with rotation augmentation with $\sigma\in\{0.1, 0.5\}$. 
    Right: translation augmentation with $\sigma\in\{0.1, 0.5\}$.
    The color convention follows that of Figure~\ref{fig:nominal-pretraining-all}.}
    \label{fig:affine-pretraining}
\end{figure}

In Section 4, we analyzed the benefits of personalization on the certified robustness when all clients are training nominally.
Next, we address the following question: what if all other clients aim to build a model robust against  transformations that are different than the desired one?
We analyze the scenario when all clients train with an augmentation that differs from the one a certain client desires. 
Thus, we deploy augmentation with the general affine deformation~\cite{deformrs} during the federated training phase, and then personalize with either rotation or translation. 
We report the results in Figure~\ref{fig:affine-pretraining} and illustrate that personalization has a positive impact on robustness even when federated training is conducted with a different augmentation (an affine deformation, in this case).

\begin{figure}[h]
    \centering
    \includegraphics[trim={0 10.4cm 0 0},clip , width=0.9\textwidth]{sections/images/common_legend.pdf}
    \includegraphics[width=0.49\textwidth]{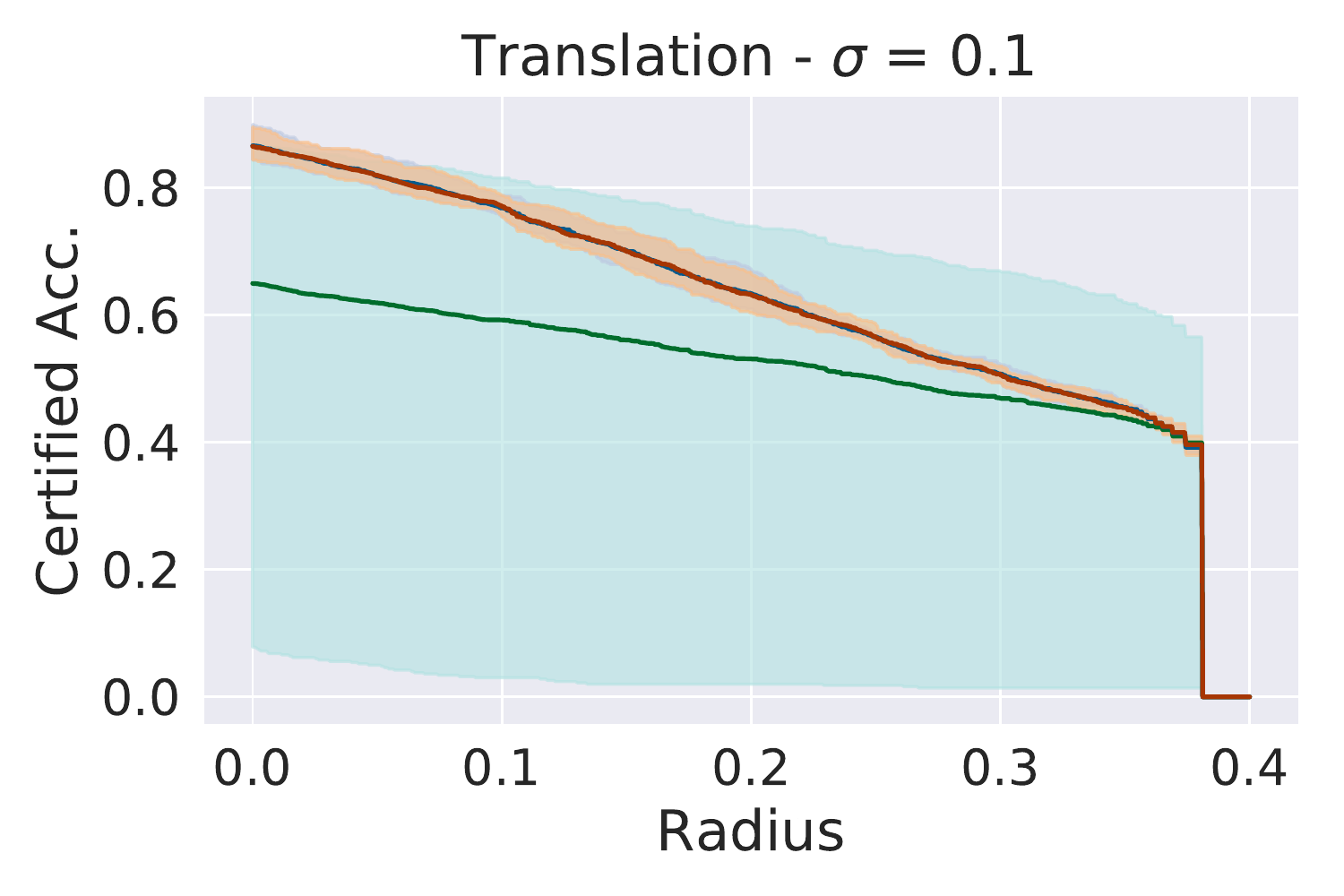}
    \includegraphics[width=0.49\textwidth]{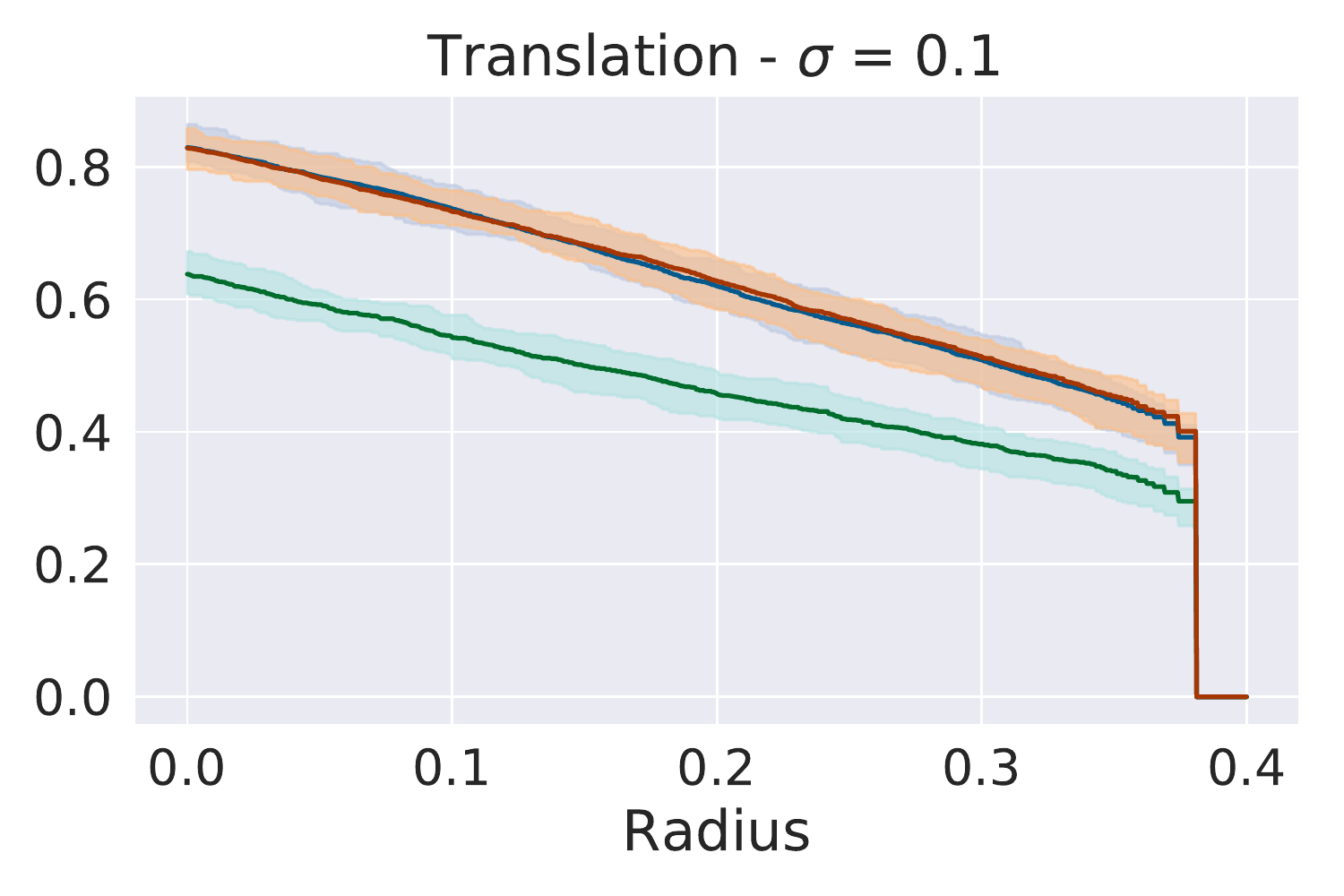}
    \includegraphics[width=0.49\textwidth]{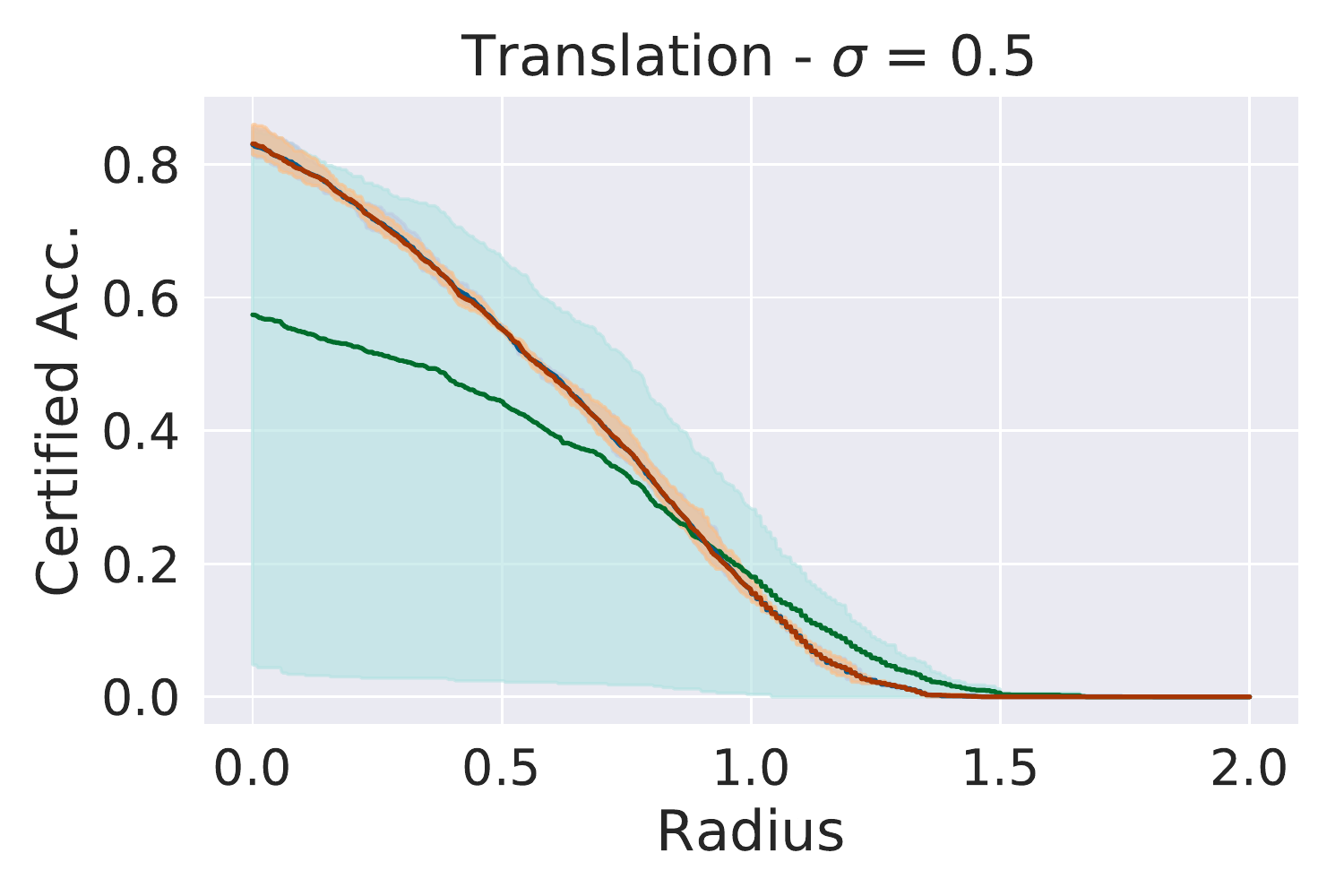}
    \includegraphics[width=0.49\textwidth]{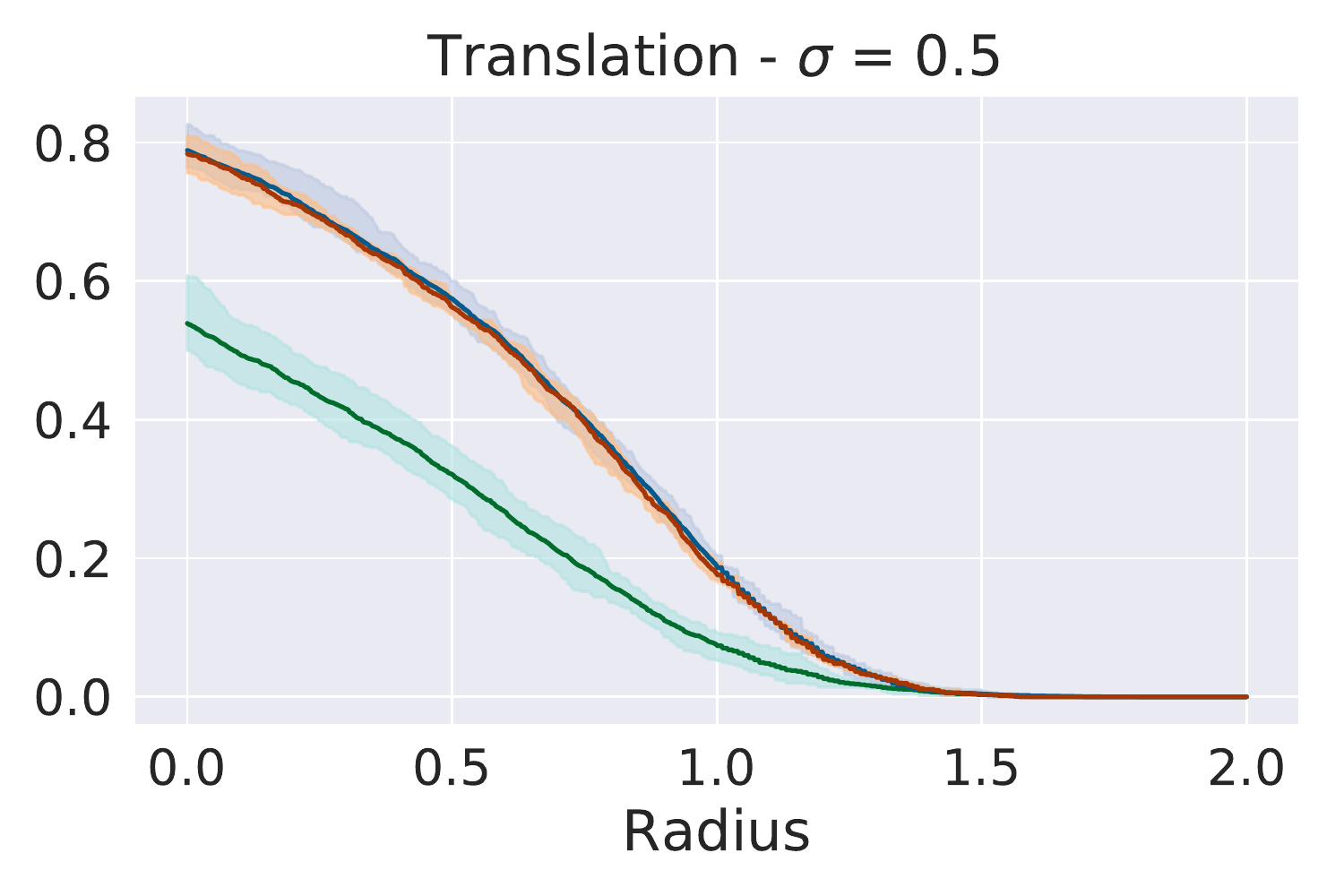}
    \caption{\textbf{Certified Accuracy against translation.} 
    We plot the certified accuracy curves for translation deformation by varying $\sigma \in\{0.1, 0.5\}$ in the top and bottom rows, respectively. 
    We vary the number of clients on which we divide the dataset. 
    Left: 4 clients, right: 10 clients. 
    As the number of clients increases, performance of local training worsens.}
    \label{fig:n_clients_ablation_on_translation}
\end{figure}

In Figure~\ref{fig:rotation-ablating-num-clients}, we analyzed the effect of varying the number of clients on the certified robustness against rotations. 
For completeness, we repeat the experiment with translations, and report the results in Figure~\ref{fig:n_clients_ablation_on_translation}. 
In particular, in this experiment we distribute CIFAR10 on 4 and 10 clients, and vary $\sigma\in\{0.1, 0.5\}$. 
Analogous to our earlier observation, as the number of clients increases, the local data becomes insufficient, and the performance and robustness of local training deteriorates. 

\begin{figure}
    \centering
    \includegraphics[trim={0 10.4cm 0 0},clip , width=0.9\textwidth]{sections/images/common_legend.pdf}
    \includegraphics[width=0.49\textwidth]{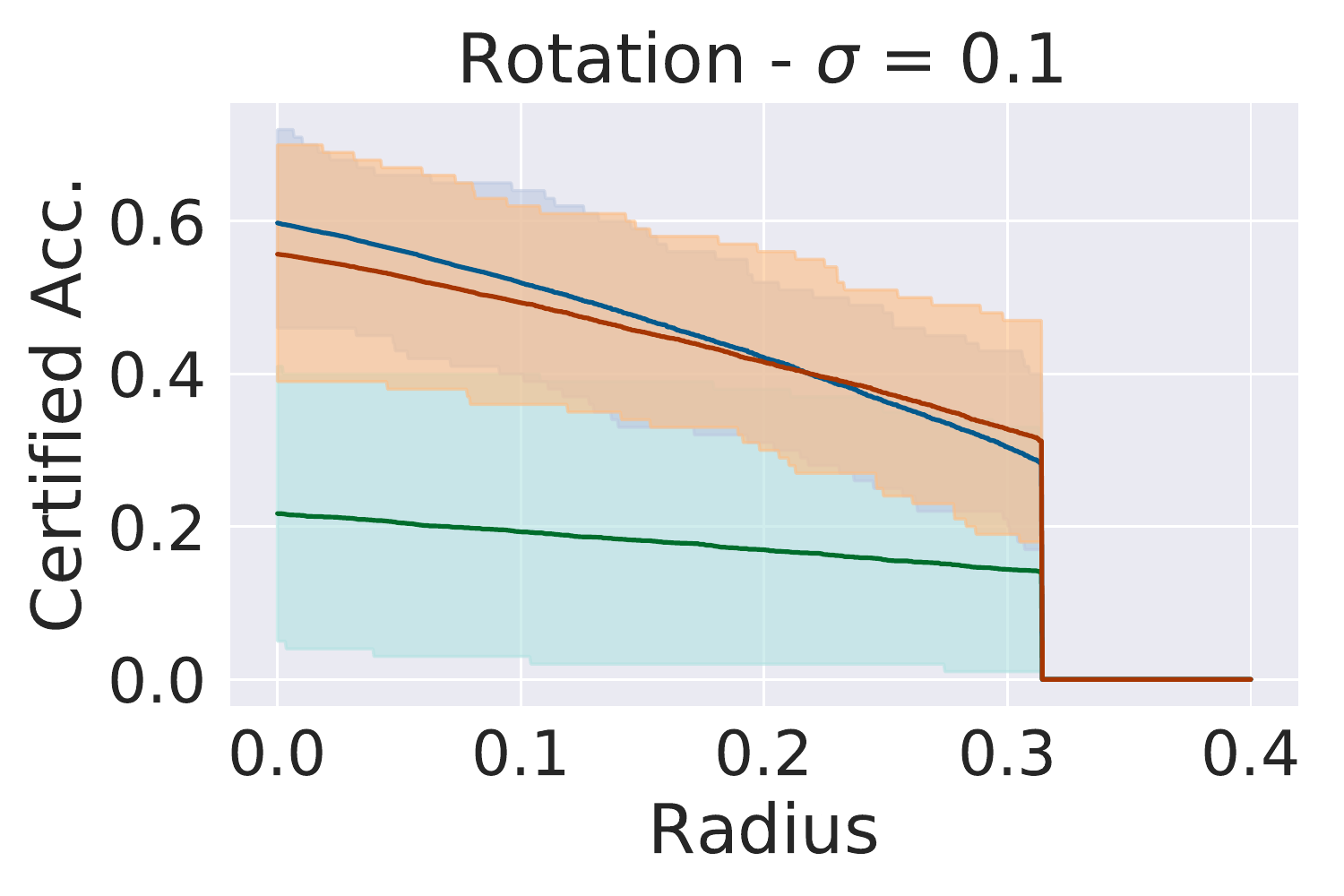}
    \includegraphics[width=0.49\textwidth]{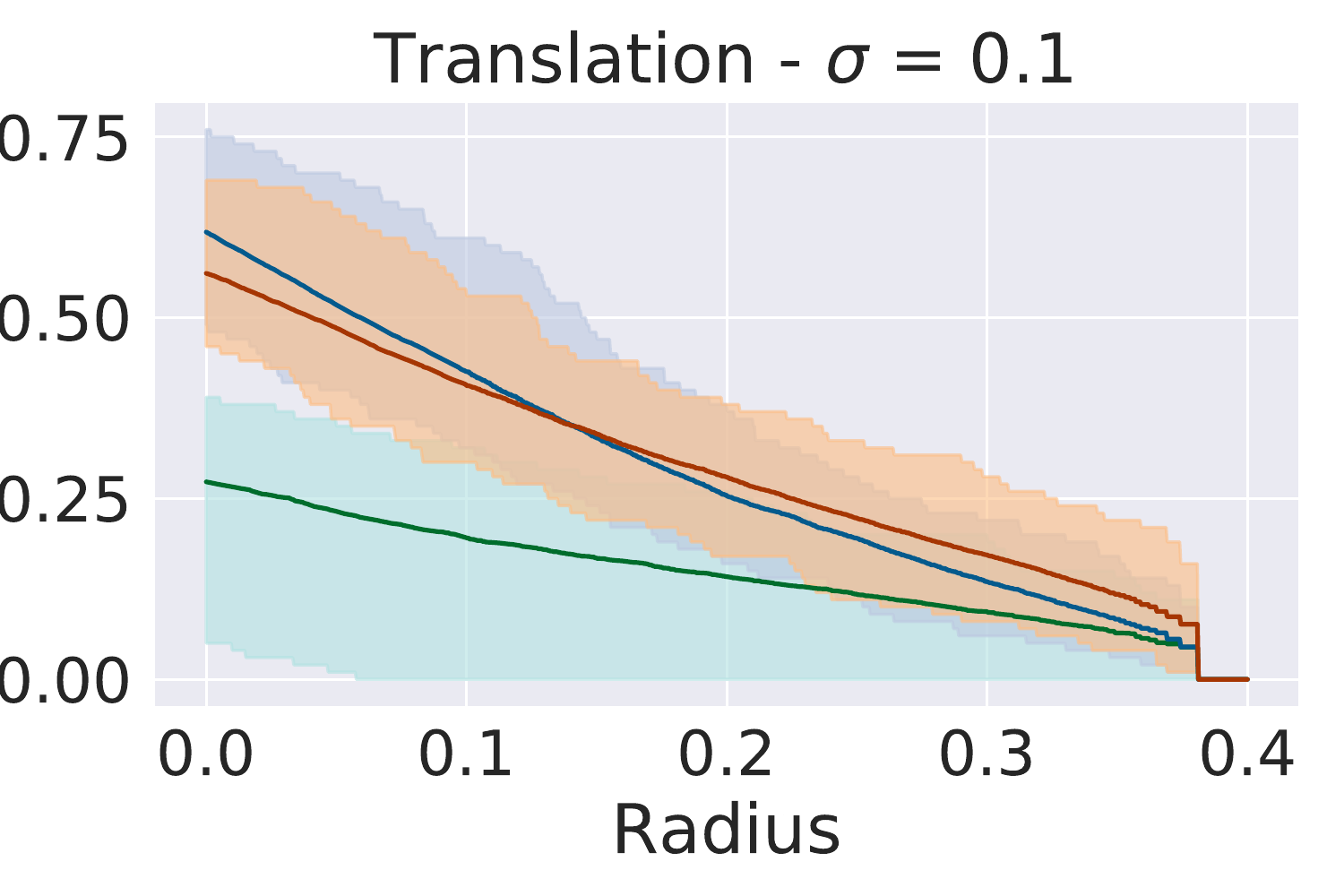}
    \includegraphics[width=0.49\textwidth]{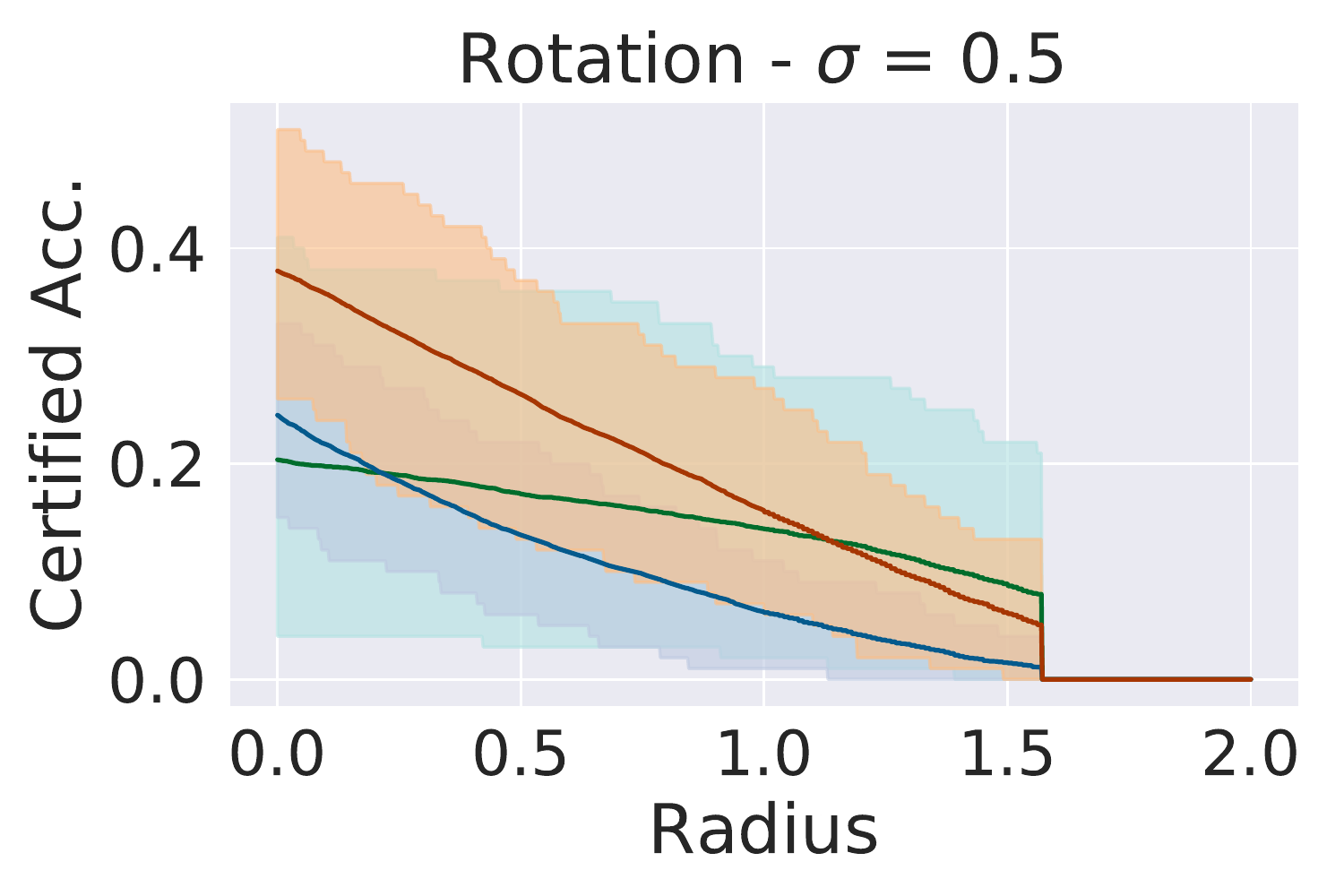}
    \includegraphics[width=0.49\textwidth]{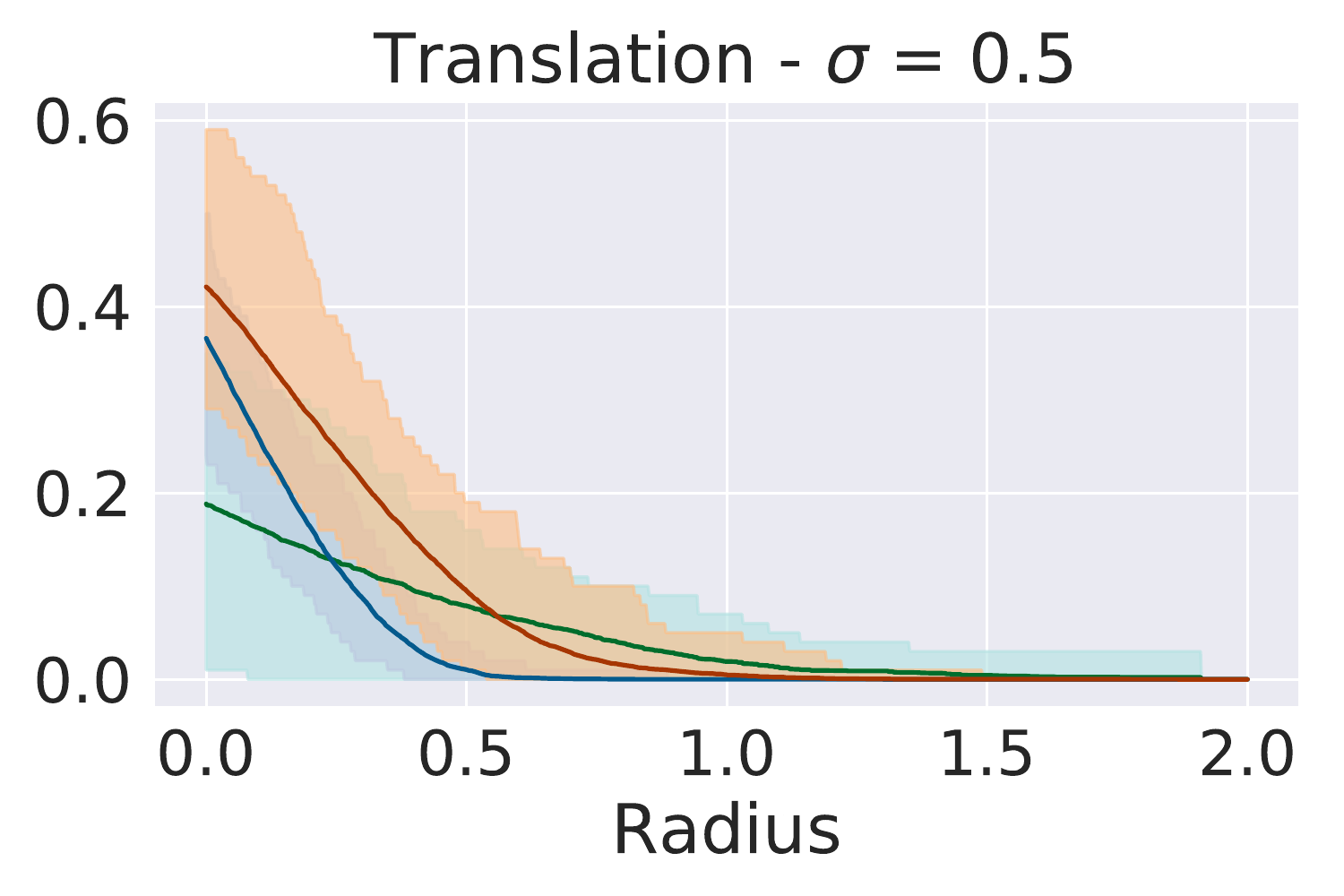}
    \caption{\textbf{Nominal pre-training and robust personalization with 100 clients.} 
    We compare local robust training to nominal federated training with robust personalization in terms of certified accuracy on CIFAR10.}
    \label{fig:100-clients-all}
\end{figure}
\subsection{100 Clients Experiment}
In Section~\ref{sec:advantages-of-personalization}, we observed that when the augmentation deployed in the federated training phase differs from the one used in the personalization phase, the performance of local models is comparable to that of personalized models.
We scale our experiments by distributing the dataset on a larger number of clients (100).
This makes the local data for each client insufficient for training local models. 
We plot the certified accuracy curves of local training, federated training, and personalized training in Figure~\ref{fig:100-clients-all}.
We observe that the performance gap increases between the personalized and local models, providing further evidence to support our hypothesis from Section~\ref{sec:advantages-of-personalization}.
That is, federated training and personalization, compared to local training, provide a more reliable approach that improves both performance and robustness in more realistic federated scenarios.
 
 \subsection{Ablation for All Clients}
\begin{figure}
    \centering
    \includegraphics[trim={0 6.8cm 0 0},clip , width=0.9\textwidth]{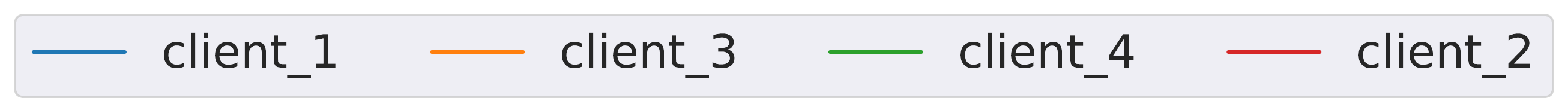}
    \includegraphics[width=0.49\textwidth]{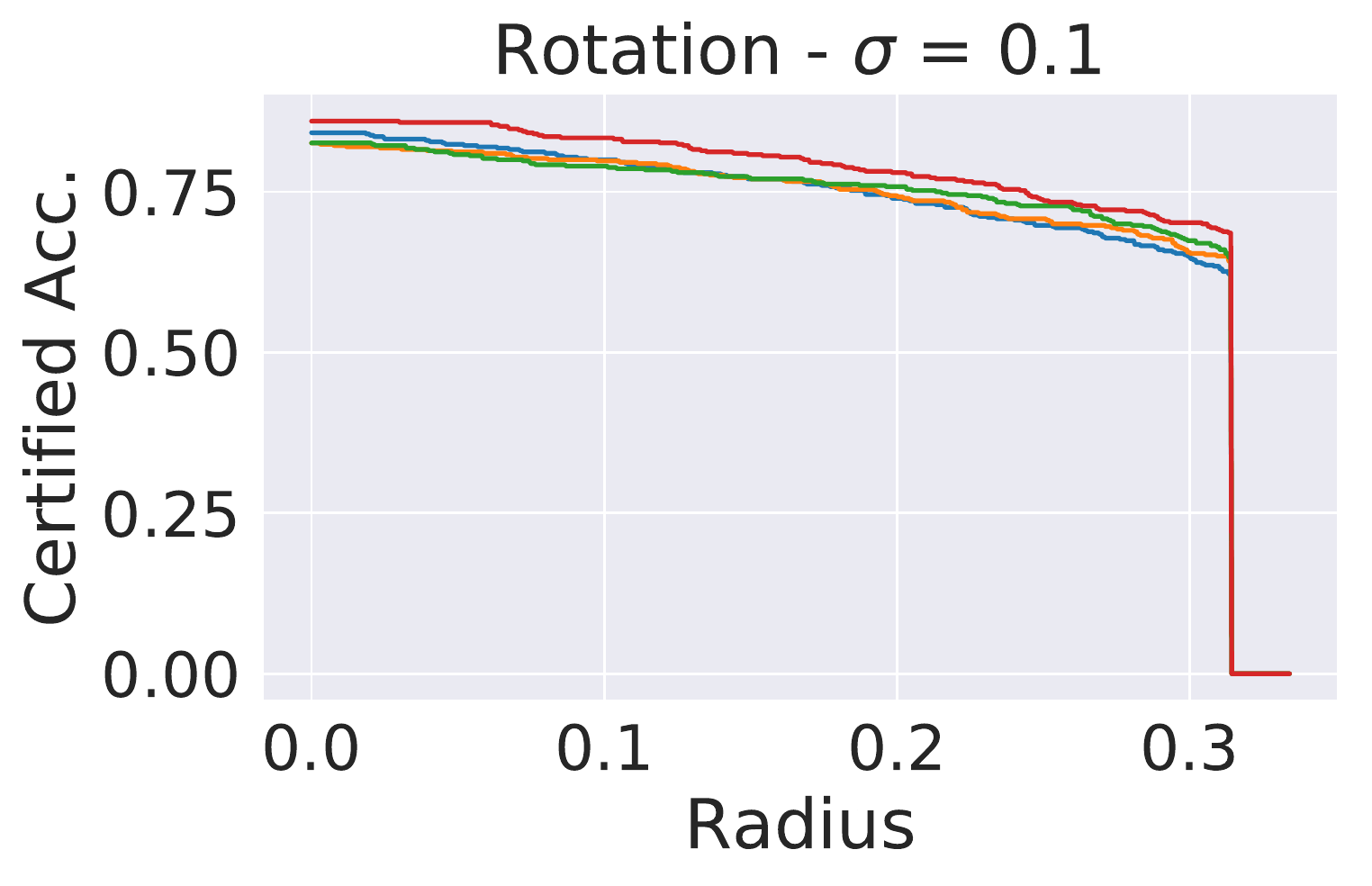}
    \includegraphics[width=0.49\textwidth]{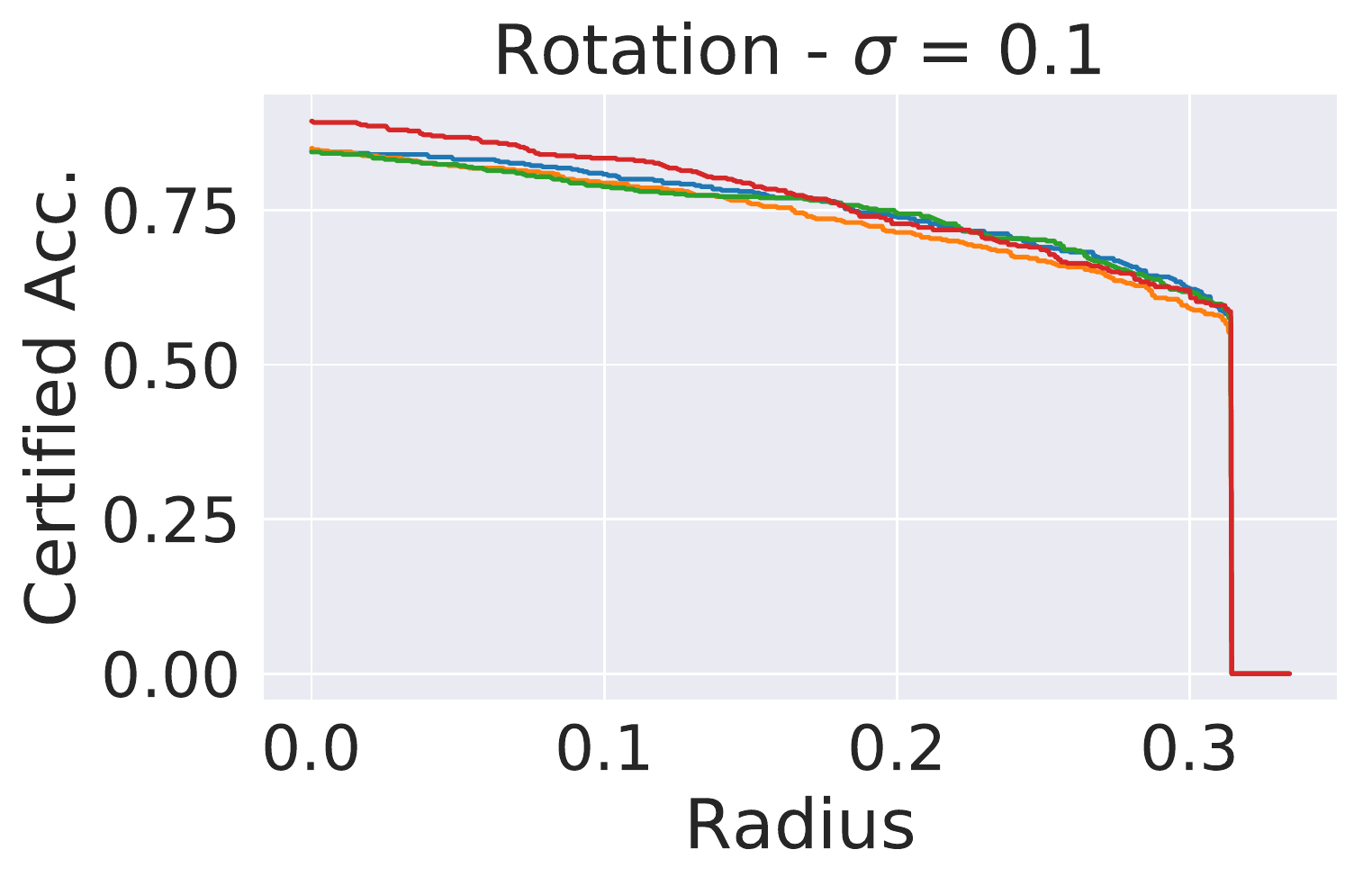}
    \includegraphics[width=0.49\textwidth]{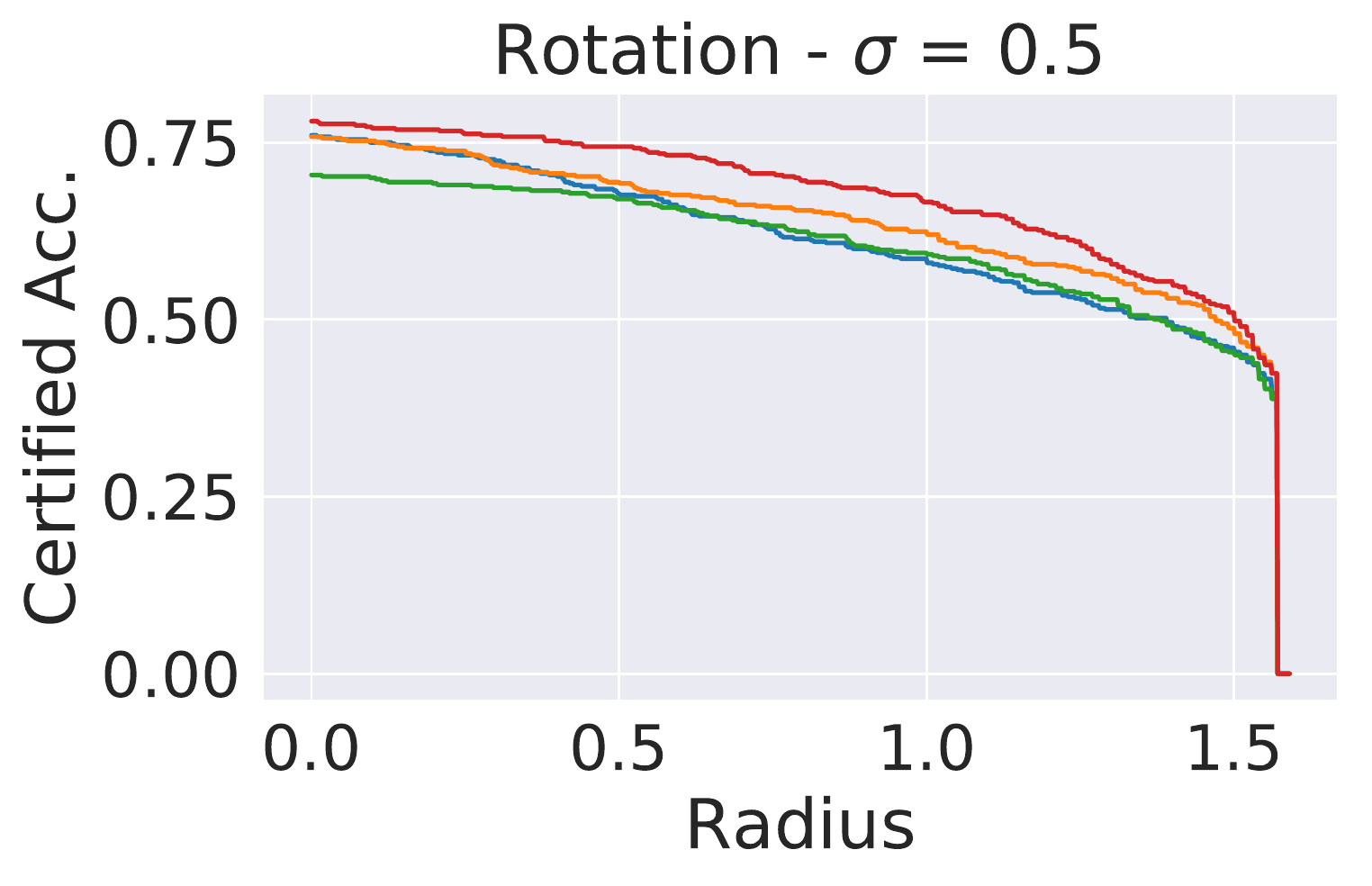}
    \includegraphics[width=0.49\textwidth]{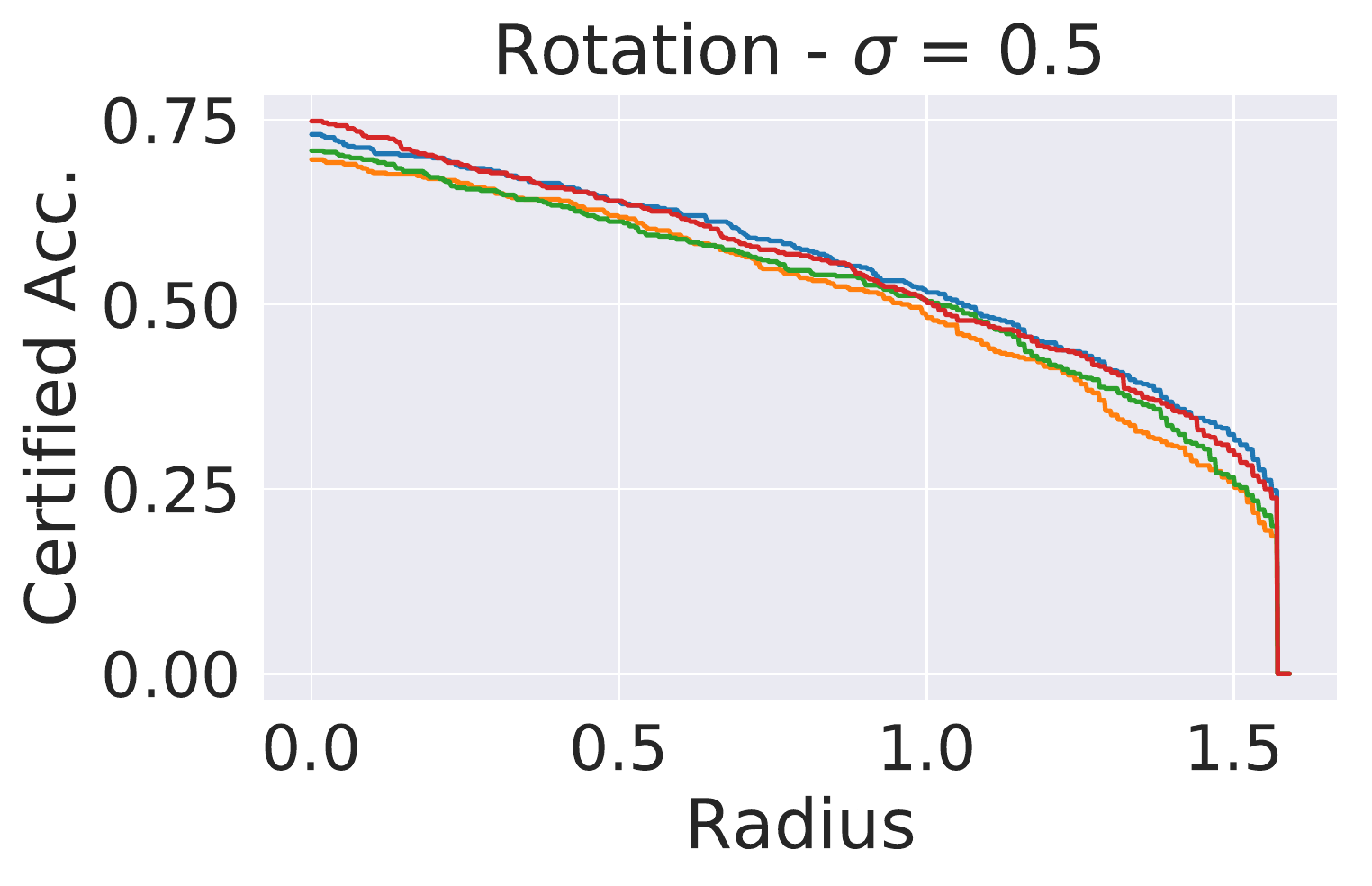}
    \caption{\textbf{Ablations for all clients. 4 Clients experiment} 
    Left: Local training. 
    Right: Personalized training.}
    \label{fig:ablations-all-clients-4-clients}
\end{figure}

\begin{figure}
    \centering
    \includegraphics[trim={0 10.8cm 0 0},clip , width=0.9\textwidth]{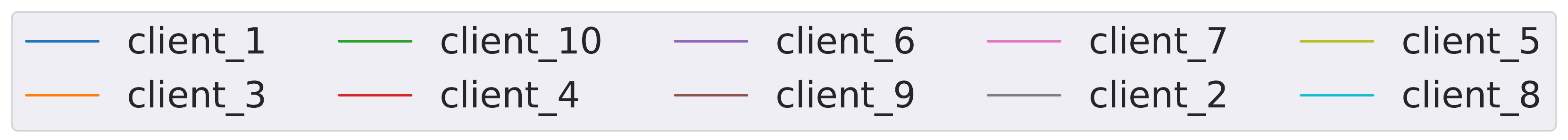}
    \includegraphics[width=0.49\textwidth]{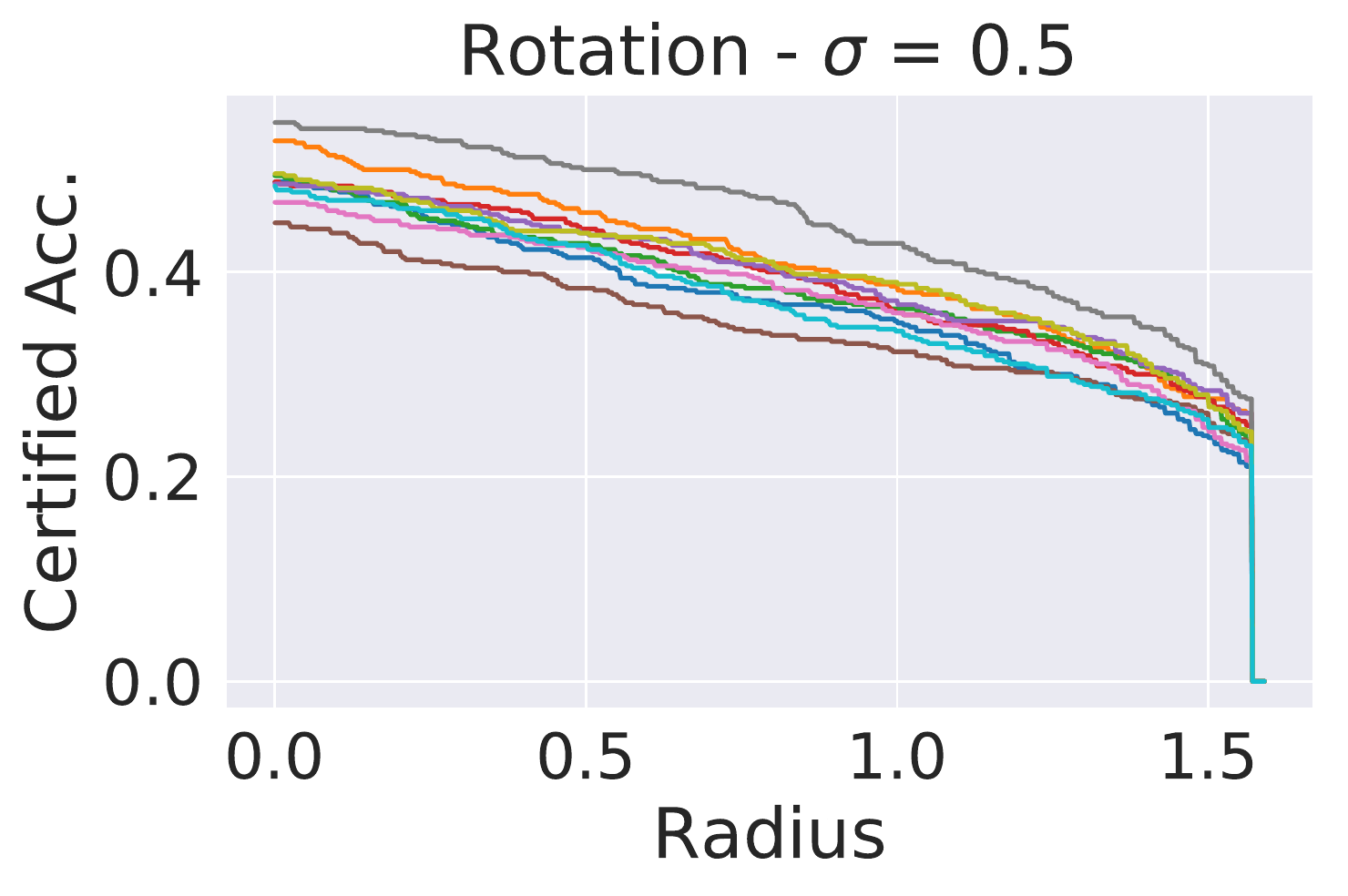}
    \includegraphics[width=0.49\textwidth]{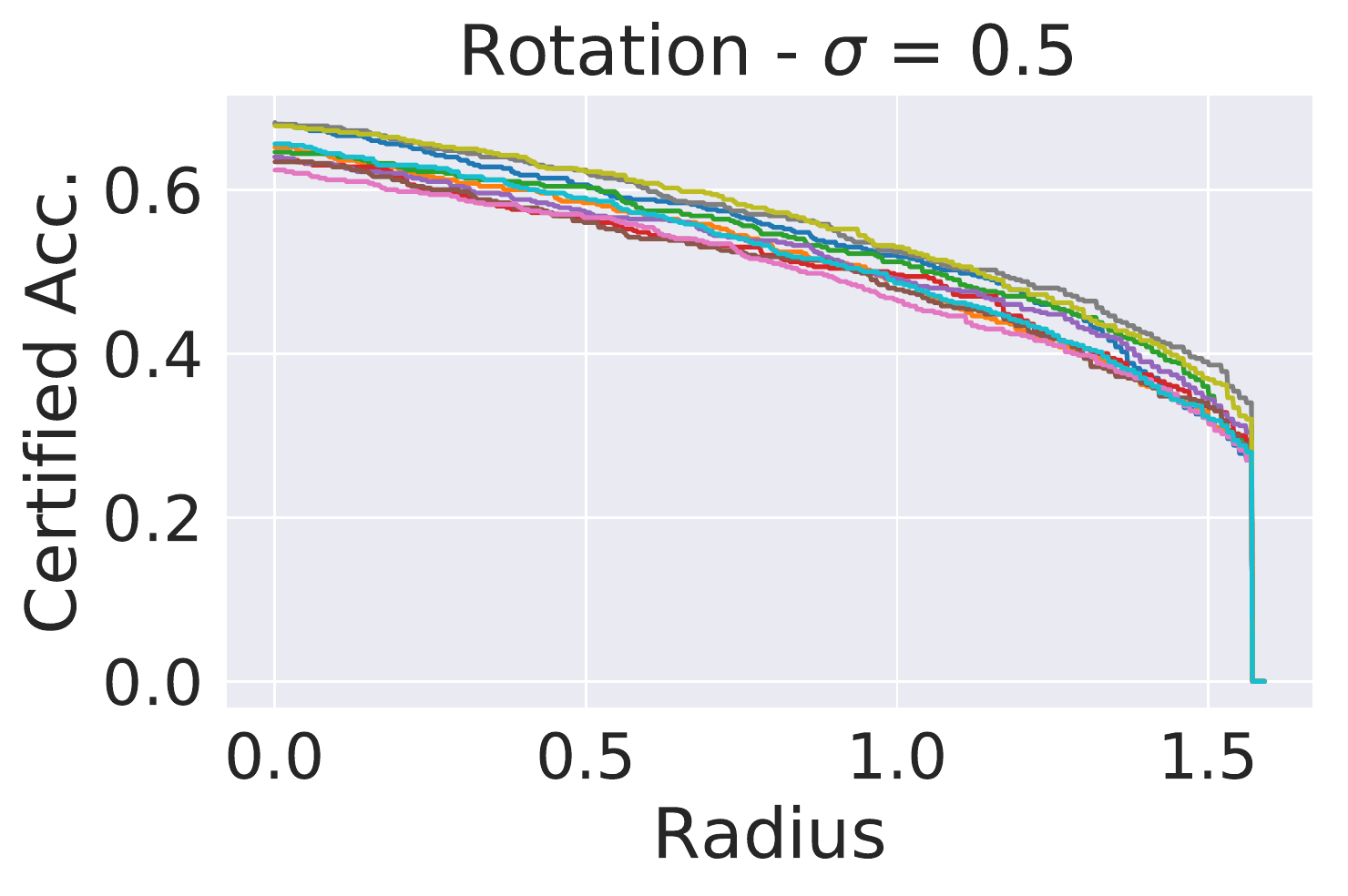}
    \caption{\textbf{Ablations for all clients. 10 Clients experiment} 
    Left: Local training. 
    Right: Personalized training.}
    \label{fig:ablations-all-clients-10-clients}
\end{figure}

 In all of our experiments, we reported the certified accuracy curves averaged across all clients, highlighting the minimum and the maximum performing clients in a shaded region.
 For completeness, we plot the certified accuracy for all clients for some of our experiments. 
 In particular, for experiments with 4 clients, we show our results in Figure~\ref{fig:ablations-all-clients-4-clients}, and show the results on 10 clients in Figure \ref{fig:ablations-all-clients-10-clients}.
 On each plot, we show the performance variation across clients when local training is conducted (compared to personalized training). 
 We observe, following our earlier observations, that the spread is significantly higher for local training. 
Note that this behaviour is despite the fact that we distributed the data in a uniform and homogeneous fashion across clients. 
Hence, in more realistic scenarios with heterogeneous splits, this spread could grow, highlighting the reliability of federated training and personalization in building accurate and robust models.
We believe this analysis is an interesting setup for future work.  

 \subsection{Experimental Details}
 The experiments reported in the paper used the MNIST and CIFAR-10~\cite{cifars} Datasets\footnote{Available \texttt{https://www.cs.toronto.edu/~kriz/cifar.html}, under an MIT license.}. 
 We used the available splits provided in PyTorch for training and testing sets.
 Moreover, we adopted the publicly-available codes for certification, as noted in the experimental section.

 \subsection{FL with Mixture Model}
\begin{figure}
    \centering
    \includegraphics[trim={0 10.4cm 0 0},clip , width=0.9\textwidth]{sections/images/mixture/common_legend.pdf}
    \includegraphics[width=0.49\textwidth]{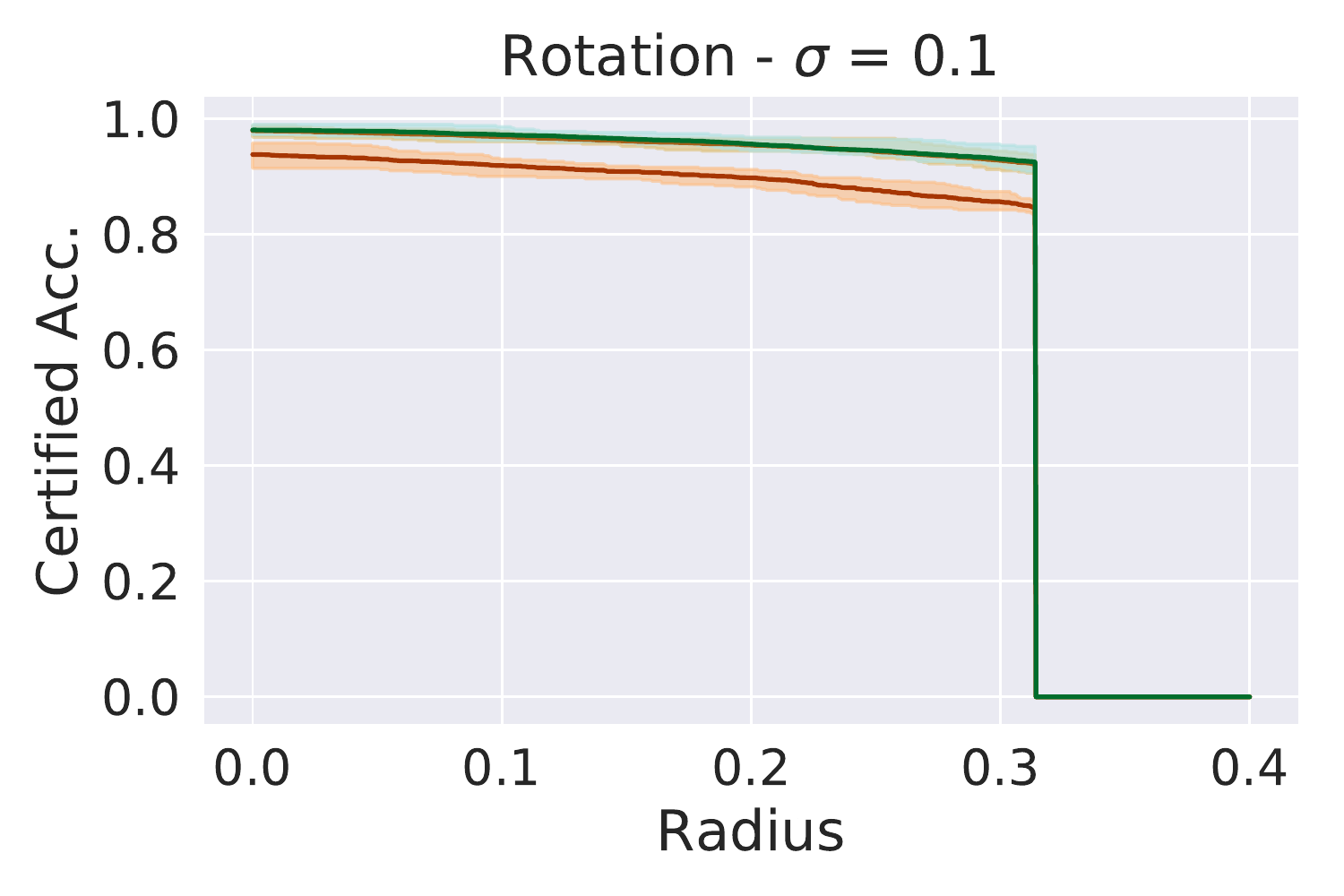}
    \includegraphics[width=0.49\textwidth]{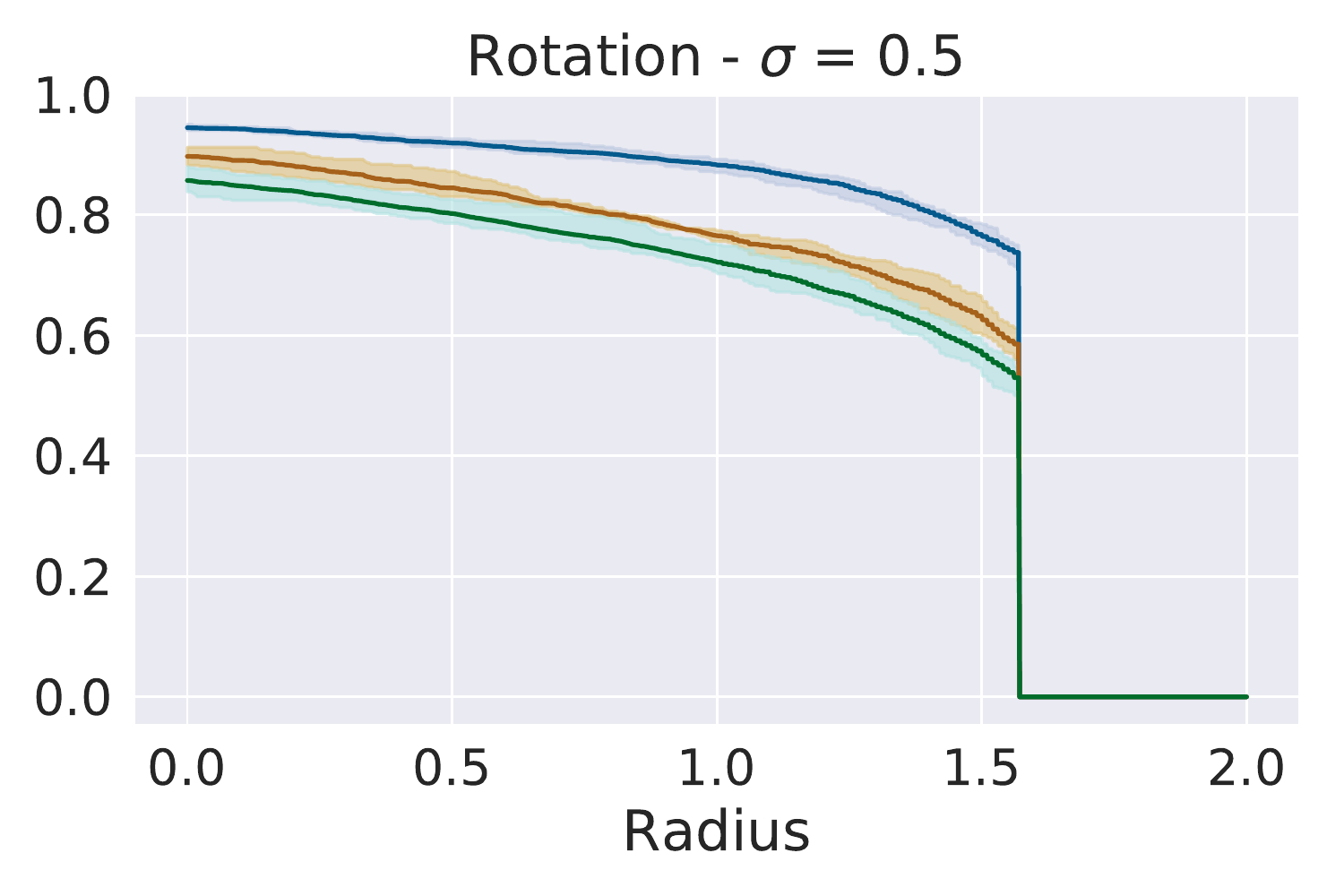}
    \includegraphics[width=0.49\textwidth]{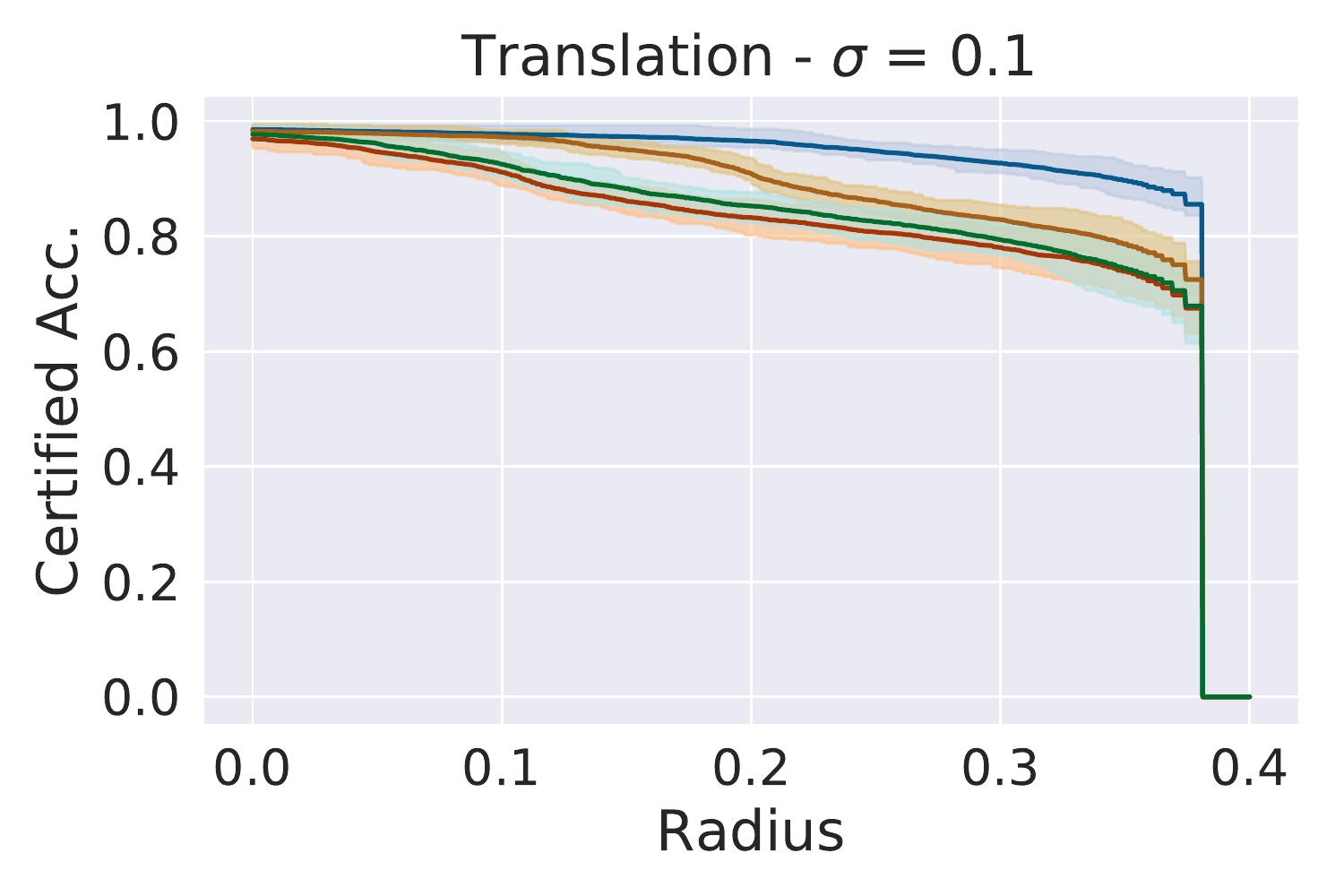}
    \includegraphics[width=0.49\textwidth]{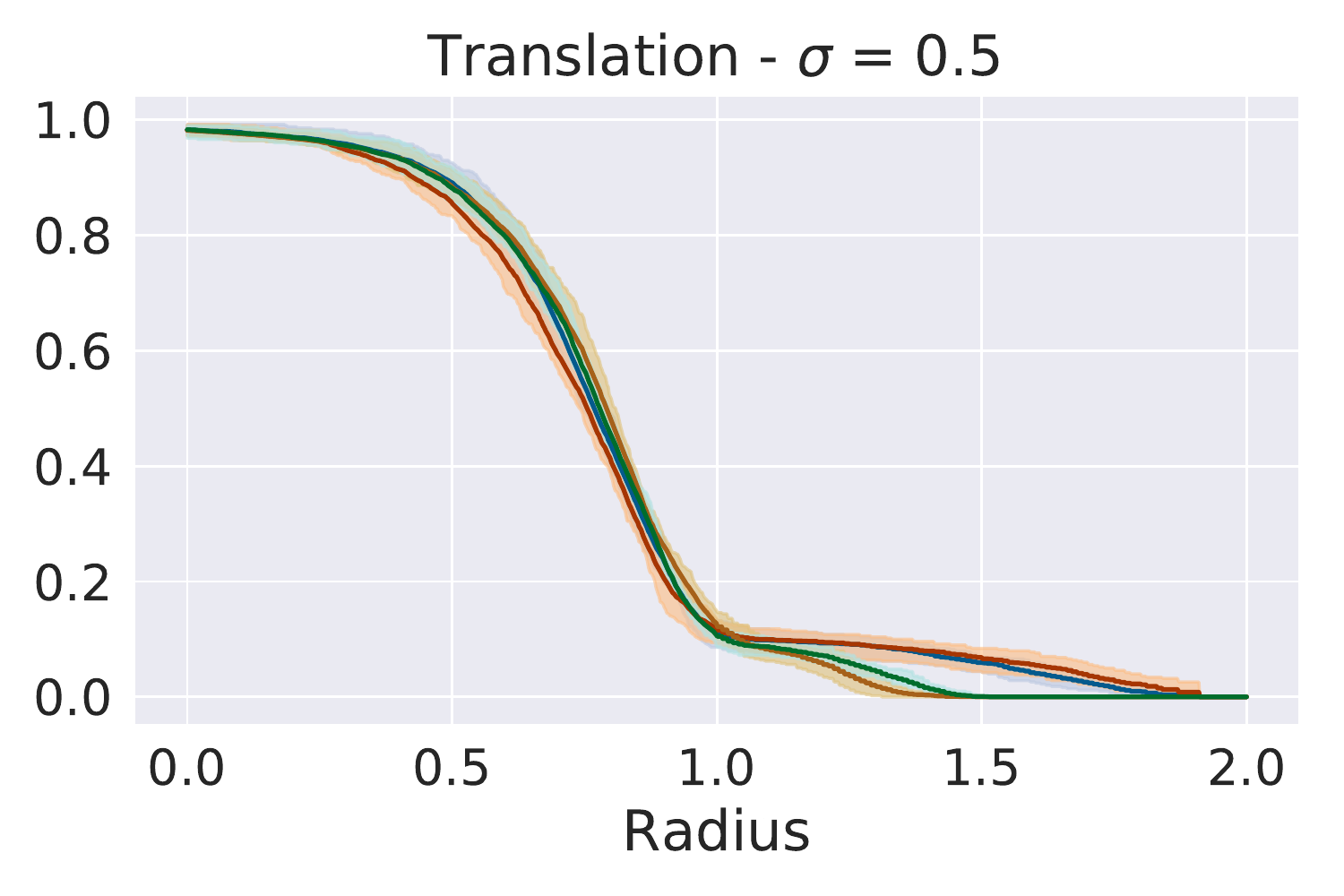}
    \includegraphics[width=0.49\textwidth]{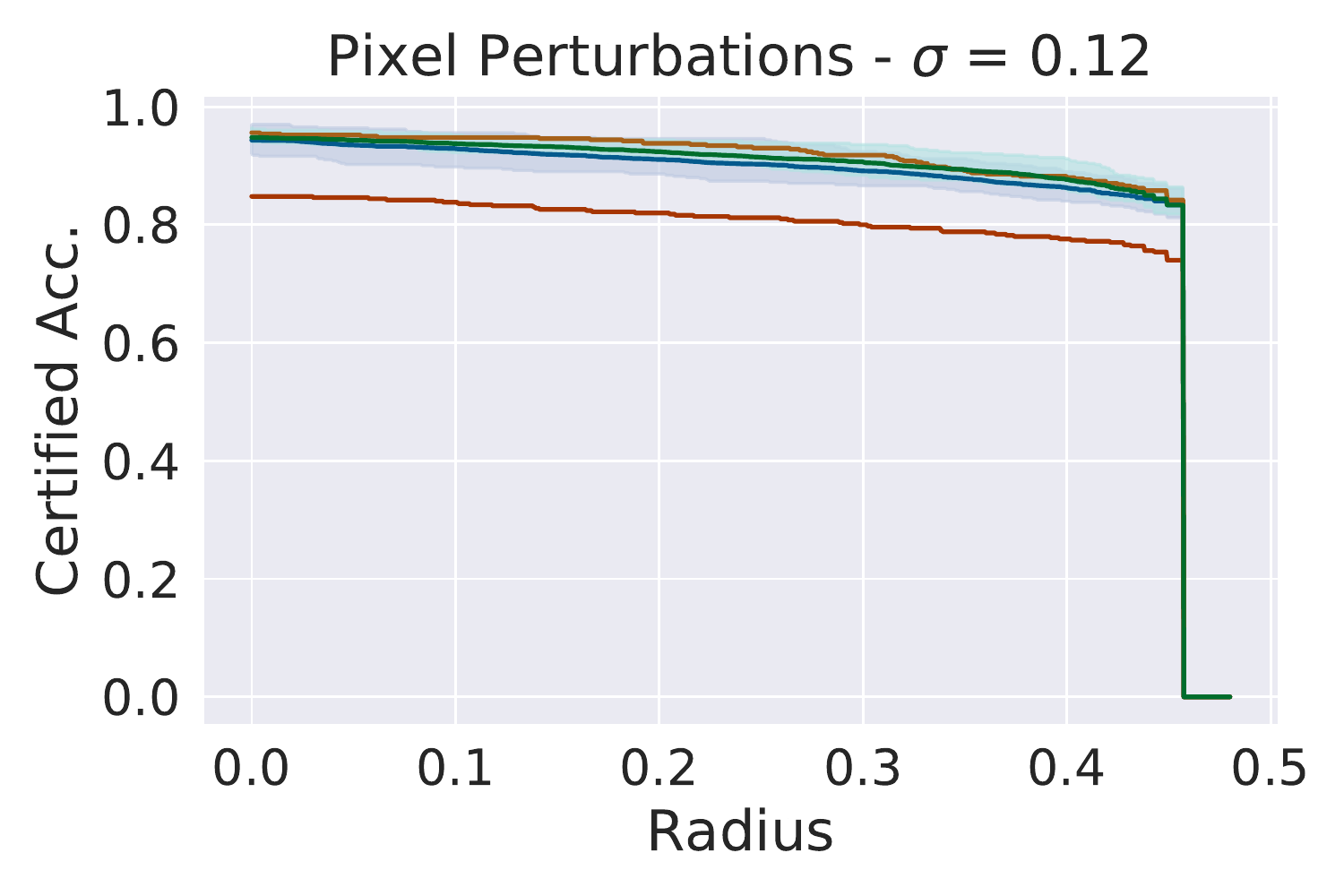}
    \includegraphics[width=0.49\textwidth]{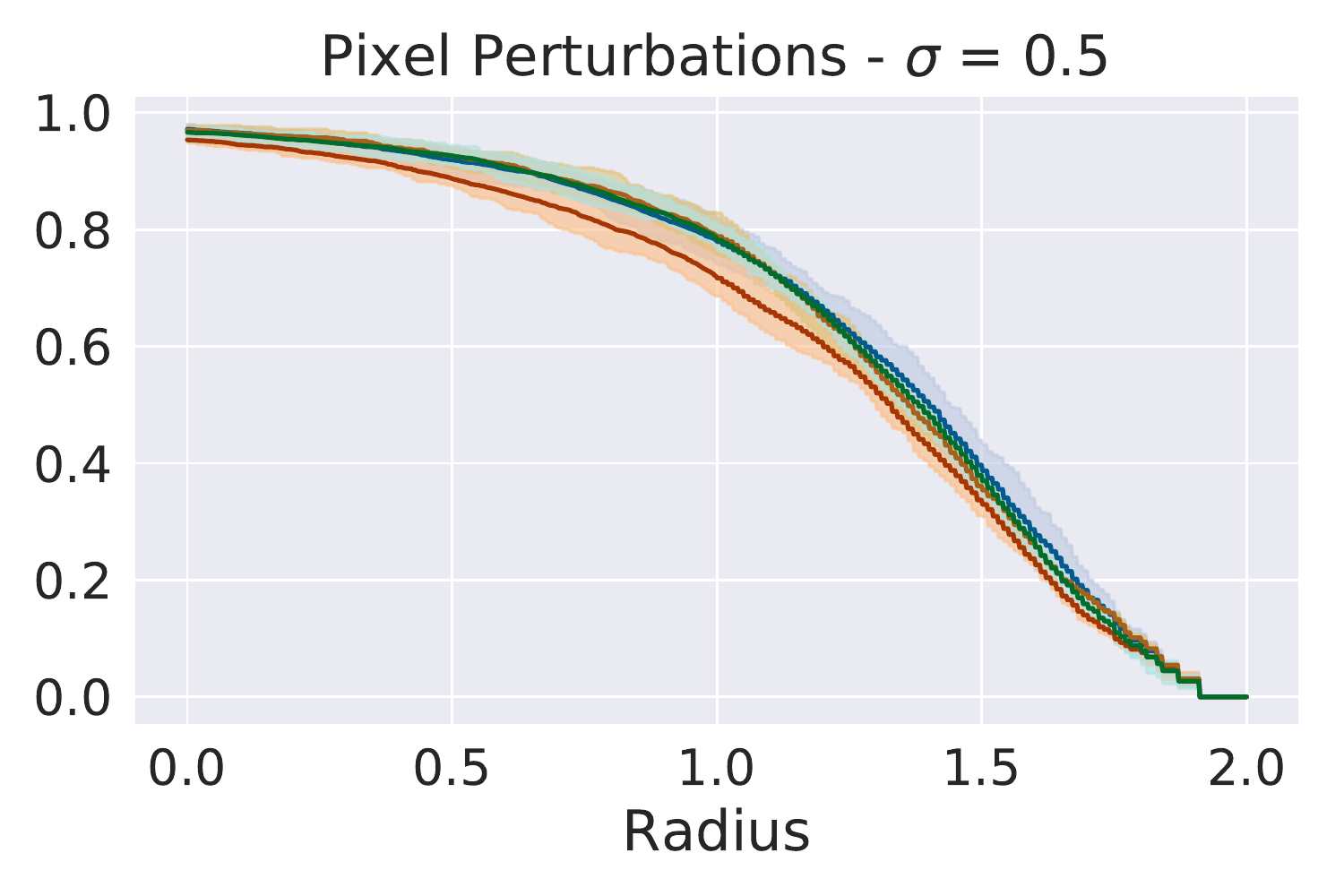}
    \includegraphics[width=0.49\textwidth]{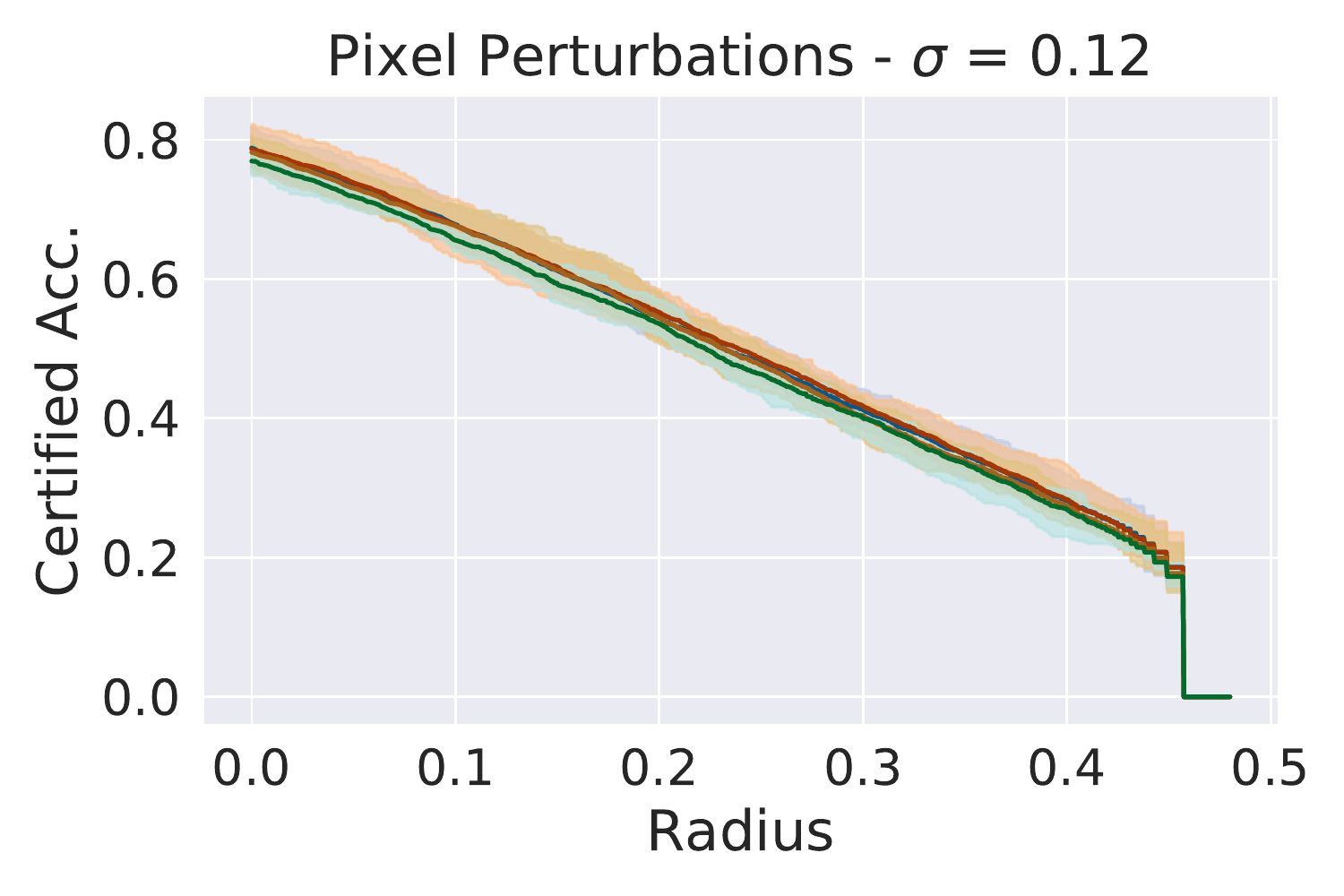}
    \includegraphics[width=0.49\textwidth]{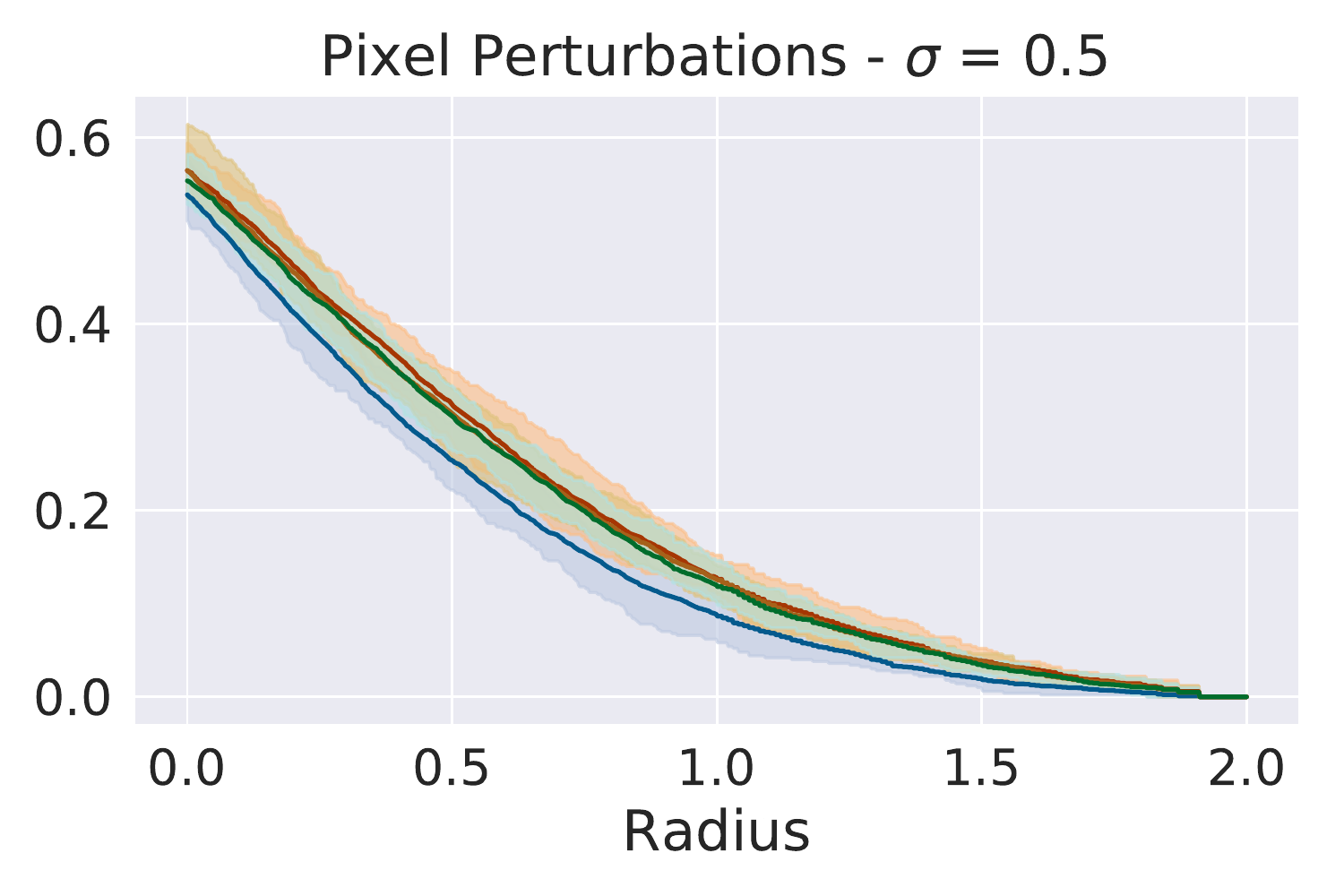}
    \caption{\textbf{FL with Mixture of Models.} We analyze the effect of mixture model on certified robustness on MNIST (first three rows) and CIFAR10 (last row). We observe that for most personalization values $\lambda$, the mixture of model outperforms the naive federated averaging model.
    }
    \label{fig:mixture-model-experiments}
\end{figure}

 At last, we experiment with the effect of the mixture model on the certified accuracy of the trained model. 
 In Section~4, our experiments analyzed rotation and translation deformations on CIFAR10. 
 For completeness, in Figure~\ref{fig:mixture-model-experiments} we also report results against (1) pixel perturbations on CIFAR10 and (2) rotation, translation, and pixel perturbations on MNIST. 
 Our results further confirm our observation in Section~4: the mixture model provides consistent improvement on the certified robustness for most considered values of $\lambda$. 
 It is also worth noting that different degrees of personalization perform the best for different perturbations and $\sigma$ values. % different personalization levels ($\lambda$ values) perform the best.
 
 \section{Social Impact}
 This paper explores (1) certified robustness and (2) federated learning.
 Both topics are associated to how machine learning settings relate to security which, in turn, has a social impact.
 In particular, robustness is associated with the secure deployment of models, hindering how malicious agents may attack recognition systems to fit their purposes.
 Furthermore, federated learning is a setting that arises naturally when data-privacy constraints are introduced into the learning objective: how to develop well-performing models that leverage user data while preserving privacy.
 
 \section{Compute Power}
 All in all, our work has the potential to be used by services targeting high-performance private models to benefit users.
 In particular, we used 20 Nvidia-V100 GPUs for all of our experiments. 
 
 \section{Limitations}
  There exist several frameworks that characterize federated training and personalization. 
A limitation of our work is the focus on the popular federated averaging technique. 
A possible interesting avenue for future work is to benchmark different federated and personalization frameworks in terms of robustness. 
Moreover, new mixture models could be devised that explicitly target robustness enhancement are left for future works.
Moreover, the development of new mixture models explicitly aimed at improving robustness and reliability is also left for future work.
At last, in this work we considered a homogeneous split of data across clients. That is, all clients have the same amount, and probably the same set of classes, of data.
We note here, while this setup is less realistic to the real world setup where different clients have heterogeneous data splits, we remark here that this serves as the best condition for local training.
That is, even when putting local training at advantage, our work showed the superiority of conducting federated training and personalization in providing models that are both accurate and robust.
We leave studying the heterogeneous setup for future works.
\end{document}